\documentclass[Afour, times, sageh]{sagej}
\usepackage{subcaption}
\usepackage{multirow}
\usepackage{tabularx}
\usepackage[T1]{fontenc} 
\usepackage{makecell}
\usepackage{enumitem}
\usepackage{moreverb,url}
\usepackage[colorlinks,bookmarksopen,bookmarksnumbered,citecolor=red,urlcolor=red]{hyperref}

\newcommand{\orcid}[1]{%
  \href{https://orcid.org/#1}{https://orcid.org/#1}%
}

\newcommand{\tabitem}{\textbullet~~}

\newcolumntype{C}{>{\centering\arraybackslash}p{1.3cm}}

\newtheorem{lemma}{Lemma}
\newtheorem{conjecture}{Conjecture}

\makeatletter
\def\@maketitle{%
\vspace*{-34pt}%
\null%
\begin{center}
\begin{sf}
\begin{minipage}[t]{\textwidth-57.625mm}
  \vskip 12.5pt%
    {\raggedright\titlesize\textbf{\@title} \par}%
    \vskip 1.5em%
    \vskip 12.5mm%
    \end{minipage}\hspace{15mm}\begin{minipage}[t]{42.625mm}
\hbox{}\scriptsize Preprint
\end{minipage}
{\par\large%
      \lineskip .5em%
      {\raggedright\textbf{\@author}
      \par}}
     \vskip 40pt%
    {\noindent\usebox\absbox\par}
    {\vspace{20pt}%
      {\noindent\normalsize\@keywords}\par}
      \end{sf}
      \end{center}
      \vspace{22pt}
        \par%
  }
\def\ps@title{%
\def\@oddhead{\parbox{\textwidth}{\mbox{}\\[-1pt]%
\noindent\rule{\textwidth}{0.5pt}}}%
\let\@evenhead\@oddhead
\def\@oddfoot{}\let\@evenfoot\@oddfoot}
\def\ps@sagepage{%
\let\@mkboth\@gobbletwo
\def\@evenhead{\parbox{\textwidth}{%
\normalsize\sagesf\thepage\hfill\itshape Preprint\\[-6pt]
\noindent\rule{\textwidth}{0.25pt}}}
\def\@oddhead{\parbox{\textwidth}{%
\normalsize\sagesf{\itshape{\leftmark}}\hfill\thepage\\[-6pt]
\noindent\rule{\textwidth}{0.25pt}}}
\def\@evenfoot{}\def\@oddfoot{}}
\makeatother

\begin{document}

\runninghead{Demont et al.}

\title{The Kinetics Observer: \\ A Tightly Coupled Estimator for Legged Robots}
% \author{
%     A. Demont, M. Benallegue, A. Benallegue, P. Gergondet, A. Dallard,\\ R. Cisneros-Limón, M. Murooka, F. Kanehiro% <-this % stops a space
% \thanks{A. Demont, M. Benallegue, P. Gergondet, R. Cisneros-Limón, M. Murooka and F. Kanehiro are with the CNRS-AIST JRL (Joint Robotics Laboratory), IRL, National Institute of Advanced Industrial Science and Technology (AIST), Tsukuba, Japan. A. Demont and A. Benallegue are also with Université Paris-Saclay, Gif-sur-Yvette, France, and Laboratoire d’Ingénierie des Systèmes de Versailles, France. A. Dallard is with the CNRS-University of Montpellier, LIRMM, UMR5506, Montpellier, France.}% <-this % stops a space
% \thanks{Manuscript received XX XX, XX; revised XX XX, XX.}
% }

\author{Arnaud Demont\affilnum{1}\affilnum{2}, Mehdi Benallegue\affilnum{1}, Abdelaziz Benallegue\affilnum{2}, Pierre Gergondet\affilnum{1}, Antonin Dallard\affilnum{3}, Rafael Cisneros-Limón\affilnum{1}, Masaki Murooka\affilnum{1}, and Fumio Kanehiro\affilnum{1}}

\affiliation{
\affilnum{1}CNRS-AIST JRL (Joint Robotics Laboratory), IRL, National Institute of Advanced Industrial Science and Technology (AIST), Tsukuba, Japan.\\
\affilnum{2}Université Paris-Saclay, Gif-sur-Yvette, France, and Laboratoire d'Ingénierie des Systèmes de Versailles, France. \\ 
\affilnum{3}CNRS-University of Montpellier, LIRMM, UMR5506, Montpellier, France.
}

\corrauth{Arnaud Demont, CNRS-AIST JRL (Joint Robotics Laboratory), National Institute of Advanced Industrial Science and Technology (AIST), 1-1-1 Umezono, Tsukuba, Ibaraki, 305-8560, Japan.}
\email{dem.arnaud2@gmail.com}

\begin{abstract}
This paper presents the Kinetics Observer, a novel proprioceptive state estimator for legged robots designed to provide the feedback required for versatile locomotion and physical interaction with the environment, with enhanced robustness to contact slippage. Its core contribution is a tight coupling between whole-body kinematics and external wrenches through a contact dynamics model, enabling the consistent fusion of leg kinematics, IMU, and wrench sensor measurements for real-time joint estimation of centroidal kinematics, contact rest poses, contact wrenches, and disturbance wrenches. By exploiting contact wrench measurements as correction terms, the proposed observer increases sensing redundancy, which provides observability of contact slippage relative to the centroid frame and enhanced robustness to modeling and sensor errors. The approach is experimentally evaluated on two humanoid robots across three scenarios totaling thirteen walking sequences, including long-distance walking with repeated contact changes, locomotion over slippery obstacles, and non-coplanar multicontact motion.
\end{abstract}

\keywords{Legged robots, State estimation, Proprioceptive odometry, Humanoid robots, Contacts estimation.}

\maketitle

\section{Introduction}
Legged robotics, particularly humanoids, have seen a major resurgence since the early 2020s. Demonstrations of robots performing increasingly complex and human-like tasks have multiplied, generating significant interest and raising expectations for applications such as manufacturing, logistics, healthcare, and assistance with daily activities. However, current control methods still lack robustness to external disturbances and modeling errors, which limits the deployment of reliable and safe behaviors~\citep{wu2024surveyLeggedRobots} in real-world environments.
Legged robots remain challenging to control because their underactuated dynamics imply that motion can only be generated through contact interactions with the environment~\citep{sugihara2020surveyDynamicsHumanoids}. Therefore, in order to achieve the desired motion of the robot, the appropriate contact forces must be applied~\citep{Wieber2016ModelingControlLeggedRobots}. 
 
In order to perform locomotion and execute autonomous tasks in the real-world, the robot must be able to locate itself in its environment. It therefore performs odometry, i.e., reconstructing its trajectory by incrementally integrating motion estimates obtained via heterogeneous measurements.
Furthermore, the dynamics and kinematics of legged robots impose constraints on the admissible motion of the center of mass that ensure balance~\citep{harada2006dynamicsBalanceHumanoid}. Beyond odometry, it is therefore essential to estimate the robot's full-body kinematics as precisely as possible to determine how the CoM should be accelerated through contact forces. This requirement is particularly critical for biped robots, whose margin for balance is relatively small. The estimation must therefore be highly reactive and robust to modeling errors, as even small deviations may lead to an irrecoverable fall.

\begin{figure}[!t]
\centering
\includegraphics[width=1\columnwidth]{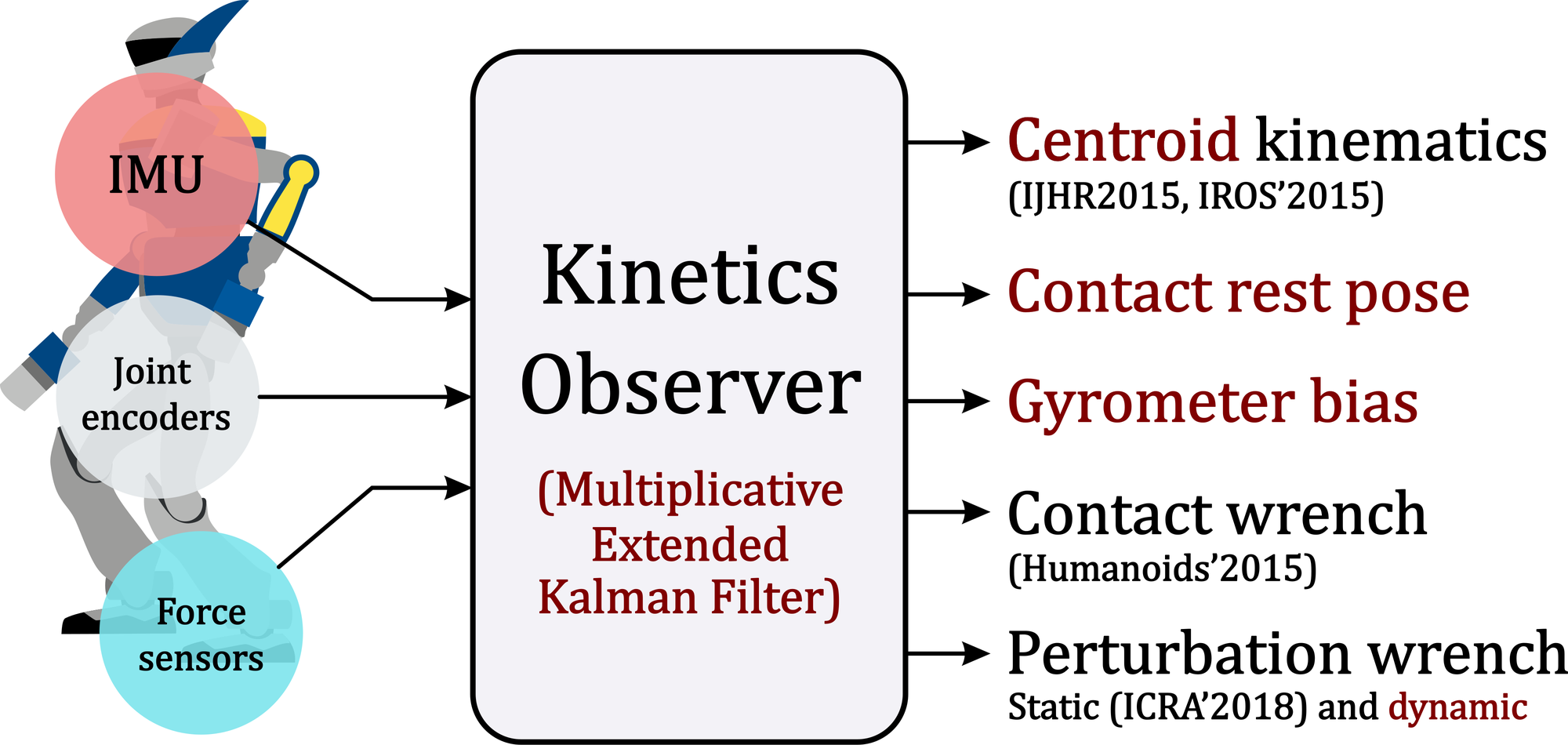}
\caption{Summary of the sensors we use and of the estimated state. We highlighted in bold red the new features compared to our prior works: IJHR2015~\citep{Benallegue2015EstimHumanoidFlexDeformationUsingOnlyImuAndContact}, IROS'2015~\citep{Mifsud2015EstimationContactForcesAndKinematicsWithOnlyImu}, Humanoids'2015~\citep{Benallegue2015FusionWrenchSensorsImuAndProprioceptionForKineDynObs}, ICRA'2018~\citep{Benallegue2018ExtForceEstimationWithNoTorqueMeas}.}\label{fig:summary}
\end{figure} 

State estimation for legged robot locomotion feedback has thus evolved along two main directions. The first focuses on kinematic estimation, seeking the best compromise between computational efficiency, robustness to dynamic motions (which notably induce contact slippage), and accuracy. The second aims to estimate external disturbance forces acting on the robot (e.g., unexpected interactions with humans), whose effects on the robot's dynamics must be accounted for in the control law. However, these two approaches have seldom converged, which hinders the deployment of agile and robust robotic behaviors capable of physically interacting with the environment.

\subsection{Motivation and Method}\label{subsec:motivation}
 
\noindent  In this section, we explain the motivation of our work by identifying what we consider to be remaining gaps in the state of the art of legged-robot state estimation. We also summarize what we propose to address them, while giving an outline of the core differences with respect to the state of the art,  which enable the contributions introduced in the next section. 
The objective of this work is to develop an estimator that provides the feedback needed for versatile legged-robot control, namely the robot's kinematics and the disturbance wrench applied to it. The estimation should be as robust as possible to sensor and modeling errors.

State estimation for legged robots has greatly advanced in the past decades, but we believe that this estimation robustness can be pushed further through the coupling between robot motion and the   wrenches that govern it (i.e. the \emph{kinetics}), which remains underexploited.  
The Kinetics Observer proposes to use kinetics to jointly estimate the contact rest poses and   wrenches alongside the robot's kinematics and disturbance wrench.   Just as including contact positions in the state enabled the  consistent fusion of IMU and leg   kinematics~\citep{bloesch2013FusionLegKineAndImu}, the Kinetics Observer  extends this consistency to wrench measurements. The latter can therefore be used to correct the estimated state, including the estimated kinematics and contact rest poses. 

This matters because kinematics-based proprioceptive estimators rely on the accuracy of their IMU and leg-kinematic measurements, and on the assumption that contacts are stationary. IMU measurements are affected by noise and bias drift, particularly when working with lower-grade sensors. Leg kinematics do not capture structural deformations under load: as illustrated in the top part of Figure~\ref{fig:Viscoelastic-model}, these deformations introduce errors in the modeled kinematic chain and thus in the leg-kinematic measurement, a core component of these estimators~\citep{bloesch2013FusionLegKineAndImu,Hartley2020RIEKF}. These errors are a known source of odometry drift~\citep{wisth2020preintegratedVelocityBias}. Additionally, contact slippage violates the stationary-contact assumption, while inaccurate contact detection or initialization introduces errors in the contact constraints. Without an additional source of information on the contacts,  these errors can be difficult to distinguish from actual robot motion and can therefore lead to odometry drift.

Contact wrench measurements provide the Kinetics Observer with a second source of information on the contacts, allowing it to identify inconsistencies between leg kinematics and measured wrenches through the viscoelastic model. It can therefore accommodate slippage and contact initialization errors without attributing them solely to the robot motion. 
This new capability is enabled by significant structural differences with standard Leg-Inertial estimators, especially in the way the prediction is generated and constrained. 
In such estimators, the prediction is directly inferred from sensor measurements: the IMU is used as an input to predict the robot motion, and \citet{Fourmy2021ContactForcesPreintegrationFactorGraphs} extend this by using contact force measurements as \emph{inputs} to additionally predict the centroid motion through the robot's dynamic model. Leg kinematics, used as observations, relate the IMU pose to that of the foot, providing a correction of the robot kinematics that anchors the robot in the world. 
In the Kinetics Observer, the prediction instead originates from the system model: the motion is predicted through the \emph{estimated} contact forces and the robot's dynamic model. The IMU is then not used as a feed-forward on the robot kinematics, but as an observation within the filter, which notably observes and thus corrects the estimated wrenches. This property is what allowed us, in previous works,  to estimate the resulting   wrench applied to the robot using the IMU alone~\citep{Mifsud2015EstimationContactForcesAndKinematicsWithOnlyImu,Benallegue2018ExtForceEstimationWithNoTorqueMeas}, assuming fixed contacts. Likewise, wrench measurements are not used as a feed-forward but as an observation. Contact wrenches are \emph{state variables}, and are \emph{predicted} by the viscoelastic model from the contact rest poses and the contact kinematics, obtained from the centroid kinematics and the leg kinematics, which are thus an input of the model rather than an observation. The discrepancy between the wrench prediction and measurement  thus \emph{corrects} the contact wrenches, the centroid kinematics \emph{and} the contact rest poses. The strength of this model is that the wrench measurement, compared to this prediction, carries information not only on the instantaneous robot motion, as the force does when used as a feed-forward, but on relative poses, providing correction even after the motion has finished.  This notably makes the slippage of the contacts observable relative to the centroid frame (Section~\ref{subsec:observability}).

Finally, to accommodate slippage, standard Leg-Inertial methods allow for some correction of the contacts through process noise on their evolution. While this has been shown to improve odometry performance~\citep{bloesch2013FusionLegKineAndImu}, it also introduces the possibility of free drift of the robot, as its anchoring in the world is loosened. Our work addresses this issue by constraining the contact-pose process uncertainty, which prevents a common drift of the contact rest poses while allowing their relative positions and orientations to be corrected.

The Kinetics Observer therefore provides robot kinematics, contact wrenches and rest poses,  and the disturbance wrench in a single real-time filter, whose robustness to sensor and modeling errors is evaluated.

\subsection{Contributions}\label{subsec:contribs}

\noindent The main contributions of this work are:
\begin{itemize}[leftmargin=1em]
    \item {The Kinetics Observer jointly estimates the robot's
    kinematics, the resulting disturbance wrench applied to it,
    and the contact wrenches and rest poses. To the best of our
    knowledge, it is the first estimator for legged robots to
    perform the \emph{consistent} fusion of leg kinematics, IMU,
    and wrench-sensor measurements, thereby improving robustness
    to modeling and sensor errors.}

    \item {Through this consistent fusion, the Kinetics Observer
    uses contact wrench measurements to \emph{correct} the contact
    rest poses. We establish the observability of these poses
    relative to the centroid frame, and thus of local contact
    slippage and local support geometry (the position and
    orientation of the support area).
    We experimentally demonstrate the resulting robustness
    to contact slippage and substantial errors in contact
    pose initialization.}

    \item {We introduce the \emph{constrained contact process
    covariances}, which relax the fixed-contact assumption while
    preventing contact process noise from introducing a common
    drift of the contact rest poses. This allows their relative
    positions and orientations to be corrected while preserving
    the collective anchoring provided by the contacts.
    This constraint can also be applied within standard Kalman
    filters for proprioceptive odometry. It was notably applied to the RI-EKF in Section~\ref{subsec:gyro-bias-drift}.}

    \item {We evaluate the estimator in thirteen experiments using
    the humanoid robots RHP Friends and HRP-5P, including
    long-distance walking, walking with slipping contacts, and
    multicontact locomotion with an additional hand contact.
    Beyond odometry accuracy, we evaluate disturbance-wrench
    estimation and robustness to gyrometer bias drift,
    contact initialization errors, and contact flexibility
    identification errors. We also discuss how the estimator
    handles dynamic-model errors.
    The KO-Lin variant further shows that combining robot dynamics,
    contact force measurements, and the linear viscoelastic model
    already provides accurate estimation, even without contact
    orientations or moment measurements, while the full model
    provides better yaw accuracy.}

    \item {The code of the Kinetics Observer is available as an
    open-source framework.\endnote{\url{https://github.com/jrl-umi3218/state-observation}}}
\end{itemize}

Compared with our previous
work~\citep{Benallegue2018ExtForceEstimationWithNoTorqueMeas},
we introduce contact rest poses into the estimated state,
establish their observability, and adapt the viscoelastic model
to compute contact deformations relative to these estimated poses.
Another core change is to express the estimated robot kinematics
as those of the centroid frame, which simplifies the robot
dynamics equations. Together with the Multiplicative Extended
Kalman Filter formulation respecting the Lie-group structure
of the state, this required adapting the entire system model.
We derive the corresponding analytical state-transition and
measurement Jacobians, detailed in~\citet{demont2025Thesis}.
The estimator now also estimates gyrometer biases.

\noindent Additional features of the Kinetics Observer include:
\begin{itemize}[leftmargin=1em]
    \item {It does not require joint torque measurements, which
    can be difficult to obtain accurately, particularly on
    position-controlled robots because of friction, backlash,
    and transmission-side uncertainties.}

    \item {It is not restricted to a specific number of contacts,
    wrench sensors, or IMUs, and their placement on the robot
    is not constrained.}

    \item {It can operate without contact orientations or moment
    measurements, as demonstrated by KO-Lin. Without wrench
    sensors, it can also perform Leg-Inertial odometry,
    including contact orientations when available.
    The contact representation can be extended to point
    and edge contacts.}
\end{itemize}

Figure~\ref{fig:summary} summarizes the measurements and estimated
state of the Kinetics Observer, highlighting the additions
relative to our previous
work~\citep{Benallegue2018ExtForceEstimationWithNoTorqueMeas}.

After reviewing the state of the art of state estimation for
legged robots in Section~\ref{sec:related-work}, we introduce
the necessary background in Section~\ref{sec:Preliminaries}.
We then define the model at the core of the Kinetics Observer
in Section~\ref{sec:ProblemStatement} and present its experimental
evaluation in Section~\ref{sec:Experiments}.
Finally, we conclude and discuss future directions
in Section~\ref{sec:conclusion}.

\section{Related Work} \label{sec:related-work}
\subsection{High-Frequency Kinematics Estimators} \label{sec:high_freq_kine}

High-frequency state estimators play a central role in the control loops of legged robots, providing reactive feedback required for balance control and locomotion execution~\citep{fallon2014driftFreeHumanoidStateEstimation}. 

These estimators typically rely on proprioceptive sensors due to their high bandwidth. Especially, IMU measurements are combined with joint encoder readings, providing leg kinematic measurements under the assumption that contacts are  zero-velocity points of the robot in the world. This fusion provides observability of the robot's tilt and local linear velocity~\citep{Lin2006HexapodDynamicalGaits}, which are essential for stabilizing the robot~\citep{englsberger2013ThreeDimBipedalWalkingDCM}. Without exteroception, position and yaw angle in the world remain unobservable, leading to drifts in the odometry performed by these methods (commonly referred to as Leg-Inertial Odometry). These drifts can be due to various reasons, such as contact slippage, compliance, uncertainties in contact detection, etc. Nevertheless, proprioceptive odometry remains necessary for walking controller feedback, since methods using exteroception run at too low frequency, introduce latency in the feedback, and may produce discontinuities in the estimated pose~\citep{fallon2014driftFreeHumanoidStateEstimation}. Efforts have thus been made to reduce the drift in proprioceptive odometry, notably by robustifying it to the main modeling errors affecting it. 

Contact positions can be added to the estimated state and thus be partially corrected, enabling consistent fusion of IMU and leg-kinematic measurements and helping reduce drift caused by contact slippage~\citep{bloesch2013FusionLegKineAndImu,Hartley2020RIEKF}. \citet{Rotella2014StateEstimationForHumanoid} extended this approach to humanoid robots by adding contact orientations to the estimation. 

Since Leg-Inertial odometry relies heavily on the assumption that contacts remain fixed in the world, improvements can also be achieved by better estimating the time a contact is firmly set and can be used in the estimation. A common approach is to threshold the Ground Reaction Force (GRF)~\citep{ito2000StandingPostureControlBasedOnGrf,Focchi2013LocalReflexGenerationForQuadrupedObstacleNegotiation,Wu2016GrfSensingAndTerrainClassification}, but it is notably affected by the perturbations of the GRF during impact and by slippage. \citet{lin2021legged} implemented a detection based on a neural network that uses the IMU measurements and the joint encoders of quadruped robots to deal with this bouncing problem. \citet{maravgakis2023probabilistic} proposed an alternative using IMUs at the end-effectors to robustify contact detection in the presence of slippage. \citet{yoon2023InvariantSmootherDynamicContactEventInformation} introduced a method to detect and handle dynamic contacts based on the thresholding of the estimated foot velocity and acceleration. Instead of ignoring contact information in the case of dynamic contacts, they increase the covariance associated with the static contact foot assumption. They also enforce the static contact condition with a dedicated cost in their invariant smoother when the conditions are met. Finally, \citet{buchanan2022LearningInertialOdometry} addressed the slippage problem by estimating the sliding displacement with a Convolutional Neural Network, which uses the IMU's measurements.

Another challenge of proprioceptive odometry, inherent to leg kinematics measurements, is that of flexibilities, either of the environment or of the robot's structure~\citep{Mikami2014IdentificationHRP2Dynamics}. The latter notably induce a deformation of the robot's geometry under load, leading to errors in its forward kinematics model, thus altering leg kinematic measurements. These flexibilities are a known problem in legged robot control, notably for bipeds, as they make their control challenging~\citep{Romualdi2022WholeBodyControlEstimation}. In state estimation, very few methods seem to try and compensate for the effect of internal flexibilities, although they cause inconsistencies between leg kinematics and those measured by the IMU, and notably to a non-negligible biased drift in the robot's estimated pose throughout the walk.  \citet{Flayols2017ExperimentalEvaluationOfSimpleEstimForHumanoidRobots} proposed to use contact wrench sensors and identified flexibilities characteristics of the robot HRP-2's ankle, to account for its deformation in the IMU's pose estimate. Observing the repeatable nature of the odometry drift caused by flexibilities, \citep{wisth2020preintegratedVelocityBias}  represented it as a bias in the velocity measurement provided by Leg odometry. To suppress this drift, they introduced a zero-velocity measurement that was incorporated into the estimation whenever at least two of the three sensing modalities (IMU, camera, and Leg odometry) indicated no motion. While this method appears to be effective, it was primarily enabled by the use of exteroceptive sensing, which provided reliable global pose observation. 

Finally, we note that other biases in the model or the sensor measurements are more commonly estimated to improve the estimation, notably that in the IMU measurements~\citep{bloesch2013LeggedRobbotsUnstableSlipperyTerrain,Yang2019ContactCentricLegOdometry,Wang2017VelocityEstimAlgorithmForLeggedRobot}, or in the kinematics of the robot's center of mass~\citep{Masuya2020ReviewStateEstimHumanoidRobots,Fourmy2021ContactForcesPreintegrationFactorGraphs}. The latter is analogous to a constant external force applied to the robot.

\subsection{Exteroceptive Odometry}\label{sec:extero-odom}
The most accurate odometry results are typically achieved by incorporating additional spatial awareness from exteroceptive information, notably adding robustness to contact slippage. In 2020, Camurri proposed to incorporate pose corrections from stereo vision and LIDAR measurements into an Extended Kalman Filter. To deal with the latency and asynchronous arrival of these measurements, the EKF was running a high-rate Leg-Inertial odometry as the nominal process at every iteration while keeping a history of about 10 seconds containing all state, inputs, measurements, and covariances. Upon arrival of a delayed exteroceptive measurement, the estimator replayed the buffered history from the time the measurement was acquired so that the correction could be applied consistently. While this method provided satisfactory accuracy overall, its performance would degrade significantly upon failure of one of the measurement sources due to the loose coupling between them. In~\citet{wisth2022vilens}, PRONTO was reformulated within a factor graph framework~\citep{dellaert2017factor}. This formulation allowed raw sensor measurements to be incorporated as factors that jointly constrain the state over a sliding window through nonlinear optimization, thereby improving robustness to failures or degradation in individual sensors. 

\subsection{External Forces Estimation}

Since contacts are a key element of the control of legged robots, controlling contact forces is crucial not only for maintaining the robot's balance, but also for accurately generating the desired motions~\citep{lengagne2013generationMultiContactMotion}. When these forces cannot be measured, they therefore must be estimated. Dedicated methods use the robot's joint torque measurements to estimate the contact forces that make the robot's dynamics consistent.   \citet{Xu2016SixLeggedRobotStepObstaclesByIndirectForceEstimation} combined the kinematic model of the leg of their hexapod robot with a model of the force applied to its tip, to estimate the latter from the output torque of the three motors actuating the leg. \citet{Cong2020contact_force_estim_and_application_in_impedance_control} used a relationship between the generalized momentum of their robot and the contact force at the newly created contact of their quadruped through the Jacobian matrix of the legs.  \citet{xinjilefu2014dynamicStateEstQuadProg} formulated their estimator as an optimization problem solved using Quadratic Programming. This enabled the inclusion of constraints, notably on joint limits, torque limits, and contact forces. However, this method requires the base pose as an input. The latter must therefore be estimated by another estimator, such as that introduced in Section~\ref{sec:high_freq_kine}, running in parallel (e.g.~\citet{piperakis2018NonlinearStateEstimationForHumanoidRobotWalking}). A direct consequence of this unidirectional dependence on the kinematics estimate is that errors in the latter inevitably propagate to the downstream estimate. \citet{varin2020constrainedKfFrictContact} proposed an alternative method based on a Constrained Kalman Filter. By introducing richer contact constraints, including non-penetration and friction constraints, they addressed some of the limitations of \citet{xinjilefu2014dynamicStateEstQuadProg}, in particular its reliance on a kinematics prior. However, this method requires prior knowledge of the ground geometry and the contact friction coefficients.

Additionally, the methods introduced above consider that external forces consist only in contact forces, and thus don't consider potential disturbances, which  are unanticipated or at least not applied at a fixed point / surface in the world (e.g. the robot being pushed by a human), yet directly affect the robot's dynamics and thus its trajectory. 

In order for the robot to engage in more complex and versatile physical interaction with the environment, including interactions with humans~\citep{Kaneko2012DisturbanceObserverExtForceOnHumanoid}, it is therefore mandatory to estimate and compensate for the resultant of the disturbance force applied to the robot's CoM. An exhaustive review of methods for its estimation is provided in~\citet{Masuya2020ReviewStateEstimHumanoidRobots}.
Notably, \citet{Kaneko2012DisturbanceObserverExtForceOnHumanoid} used a simple model to estimate the resultant external disturbance force exerted on the center of mass from its measured linear acceleration (assuming that the IMU is located at the center of mass) and the measured force at the feet. More recent methods have additionally allowed the estimation of the external moment applied to the robot. To achieve this, \citet{flacco2016ResidualBasedContactsEstimationHumanoidRobots} notably adapted the residual method, already used for fixed-base robots, to floating-base robots equipped with joint torque sensors. The introduced estimators show strong overall performance. However, similarly to estimators for contact forces such as~\citet{Xinjilefu2015StateEstimationHumanoidRobots}, they rely on the a priori knowledge of the robot's kinematics and are thus also highly dependent on their accuracy and robustness. 

In a previous work~\citep{Benallegue2018ExtForceEstimationWithNoTorqueMeas}, we proposed to couple kinematics and contact wrenches via a dedicated dynamic model of contacts, allowing for the simultaneous estimation of the kinematics and the disturbance wrench. This mitigates the impact of errors in one estimate on the other. This method also has the advantage of using solely IMUs, joint encoders, and contact force sensors. It thus does not rely on joint torque measurements, which are not necessarily available in legged robots. The Kinetics Observer introduced in this paper extends that work by incorporating contact rest poses into the estimated state. This enables full consistency between leg kinematics, IMU measurements, and wrench sensor data, thereby further improving robustness to contact slippage and other modeling errors.

\subsection{State Estimation Methods for Legged Robots}

Since the 2000s, state estimation for legged robots has increasingly focused on providing reliable feedback for closed-loop control in real-world deployment. Extended Kalman Filters (EKFs) have remained predominant due to their computational efficiency and their ability to fuse system models with sensor measurements under Gaussian noise assumptions~\citep{barshan1993inertialNavigationSystemMobile}. However, their performance strongly depends on how estimation errors are represented and propagated.

One important development was the adoption of mathematically consistent representations of orientation. Earlier approaches commonly used a three-parameter vector representations such as Euler angles~\citep{Setoodeh2004AttitudeEstimSeparateBiasKalman, Lin2006HexapodDynamicalGaits}. However, these parameterizations introduce singularities and do not respect the Lie group structure of rotations ($SO(3)$), inducing improper uncertainty propagation and degrading estimation performance. To overcome these issues, the Multiplicative EKF~\citep{markley2004MEKF}, including the Quaternion EKF~\citep{bloesch2013FusionLegKineAndImu}, relies on  nonsingular orientation representations while expressing errors with a three-dimensional perturbation. This formulation enables consistent composition and uncertainty propagation by operating directly on the geometry of $SO(3)$. The use of Lie group theory into EKFs was later formalized by~\citet{Bourmaud2013DiscreteEkfLieGroups}, and has since become standard practice in legged robot state estimation~\citep{Camurri2020ProntoAM, Ramadoss2021DiligentKIO, teng2021legged, kim2022step}, which we also follow in this work.

Another approach to improve closed-loop control aims at providing feedback signals with formal guarantees on the stability of the estimation error. This ensures that the estimation error converges to zero rather than diverging when the estimate deviates from the true trajectory, which may notably happen in classical EKFs~\citep{Barrau2017IEKF}.  
To address this issue, Bonnabel proposed the Invariant Extended Kalman Filter (IEKF), which exploits the symmetry properties of systems evolving on Lie groups~\citep{Bonnabel2007LeftInvariantEkfAndAttitudeEstim,Bonnabel2008SymmetryPreservingObservers, Barrau2018InvariantKalmanFiltering}. By defining invariant estimation errors that respect these symmetries, the resulting error dynamics become independent of the system trajectory, reducing  errors due to linearization and providing local asymptotic stability guarantees. However, enforcing invariance can become challenging for systems with complex variable couplings~\citep{ramadoss2022StateEstimHumanMotionLocomotion}. 
Besides EKFs, nonlinear complementary filters~\citep{mahony2008NonlinearComplementaryFiltersOnSO3} have also gained in popularity in this context, as they enable the use of nonlinear analysis tools to establish stability guarantees that may even hold globally. ~\citet{benallegue2020LyapunovStableOrientationEstimatorHumanoids} proposed a nonlinear complementary filter for estimating a legged robot's IMU tilt and local linear velocity, for which almost global asymptotic stability of the estimation error is proven.

Finally, factor graph methods have gained increasing attention for state estimation in legged robotics since the late 2010s. In this framework, state variables are represented as nodes in a graph, while factors encode constraints derived from system models or sensor measurements. State estimation is formulated as a maximum a posteriori (MAP) inference problem over the graph. As discussed in Section~\ref{sec:extero-odom}, this formulation   notably enabled the incorporation of delayed exteroceptive measurements with enhanced robustness to individual sensor failures compared to EKF-based approaches.

Factor graphs have also been applied to purely proprioceptive estimation, particularly due to their ability to represent more complex system models and to relinearize past states during optimization~\citep{Agrawal2022ProprioceptiveEstimWithKinemChainModeling, Fourmy2021ContactForcesPreintegrationFactorGraphs}. \citet{yoon2023InvariantSmootherDynamicContactEventInformation} further proposed a factor-graph-based smoother for legged robot kinematic state estimation that exploits the same invariance properties as the IEKF~\citep{Chauchat2018InvariantSmoothingLieGroups}, and allowed to incorporate  dynamic contact events classification and apply the corresponding constraints into the estimation process.

\section{Preliminaries}\label{sec:Preliminaries}

\subsection{General Notations}
\begin{itemize}[leftmargin=1em]
\item {Different notations are used to distinguish estimated ($\hat{\boldsymbol{\circ}}$), predicted ($\bar{\boldsymbol{\circ}}$), and error ($\tilde{\boldsymbol{\circ}}$) variables.}
\item {$\mathbb{I}_{n\times n}$ and $\mathbb{O}_{n\times n}$ are, the $n\times n$ identity and zero square matrices, respectively. If the dimension $n$ is not given, the $3\times 3$ matrix is implicitly referred to.}
\item {Reference frames are represented by uppercase calligraphy variables (e.g., $\mathcal{W}$ for the world frame and $\mathcal{C}$ for contact frames) or uppercase Greek letters (e.g., $\Gamma$ for the centroid frame).}
\item {Kinematic variables (position, orientation, velocities and accelerations) are represented using the notation $^{\mathcal{B}}\bigcirc_{\mathcal{A}}$, which expresses the kinematics of the frame $\mathcal{A}$ in the frame $\mathcal{B}$. To simplify the notation, we omit the world frame symbol $\mathcal{W}$ whenever it can be inferred from the context: $^{\mathcal{W}}\bigcirc_{\mathcal{A}}=\bigcirc_{\mathcal{A}}.$}
\item {The matrix representation of a rotation $\boldsymbol{R}$ is used in the developed equations, but to keep rigorous notations, we also use a vector representation $\boldsymbol{\Omega}$ when necessary. For example, $^{\mathcal{B}} \boldsymbol{\Omega}_{\mathcal{A}}$ refers to the vector representation (e.g., quaternions, Euler angles, etc.) of the rotation $^{\mathcal{B}}\boldsymbol{R}_{\mathcal{A}}$.}
\end{itemize}
The state of our system lies in a high-dimensional Lie group consisting of components from $\mathbb{R}^{3}$ and components of the Lie group of 3D rotation matrices $\text{SO}\!\left(3\right)$. While we recommend reading \citet{Barfoot2017StateEstimationForRobotics} for a more in-depth explanation of Lie groups, we provide here the notations we use in this paper for the $\text{SO}\!\left(3\right)$ group and its Lie Algebra the group of skew-symmetric matrices, denoted $\mathrm{so}\!\left(3\right)$: 

\begin{itemize}[leftmargin=1em]
   \item {$\oplus$ and $\ominus$ are the sum and difference operators between two elements of a Lie group, respectively.}
   \item {$\left[.\right]_{\times}:\mathbb{R}^{3}\rightarrow\mathrm{so}\!\left(3\right)$ is the \emph{hat} operator of $\text{SO}\!\left(3\right)$. It corresponds to the skew-symmetric operator.}
\item {$\left[.\right]_{\vee}:\mathrm{so}\!\left(3\right)\rightarrow\mathbb{R}^{3}$ is the \emph{vee} operator of $\text{SO}\!\left(3\right)$, defined so that $\left[\left[\boldsymbol{v}\right]_{\times}\right]_{\vee}=\boldsymbol{v}\text{ .}$}
   \item {We note $\text{Exp}:\mathbb{R}^{3}\rightarrow\text{SO}\!\left(3\right)$ the operator that combines the $\text{SO}\!\left(3\right)$ \emph{exponential map} with the \emph{hat} operator.}
   \item {Conversely, we note $\text{Log}:\text{SO}\!\left(3\right)\rightarrow\mathbb{R}^{3}$ the operator that combines the $\text{SO}\!\left(3\right)$ \emph{logarithm map} with the \emph{vee} operator.}
\end{itemize}

\subsection{Frames and Variables Definition} \label{subsec:FramesVarsDef} 
\begin{figure}
   \centering
         \includegraphics[width=\columnwidth]{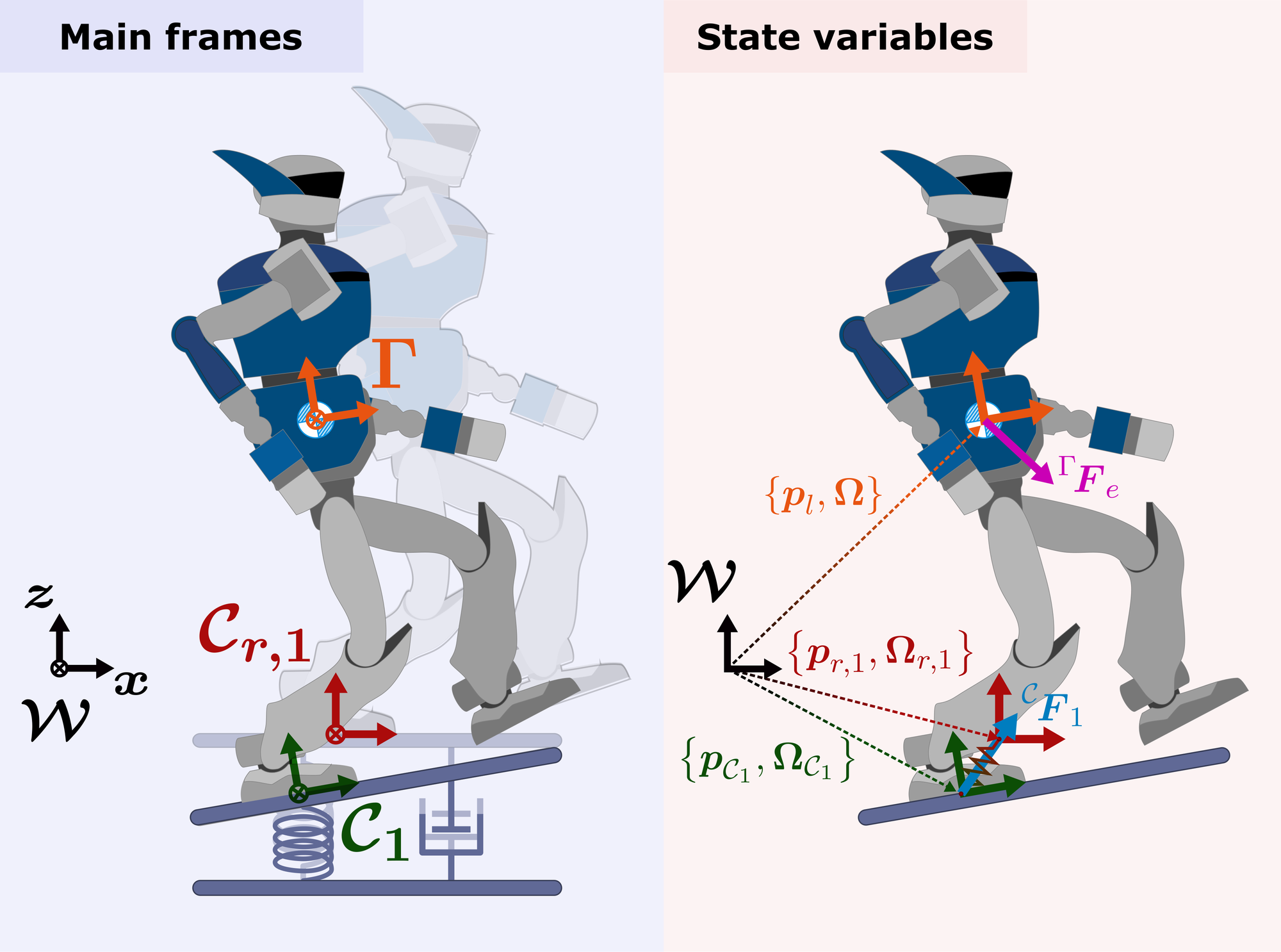}
   \caption{ \textbf{Left}: Main frames introduced in the Kinetics Observer. $\mathcal{W}$: \emph{world frame}, $\Gamma$: \emph{centroid frame}, $\mathcal{C}_{i}$: \emph{i-th contact's frame}, $\mathcal{C}_{r,i}$: \emph{i-th contact's rest frame}. The transparent robot corresponds to the state of the robot at the instant the contact is created, while the non-transparent one represents it after the deformation resulted from the contact. This is a virtual case with unrealistic and exaggerated deformations, in order to make the frames clearer. \textbf{Right}: Main state variables involved in the Kinetics Observer. For more clarity, the forces are represented using their notation in the state vector and the associated moments are omitted.}
   \label{fig:FramesAndVars}
   \end{figure}

\noindent In this section, we introduce the frames the Kinetics Observer relies on. To help visualize them, they are represented on Figure~\ref{fig:FramesAndVars} along the main state variables defined in Section~\ref{subsec:state-vec-def}.\\
We first introduce the \emph{centroid frame} denoted $\Gamma$, which relies on the centroid definition specific to humanoid robots~\citep{Orin2013Centroidal_Dynamics_Humanoid_Robot}. This frame is attached to the robot and is located at the center of mass. The conditions that its orientation must satisfy are that (i) its trajectory in the world frame is differentiable twice and (ii) knowing the kinematics (pose, velocities, and accelerations) of the centroid in the world frame allows to obtain the kinematics of all the robot limbs in the world frame using simple frame transformations. In this paper, we define this orientation as that of the floating base. Note that besides the rotations, the estimated kinematics of the centroid frame in the state vector are expressed in the centroid frame and not in the world frame.  The relation to their expression in the world frame is reminded in Table~\ref{tab:state-kinematics} in Appendix~\ref{sec:appendix-notations}. To make this specificity clearer throughout the paper, we add the subscript $l$ to these variables.

Each contact $i$ is associated with a \emph{contact frame} $\mathcal{C}_{i}$. This frame is attached to the robot and its kinematics $\left\{ \boldsymbol{p}_{\mathcal{C}_{i}},\boldsymbol{\Omega}_{\mathcal{C}_{i}},\dot{\boldsymbol{p}}_{\mathcal{C}_{i}},\boldsymbol{\omega}_{\mathcal{C}_{i}}\right\}$ in the world frame are obtained by forward kinematics from those of the centroid frame. A reaction wrench is applied to the contact. Its expression in $\mathcal{C}_{i}$ is denoted $\left\{^{\mathcal{C}}\boldsymbol{F}_{i},^{\mathcal{C}}\boldsymbol{T}_{i}\right\}$. A wrench sensor can be associated with the contact, providing a direct measurement of the reaction wrench.
As detailed in Section~\ref{subsec:Viscoelastic-model}, we use a model of viscoelastic contacts that additionally considers a \emph{rest frame} $\mathcal{C}_{r,i}$ for the contact. This frame is attached to the environment and denotes the pose of the contact such that the deformation would be null, similarly to the rest position of a linear spring. We can see it as the anchor of the robot in the environment, that describes the environment's local geometry. This pose can then vary in time to account for slippage. Figure~\ref{fig:FramesAndVars} helps to grasp the difference between $\mathcal{C}_{i}$ and $\mathcal{C}_{r,i}$.
To simplify the notation, we denote $\left\{\boldsymbol{p}_{r,i}\triangleq\boldsymbol{p}_{\mathcal{C}_{r,i}}, \boldsymbol{\Omega}_{r,i}\triangleq\boldsymbol{\Omega}_{\mathcal{C}_{r,i}}\right\}$ the \emph{rest pose} of the contact $i$ in the world frame.

\noindent Finally, we note that a summary of the notation is available in Appendix~\ref{sec:appendix-notations}.

\section{Model of the Kinetics Observer}\label{sec:ProblemStatement}

\noindent The Kinetics Observer estimates the extrinsic state of a legged robot in contact with the environment. Below are the system specifications and the requirements for the observer.
We consider that some of the contacts with the robot may be equipped with wrench sensors, and some may not. The robot may also be equipped with any number of IMUs, consisting of a 3D accelerometer and a 3D gyrometer. Their placement on the robot is not constrained. The sensors may operate at a lower frequency than the control loop, and thus may not provide a value at each iteration. Also for each iteration, we consider:
\begin{itemize}[leftmargin=1em]
    \item $n_{c}$ the number of current contacts.
    \item $n_{w}$ the number of available contact wrench signals.
    \item $n_{I}$ the number of available IMU signals.
\end{itemize}
\subsection{State-Space Representation of the System} \label{subsec:state-vec-def}
\subsubsection{Definition of the State Vector}

\noindent The observer is designed to estimate the following state components: 
\begin{itemize}[leftmargin=1em]
\item The kinematics of the centroid frame in the world frame $\left\{ \boldsymbol{p}_{l}, \boldsymbol{\Omega}, \boldsymbol{v}_{l}, \boldsymbol{\omega}_{l} \right\}$ together with the predictions of the linear and angular acceleration, $\boldsymbol{a}_{l}$ and $\dot{\boldsymbol{\omega}}_{l}$, respectively.
\item The bias $\boldsymbol{b}_{g,j}$ that alters the signals of each gyrometer.
\item The resulting disturbance forces and moments $\left\{ ^{\Gamma}\boldsymbol{F}_{e},{}^{\Gamma}\boldsymbol{T}_{e}\right\}$, i.e. the wrench applied to the robot that is not associated with our model of contacts and not measured by sensors.
\item A contact state $\boldsymbol{x}_{c,i}\triangleq\left\{ \boldsymbol{p}_{r,i},\boldsymbol{\Omega}_{r,i},{}^{\mathcal{C}}\boldsymbol{F}_{i},{}^{\mathcal{C}}\boldsymbol{T}_{i}\right\}$ for each of the current $n_{c}$ contacts. It contains the pose of the contact rest frame $\mathcal{C}_{r,i}$ in the world frame and the contact wrench expressed in the contact frame $\mathcal{C}_{i}$. 
\end{itemize}

\noindent Our state vector is thus defined as follows:
\begin{equation}
\boldsymbol{x}\triangleq\left(\boldsymbol{p}_{l},\boldsymbol{\Omega},\boldsymbol{v}_{l},\boldsymbol{\omega}_{l},\left\{ \boldsymbol{b}_{g,j}\right\} _{j=0}^{n_{I}},{}^{\Gamma}\boldsymbol{F}_{e},{}^{\Gamma}\boldsymbol{T}_{e},\left\{ \boldsymbol{x}_{c,i}\right\} _{i=0}^{n_{c}}\right) \text{ .} \label{eq:vector_x}
\end{equation}
It includes components from $\mathbb{R}^{3}$ and components of $SE\left(3\right)$, defining a Lie group as the state space. Note that our state vector has a dynamic size that changes as the robot creates or breaks contacts with the environment.

\subsubsection{Definition of the Output}
The measurements vector, defined in $\mathbb{R}^{6\left(n_{I}+n_{w}\right)}$, is the following:
\begin{equation}
\boldsymbol{y}\triangleq\left(\left\{ \boldsymbol{y}_{a,j},\boldsymbol{y}_{g,j}\right\} _{j=0}^{n_{I}},\left\{ \boldsymbol{y}_{F,i},\boldsymbol{y}_{T,i}\right\} _{i=0}^{n_{w}}\right) \text{ ,}
\end{equation}
with $\boldsymbol{y}_{a,j}$ and $\boldsymbol{y}_{g,j}$ the accelerometer and gyrometer measurements of the $j$-th IMU, respectively. Similarly, $\boldsymbol{y}_{F,i}$ and $\boldsymbol{y}_{T,i}$ are the force and moment measurement of the $i$-th contact's wrench sensor. Since the number of IMU signals $n_{I}$ and wrench sensor signals $n_{w}$ can vary over time, the size of the measurement vector will evolve accordingly.

\subsubsection{Definition of the Inputs}
The input vector is composed of dynamic variables of the system and information about the considered contacts and IMUs:
\begin{equation}
\boldsymbol{u}\triangleq\left(^{\Gamma}\boldsymbol{I},{}^{\Gamma}\dot{\boldsymbol{I}},{}^{\Gamma}\boldsymbol{\sigma},{}^{\Gamma}\dot{\boldsymbol{\sigma}},\left\{ \boldsymbol{\Xi}_{i}\right\} _{i=0}^{n_{c}},\left\{ \boldsymbol{\Psi}_{j}\right\} _{j=0}^{n_{I}}, {}^{\Gamma}\boldsymbol{F}_{res}, {}^{\Gamma}\boldsymbol{T}_{res}\right) \text{,} \label{eq:vector_u}
\end{equation}
with $^{\Gamma}\boldsymbol{I},{}^{\Gamma}\dot{\boldsymbol{I}},{}^{\Gamma}\boldsymbol{\sigma}$ and $^{\Gamma}\dot{\boldsymbol{\sigma}}$ the total inertia matrix, the total angular momentum of the robot expressed at its centroid, and their derivatives. 

\noindent $\boldsymbol{\Xi}_{i}\triangleq\left\{ {}^{\Gamma}\boldsymbol{p}_{\mathcal{C}_{i}},{}^{\Gamma}\boldsymbol{\Omega}_{\mathcal{C}_{i}},{}^{\Gamma}\dot{\boldsymbol{p}}_{\mathcal{C}_{i}},{}^{\Gamma}\boldsymbol{\omega}_{\mathcal{C}_{i}}, \check{\boldsymbol{p}}_{r,i},\check{\boldsymbol{\Omega}}_{r,i}\right\}$ corresponds to the input variables associated with each contact $i$.
$\left\{ ^{\Gamma}\boldsymbol{p}_{\mathcal{C}_{i}},{}^{\Gamma}\boldsymbol{\Omega}_{\mathcal{C}_{i}},{}^{\Gamma}\dot{\boldsymbol{p}}_{\mathcal{C}_{i}},{}^{\Gamma}\boldsymbol{\omega}_{\mathcal{C}_{i}}\right\}$ are the kinematics of the contact's frame in the centroid frame and $\left\{ \check{\boldsymbol{p}}_{r,i},\check{\boldsymbol{\Omega}}_{r,i}\right\} $ is the initial value of its rest pose in the world frame at the time it is created. This value is used to initialize the contact in the state vector and is thus given only upon its creation.

\noindent $\boldsymbol{\Psi}_{j}\triangleq\left\{ ^{\Gamma}\boldsymbol{p}_{\mathcal{S}_{j}},{}^{\Gamma}\boldsymbol{\Omega}_{\mathcal{S}_{j}},{}^{\Gamma}\dot{\boldsymbol{p}}_{\mathcal{S}_{j}},{}^{\Gamma}\boldsymbol{\omega}_{\mathcal{S}_{j}},{}^{\Gamma}\ddot{\boldsymbol{p}}_{\mathcal{S}_{j}}\right\} $ are the input variables associated with each IMU $j$. They correspond respectively to the pose, the linear and angular velocities, and the linear acceleration of the IMU's frame $\mathcal{S}_{j}$ in the centroid frame. 
This acceleration depends on the joint accelerations, which are not measurable, it is computed from the reference joint accelerations used to control the robot. If not available, one can consider it zero, which is generally a reasonable approximation if the IMU is placed in the torso of the robot.
$\left\{^{\Gamma}\boldsymbol{F}_{res}, ^{\Gamma}\boldsymbol{T}_{res}\right\}$ is the resultant wrench measured by all the sensors that are not associated with currently set contacts. This allows us to account for all the measured external wrenches, ensuring the continuity of the total wrench exerted on the robot. 

\subsection{State-Transition Model}\label{subsec:State-transition-model}
\subsubsection{Kinematics of the centroid frame}\label{subsubsec:centroid-kinematics-transi}
The state variables are dynamically linked to each other. Their relations are modeled with a discrete-time state-transition function that allows to predict the future state $\boldsymbol{x}_{k+1}$ of the system based on the current state $\boldsymbol{x}_{k}$ and the system inputs $\boldsymbol{u}_{k}$. In other words,
\begin{equation}
    \boldsymbol{x}_{k+1}=f\left(\boldsymbol{x}_{k},\boldsymbol{u}_{k}\right)\text{ .}\label{eq:raw-state-transition}
\end{equation}
\noindent The prediction of the centroid frame kinematics is done by the discrete integration of the current ones: 
\begin{equation}
    \left\{
    \begin{aligned}
        \boldsymbol{p}_{l,k+1}  \! & = \text{Exp} \! \left(\boldsymbol{\theta}_k\right)^{\top} \boldsymbol{p}_{l,k}
    + \boldsymbol{V}_{\boldsymbol{\theta}_{k}}^{\top} . \! \left(\delta_{T}\boldsymbol{v}_{l,k}+\frac{\delta_{T}^{2}}{2}\boldsymbol{a}_{l,k} \right)  \text{,}  \\
     \boldsymbol{R}_{k+1} \! & = \boldsymbol{R}_{k} \text{Exp} \! \left(\boldsymbol{\theta}_{k} \right)\text{,}   \\   
     \boldsymbol{v}_{l,k+1}   \!   & = \boldsymbol{v}_{l,k}+\delta_{T}\left(-\left[\boldsymbol{\omega}_{l,k}\right]_{\times}\boldsymbol{v}_{l,k}+\boldsymbol{a}_{l,k}\right)\text{,}   \\  
     \boldsymbol{\omega}_{l,k+1} \! & = \boldsymbol{\omega}_{l,k}+\delta_{T}\dot{\boldsymbol{\omega}}_{l,k}\text{,} 
    \end{aligned}
    \right.
    \label{eq:state-kinematics-pred}
\end{equation}

\noindent with $\boldsymbol{\theta}_{k} = \delta_{T}\boldsymbol{\omega}_{l,k} + \frac{\delta_{T}^{2}}{2}\dot{\boldsymbol{\omega}}_{l,k}$.
The matrix $\boldsymbol{V}_{\boldsymbol{\theta}}$ is the one defined in the exponential map of the SE$\left(3\right)$ Lie group~\citep{Blanco2021_SE3_Transformation_Parametrizations}.
\begin{equation}
    \boldsymbol{V}_{\boldsymbol{\theta}} = \mathbb{I} +\frac{\left(1-\cos(\Vert\boldsymbol{\theta}\Vert)\right)}{\Vert\boldsymbol{\theta}\Vert^{2}} \left[\boldsymbol{\theta}\right]_{\times} + \frac{\Vert\boldsymbol{\theta}\Vert-\text{sin}\left(\Vert\boldsymbol{\theta}\Vert\right)}{\Vert\boldsymbol{\theta}\Vert^{3}}\left[\boldsymbol{\theta}\right]_{\times}^{2}\text{.} 
\end{equation}
This term allows to integrate the linear velocity expressed in the centroid frame into the position while taking into account the evolution of the orientation over the integration interval.\\
Eq.~\eqref{eq:state-kinematics-pred} integrates the linear and angular accelerations of the centroid frame in the world frame, $\boldsymbol{a}_{l,k}=f_{a}\left(\boldsymbol{x}_{k},\boldsymbol{u}_{k}\right)$ and $\dot{\boldsymbol{\omega}}_{l,k}=f_{\dot{\omega}_{l}}\left(\boldsymbol{x}_{k},\boldsymbol{u}_{k}\right)$ respectively, which are considered constant over the integration interval $\delta_{T}$. They are obtained from Newton-Euler's equations considering the robot as a rigid body:
\begin{align}
\boldsymbol{a}_{l} = & \ \frac{{{}^{\Gamma}\boldsymbol{F}_{res}+ {}^{\Gamma}}\boldsymbol{F}_{e}+\sum_{i=0}^{n_{c}}{}^{\Gamma}\boldsymbol{R}_{\mathcal{C}_{i}}{}^{\mathcal{C}}\boldsymbol{F}_{i}}{m}-g_{0}\boldsymbol{R}^{\top}\boldsymbol{e}_{z}\text{ ,}\label{eq:newtonLinAcc} \\
\dot{\boldsymbol{\omega}}_{l} = & \ {}^{\Gamma}\boldsymbol{I}^{-1}\left(^{\Gamma}\boldsymbol{T}_{res} + ^{\Gamma}\boldsymbol{T}_{c}+{}^{\Gamma}\boldsymbol{T}_{e}\right) \label{eq:eulerAngAcc}\\
 & \ - {}^{\Gamma}\boldsymbol{I}^{-1}\left({}^{\Gamma}\dot{\boldsymbol{I}}\boldsymbol{\omega}_{l}+{}^{\Gamma}\dot{\boldsymbol{\sigma}} + \left[\boldsymbol{\omega}_{l}\right]_{\times}\!\left(^{\Gamma}\boldsymbol{I}\boldsymbol{\omega}_{l}+{}^{\Gamma}\boldsymbol{\sigma}\right)\right)\text{ ,}\nonumber
\end{align}
where $^{\Gamma}\boldsymbol{T}_{c}$ is the total contact moment, expressed in the centroid frame and defined by
\begin{equation}
^{\Gamma}\boldsymbol{T}_{c}=\sum_{i=0}^{n_{c}}\left(^{\Gamma}\boldsymbol{R}_{\mathcal{C}_{i}}{}^{\mathcal{\mathcal{C}}}\boldsymbol{T}_{i}+ \left[^{\Gamma}\boldsymbol{p}_{\mathcal{C}_{i}}\right]_{\times}\!{}^{\Gamma}\boldsymbol{R}_{\mathcal{C}_{i}}{}^{\mathcal{C}}\boldsymbol{F}_{i}\right)\text{ .}
\end{equation}

\noindent $m$ is the total mass of the robot and $g_{0}$ is the gravitational acceleration constant. The remaining variables are state and input variables, defined in Eq.~\eqref{eq:vector_x} and Eq.~\eqref{eq:vector_u}.
In particular, we note that the accelerations $\boldsymbol{a}_{l}$ and $\dot{\boldsymbol{\omega}}_{l}$ are functions of the wrenches applied to the robot. By integrating them to predict the centroid frame kinematics, we tighten the coupling between the kinematics and the dynamics of our system.

\subsubsection{Gyrometer bias, disturbance wrench and contacts rest pose} \label{subsec:gyro-extWrench-restPose-process}
\noindent The bias on each gyrometer, the disturbance wrench applied to the robot, and the rest poses of the contacts, are assumed unpredictable and to be subject to slow variations. Therefore, they are considered constant in the state-transition model.

\subsubsection{Viscoelastic model of the contacts}\label{subsec:Viscoelastic-model}
\noindent We cannot rely solely on the force measurements to predict the acceleration of our system. This is firstly because they may not be available, but secondly and more importantly, relying only on forces to predict the accelerations leads to unbounded drifts in the absolute position. This is due to the uncertainties in the sensor measurements and the robot models. Nevertheless, we know that the robot has relatively reliable anchors in the environment: the contacts. However, enforcing this anchor kinematically would conflict with the Newton-Euler dynamics.

To maintain model consistency, drift correction must be applied through a wrench at the contact. To this end, we need a differentiable model that relates the kinematics of the robot to the contact reaction wrench~\citep{Mifsud2015EstimationContactForcesAndKinematicsWithOnlyImu,varin2020constrainedKfFrictContact}. In this paper, we use a viscoelastic contact model that penalizes the robot's drift from its contact points proportionally. Such a penalization model has proven effective for both simulation of biped walking (e.g.~\citet{yamane2006stablePenaltyBasedModelContact, rengifo2009penaltyBasedApproachContactForce}) and state estimation (e.g.~\citet{Benallegue2018ExtForceEstimationWithNoTorqueMeas}), due to its simplicity and ability to maintain smooth dynamics. Another advantage is that it gives a simple model of the contact compliance, including both the compliance of the ground and that of the robot. Indeed, legged robots generally feature flexibilities in their kinematic chain~\citep{wisth2020preintegratedVelocityBias, Mikami2014IdentificationHRP2Dynamics}. Their deformation under load can thus be modeled by viscoelastic models, in order to approximate it and compensate for it in controllers (e.g.~\citet{Romualdi2022WholeBodyControlEstimation,hamze2024compliantControlLegged}) and estimators (e.g.~\citet{Flayols2017ExperimentalEvaluationOfSimpleEstimForHumanoidRobots}).  

\begin{figure}[!t]
    \centering
    \includegraphics[width=1\columnwidth]{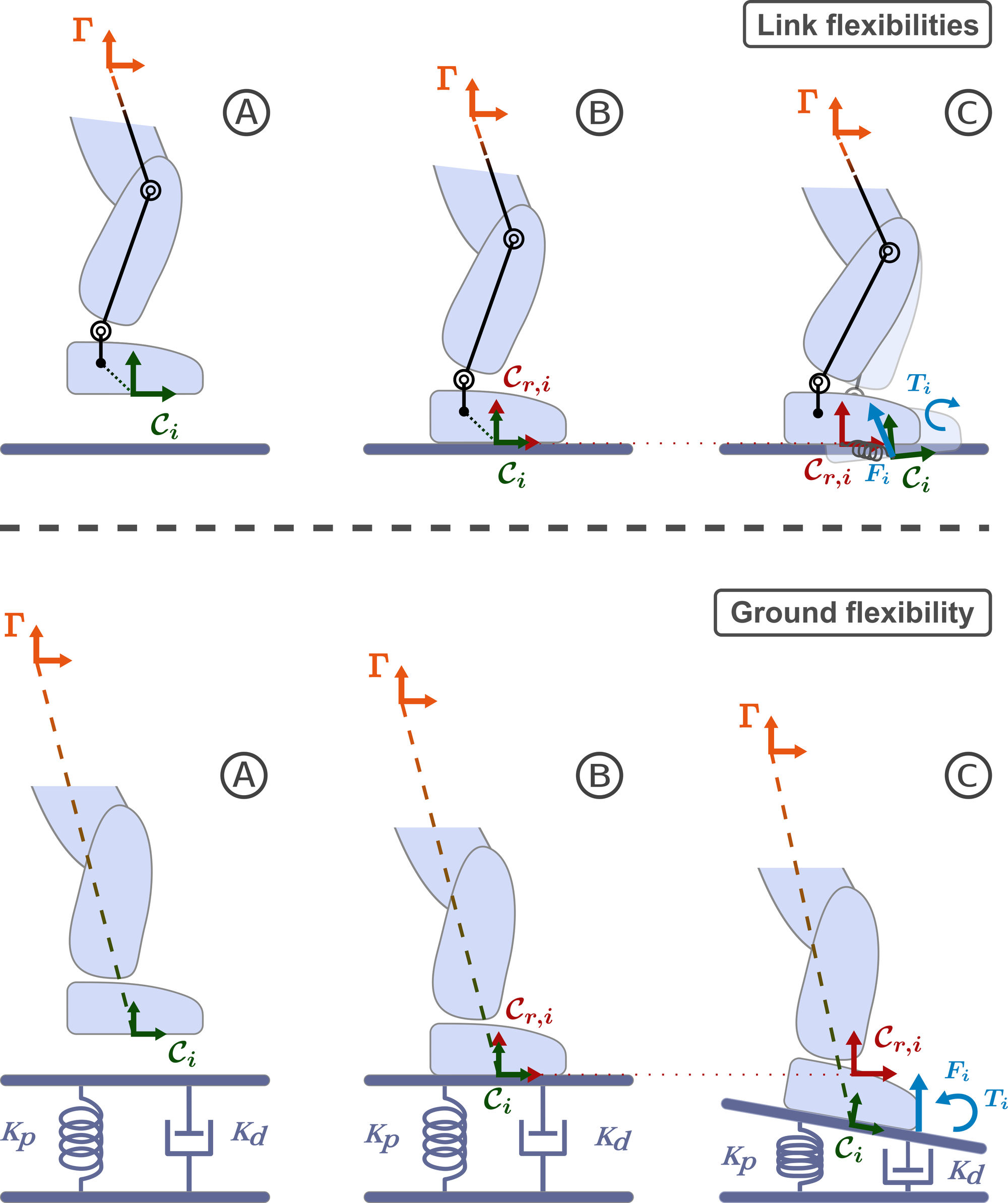}
    \caption{Viscoelastic model of contacts applied to the robot's foot at three stages. \textbf{(A)} Foot before the contact with the ground. \textbf{(B)} Creation of the contact. No force is applied, the contact frame matches the rest frame. \textbf{(C)} Generation of a reaction wrench due to the deformation. \textbf{Top:} perfectly rigid environment and flexible links, represented by bushings (6D deformable elements) \citep{Franco2023ModelingRigidLinkCompliant}. \textbf{Bottom:} perfectly rigid link flexibilities and deformable ground. The actual deformed leg is shown opaque, while the leg configuration assumed by the geometric model and joint encoders (forward kinematics) is shown transparent. Dashed lines symbolize perfectly known rigid transformations.}
    \label{fig:Viscoelastic-model} 
\end{figure}

In our model, each contact is described by two frames: the \emph{contact frame} $\mathcal{C}_{i}$, and the \emph{rest frame} $\mathcal{C}_{r,i}$. The contact frame is the frame \emph{attached to the limb} of the robot, which corresponds to the kinematics of the contact as known by the robot: its kinematics are obtained by forward kinematics from the estimated kinematics of the centroid. 
The rest frame is the frame \emph{attached to the environment}, representing its pose at the exact moment the contact is established, where it coincides with the contact frame. 
As soon as the contact material deforms, the contact frame will \textquotedbl follow\textquotedbl{} the robot's limb while the rest frame will remain fixed. 
We model the flexibility of the contact material with a {spring-damper} system between the current contact frame and the rest frame. The reaction wrench can thus be obtained from the relative errors between their kinematics. Figure~\ref{fig:Viscoelastic-model} illustrates this concept in two simplified cases where either the robot's leg or the ground is perfectly rigid.

It is important to note that with the visco-elastic model, the contact reaction force can be estimated solely from the estimated kinematics of both the contact's frames. The kinematics $\left\{ \boldsymbol{p}_{r,i},\boldsymbol{R}_{r,i}, \boldsymbol{0}, \boldsymbol{0}\right\} $ of the rest frame in the world are composed of its pose, which is part of our state vector, and of zero linear and angular velocities. The kinematics $\left\{ \boldsymbol{p}_{\mathcal{C}_{i}},\boldsymbol{R}_{\mathcal{C}_{i}},\dot{\boldsymbol{p}}_{\mathcal{C}_{i}},\boldsymbol{\omega}_{\mathcal{C}_{i}}\right\} $ of the contact frame are obtained by forward kinematics from the state kinematics of the centroid frame in the world and the input kinematics of the contact frame in the centroid frame:
\begin{equation}
    \left\{
    \begin{aligned}
        \boldsymbol{p}_{\mathcal{C}_{i}} &= \boldsymbol{R} \left(^{\Gamma}\boldsymbol{p}_{\mathcal{C}_{i}}+\boldsymbol{p}_{l}\right) \text{ ,} \\
        \boldsymbol{R}_{\mathcal{C}_{i}} &= \boldsymbol{R} \ {}^{\Gamma}\boldsymbol{R}_{\mathcal{C}_{i}} \text{ ,}  \\   
        \dot{\boldsymbol{p}}_{\mathcal{C}_{i}} &= \boldsymbol{R}\left(^{\Gamma}\dot{\boldsymbol{p}}_{\mathcal{C}_{i}}+ \left[\boldsymbol{\omega}_{l}\right]_{\times}\!{}^{\Gamma}\boldsymbol{p}_{\mathcal{C}_{i}}+\boldsymbol{v}_{l}\right)  \text{ ,} \\  
        \boldsymbol{\omega}_{\mathcal{C}_{i}} &= \boldsymbol{R}\left(^{\Gamma}\boldsymbol{\omega}_{\mathcal{C}_{i}}+\boldsymbol{\omega}_{l}\right) \text{ .}
    \end{aligned}
    \right.
\end{equation}
We can then compute the relative errors between the kinematics of the contact frame and that of the rest frame. We denote them $\left\{ \tilde{\boldsymbol{p}}_{i},\tilde{\boldsymbol{v}}_{i}, \tilde{\boldsymbol{R}}_{i},\tilde{\boldsymbol{\omega}}_{i}\right\} $:
\begin{equation}
    \begin{array}{c c}
        \left\{
        \begin{aligned}
            \tilde{\boldsymbol{p}}_{i} & =\boldsymbol{R}_{\mathcal{C}_{i}}^{\top}\left(\boldsymbol{p}_{\mathcal{C}_{i}}-\boldsymbol{p}_{r,i}\right)  \text{ ,}\\
            \tilde{\boldsymbol{v}}_{i} & =\boldsymbol{R}_{\mathcal{C}_{i}}^{\top}\dot{\boldsymbol{p}}_{\mathcal{C}_{i}}  \text{ ,}
        \end{aligned}
        \right.
        &
        \left\{
        \begin{aligned}
            \tilde{\boldsymbol{R}}_{i} & = \boldsymbol{R}_{r,i}^{\top} \boldsymbol{R}_{\mathcal{C}_{i}}   \text{ ,} \\
            \tilde{\boldsymbol{\omega}}_{i} & = \boldsymbol{R}_{\mathcal{C}_{i}}^{\top} \boldsymbol{\omega}_{\mathcal{C}_{i}} \text{ .}
        \end{aligned}
        \right.
    \end{array}
\end{equation}
Using our viscoelastic model, we finally obtain the reaction wrench at the contact:
\begin{alignat}{2}
 & ^{\mathcal{C}}\boldsymbol{F}_{i} && =-\boldsymbol{K}_{p,t}\tilde{\boldsymbol{p}}_{i}-\boldsymbol{K}_{d,t}\tilde{\boldsymbol{v}}_{i} \text{ ,} \label{eq:ViscoElasticForce}\\
 & ^{\mathcal{C}}\boldsymbol{T}_{i} && =-\frac{1}{2}\boldsymbol{K}_{p,r} \left[\tilde{\boldsymbol{R}}_{i}-\tilde{\boldsymbol{R}}_{i}^{\top}\right]_{\vee} \! -\boldsymbol{K}_{d,r}\tilde{\boldsymbol{\omega}}_{i} \text{ .} \label{eq:ViscoElasticTorque}
\end{alignat}
where $\boldsymbol{K}_{p,t}$ and $\boldsymbol{K}_{d,t}$ are the $3\times3$ SPD matrices corresponding to the linear stiffness and damping of the contact. Similarly, $\boldsymbol{K}_{p,r}$ and $\boldsymbol{K}_{d,r}$ correspond to its angular stiffness and damping. The tuning of these flexibilities is explained in Appendix~\ref{subsec:tuning-of-KO}. 
Note that the expression of the contact moment relies on the property that for a rotation matrix $\boldsymbol{R}$ we have 
\begin{equation}
\frac{1}{2} \left[\boldsymbol{R}-\boldsymbol{R}^{\top}\right]_{\vee} =\frac{\sin\left(\left\Vert \text{Log}\left(\boldsymbol{R}\right)\right\Vert \right)}{\left\Vert \text{Log}\left(\boldsymbol{R}\right)\right\Vert }\text{Log}\left(\boldsymbol{R}\right)\text{ ,}\label{eq:r-rt-property}
\end{equation}
which approximates the rotation vector equivalent to $\boldsymbol{R}$ for small rotation angles.

In summary, the viscoelastic model of contacts  has the following strengths: 
\begin{itemize}[leftmargin=1em]
\item It prevents divergence of the centroid frame from the contacts by associating their difference with a proportional reaction. 
\item It enables wrench sensor measurements to actively correct both the centroid frame kinematics and the contact rest pose. This correction is distributed according to consistency with the other available information, such as the IMU measurements.
As proven in Section~\ref{subsec:observability}, this makes the rest pose observable relative to the centroid frame, thereby making slippage observable relative to this frame and enabling estimation of the local geometry of the environment at the contact. 
\item Our method explicitly models the contact material's compliance, to account for its impact on forward kinematics and mitigate inconsistencies between the leg kinematic and the IMU measurements.
\item In the absence of wrench sensors on the contacts, it allows to estimate the reaction wrench based on the kinematics. In that case, the performed odometry would behave similarly to a standard Inertial-Leg odometry.
\item The formulation accounts for both contact position and orientation. Nevertheless, point contacts can still be represented by simply setting the angular stiffness $\boldsymbol{K}_{p,r}$ and damping $\boldsymbol{K}_{d,r}$ to zero.
\end{itemize}

Finally, although this model seems highly dependent on accurate tuning of the contact flexibilities, we explain in Appendix~\ref{subsec:tuning-of-KO} that the estimator is in fact robust to approximations in these parameters, and we demonstrate this experimentally for errors of up to one order of magnitude.

\subsection{Observation Model}\label{subsec:observationModel}

\noindent The observation model allows to predict the measurement vector from the current state and input vectors:
\begin{equation}
\boldsymbol{y}_{k}=g\left(\boldsymbol{x}_{k},\boldsymbol{u}_{k}\right)\text{ .}\label{eq:raw-meas-model}
\end{equation}

\noindent The estimated measurements of the wrench sensor at each contact $i$ directly correspond to the contact wrench in the state vector:
\begin{align}
\boldsymbol{y}_{F,i} & ={}^{\mathcal{C}}\boldsymbol{F}_{i}\text{ ,}\label{eq:measModelForce} \\
\boldsymbol{y}_{T,i} & ={}^{\mathcal{\mathcal{C}}}\boldsymbol{T}_{i}\text{ .}\label{eq:measModelTorque}
\end{align}
We obtain the gyrometer and the accelerometer measurements of each IMU $j$ using forward kinematics from the input and state variables:
\begin{align}
\boldsymbol{y}_{g,j} = & \: {^{\Gamma}}\boldsymbol{R}_{\mathcal{S}_{j}}^{\top}\left(\boldsymbol{\omega}_{l} + {}^{\Gamma}\boldsymbol{\omega}_{\mathcal{S}_{j}}\right)+\boldsymbol{b}_{g,j}\text{ ,}\label{eq:measModelGyro} \\
\boldsymbol{y}_{a,j} = & \: {^{\Gamma}}\boldsymbol{R}_{\mathcal{S}_{j}}^{\top}\left(\boldsymbol{a}_{l}+{}^{\Gamma}\ddot{\boldsymbol{p}}_{\mathcal{S}_{j}} + g_{0}\boldsymbol{R}^{\top}\boldsymbol{e}_{z} \right) \label{eq:measModelAccelero}\\
 & + {^{\Gamma}}\boldsymbol{R}_{\mathcal{S}_{j}}^{\top}\left(\left( \left[\boldsymbol{\omega}_{l}\right]_{\times}^{2} + \left[\dot{\boldsymbol{\omega}}_{l}\right]_{\times} \right){}^{\Gamma}\boldsymbol{p}_{\mathcal{S}_{j}}+2\left[\boldsymbol{\omega}_{l}\right]_{\times} \! {}^{\Gamma}\dot{\boldsymbol{p}}_{\mathcal{S}_{j}}\right) \text{ .}\nonumber 
\end{align}

\noindent In particular, $\boldsymbol{a}_{l}$ and $\dot{\boldsymbol{\omega}}_{l}$, the linear and angular accelerations of the centroid frame, are obtained from \eqref{eq:newtonLinAcc} and \eqref{eq:eulerAngAcc}, respectively. Based on these relations, the prediction of the accelerometer measurement is therefore a function of the total wrench exerted on the robot, which is part of the state vector. The accelerometer is thus comparable to an additional wrench sensor, which embraces the principle of estimating the \emph{kinetics} of the robot.

\subsection{Observability Analysis}\label{subsec:observability}
\noindent In this section we provide an observability analysis of the proposed model, inspired by~\citep{mifsud2017These}. Our analysis focuses on the full system leveraging all sensor measurements. For the sake of the analysis, we need to consider the hypothesis that gyrometer measurements are unbiased.
\paragraph{From IMU measurements}
From \eqref{eq:measModelGyro}, we notice that the gyrometer measurement depends solely on input variables and on the angular velocity of the centroid frame $\boldsymbol{\omega}_{l}$. This measurement thus gives us the observability of $\boldsymbol{\omega}_{l}$ and of its derivative $\dot{\boldsymbol{\omega}}_{l}$. Following the same principle, based on \eqref{eq:measModelAccelero}, the accelerometer measurement gives the observability of the term $\boldsymbol{a}_{l} + g_{0}\boldsymbol{R}^{\top}\boldsymbol{e}_{z}$. From \eqref{eq:newtonLinAcc} and \eqref{eq:eulerAngAcc}, we conclude that the IMU measurements provide the observability of the total wrench applied to the robot, whatever the number of contacts.
\paragraph{Adding wrench sensor measurements}
Using the previous result, we can directly conclude on the observability of the disturbance wrench $\left\{ ^{\Gamma}\boldsymbol{F}_{e},{}^{\Gamma}\boldsymbol{T}_{e}\right\}$.
We now use the Lemma~\ref{lemma:observability}, which provides the condition of the observability of the position and velocity errors $\tilde{\boldsymbol{p}}_{i}$ and $\tilde{\boldsymbol{v}}_{i}$, and of the robot's tilt $\boldsymbol{R}^{\top}\boldsymbol{e}_{z}$. Based on the conclusions above, $\boldsymbol{a}_{l}$ the centroid frame's linear acceleration is also observable.

 Noticing that $^{\Gamma}\boldsymbol{p}_{r,i}$ the rest position of the i-th contact in the centroid frame and $\boldsymbol{v}_{l}$ the velocity of the centroid frame in the world frame, can be expressed as:
\begin{align}
    ^{\Gamma}\boldsymbol{p}_{r,i} & ={}^{\Gamma}\boldsymbol{p}_{\mathcal{C}_{i}}-{}^{\Gamma}\boldsymbol{R}_{\mathcal{C}_{i}}\tilde{\boldsymbol{p}}_{i} \text{ ,} \label{eq:obs-pri}\\
    \boldsymbol{v}_{l} & =\tilde{\boldsymbol{v}}_{i}-\left({}^{\Gamma}\dot{\boldsymbol{p}}_{\mathcal{C}_{i}} + \left[\boldsymbol{\omega}_{l}\right]_{\times}{}^{\Gamma}\boldsymbol{p}_{\mathcal{C}_{i}}\right) \text{ ,}
\end{align}
we can conclude that they rely solely on observable variables and are thus observable. 

\noindent From \eqref{eq:ViscoElasticTorque}, after removing the damping part since it is clearly observable, we can write:
\begin{equation}
    \left[\tilde{\boldsymbol{R}}_{i}-\tilde{\boldsymbol{R}}_{i}^{\top}\right]_{\vee} = -2{\boldsymbol{K}_{p,r}^{-1}}{^{\mathcal{C}}\boldsymbol{T}_{i}} \text{ ,}
\end{equation}
with $\tilde{\boldsymbol{R}}_{i}$ the orientation error due to the deformation of the contact material. Noting that the right part of this equation is observable and using~\eqref{eq:r-rt-property}, we can conclude on the observability of the unit rotation vector corresponding to $\tilde{\boldsymbol{R}}_{i}$. With the hypothesis that the norm of the orientation error is lower than $\pi/2$, this norm is also determinable without ambiguity and $\tilde{\boldsymbol{R}}_{i}$ is thus observable. This is not a strong hypothesis since this error represents the angular deformation of the robot and environment materials in contact. 
The rest orientation $^{\Gamma}\boldsymbol{R}_{r,i}$ of the $i$-th contact frame in the centroid frame can be expressed by $^{\Gamma}\boldsymbol{R}_{r,i}={}^{\Gamma}\boldsymbol{R}_{\mathcal{C}_{i}}\tilde{\boldsymbol{R}}_{i}^{\top}$, with ${}^{\Gamma}\boldsymbol{R}_{\mathcal{C}_{i}}$ available in the input. We conclude that $^{\Gamma}\boldsymbol{R}_{r,i}$ is observable. 
Consequently, the inclination of the contact frame in the world frame, $\boldsymbol{R}_{r,i}^{\top}\boldsymbol{e}_{z} = {}^{\Gamma}\boldsymbol{R}_{r,i}^{\top}\boldsymbol{R}^{\top}\boldsymbol{e}_{z}$, is also observable.

We can conclude that the viscoelastic model of contacts provides the observability of contact rest poses, and thus of contact slippage with respect to the centroid frame, both in translation and in rotation.

Pure proprioception does not provide observability of the robot's position and yaw orientation in the world. We conclude this observability analysis by showing that they would be observable, under the additional hypotheses that the weighted averages \eqref{eq:loadweights} of the contacts rest positions and of their yaw orientations in the world are known. 
 
For simplification purpose, we make the assumption that the yaw corresponds to the vector $\boldsymbol{R}^{\top}\boldsymbol{e}_{x}$.
The average contact rest yaw orientation can thus be expressed as:
\begin{equation}
    \sum_{i}^{n_{w}}w_{i}\boldsymbol{R}_{r,i}^{\top}\boldsymbol{e}_{x}=\sum_{i}^{n_{w}}w_{i}{^{\Gamma}\boldsymbol{R}_{r,i}^{\top}}\boldsymbol{R}^{\top}\boldsymbol{e}_{x} \text{ .}
\end{equation}
Using the results and assumptions above, we thus have the observability of $\boldsymbol{R}^{\top}\boldsymbol{e}_{x}$. $\boldsymbol{R}$ can be reconstructed from $\boldsymbol{R}^{\top}\boldsymbol{e}_{x}$ and $\boldsymbol{R}^{\top}\boldsymbol{e}_{z}$, so it is observable.

\noindent Finally, we can write
\begin{equation}
    \boldsymbol{p} = \boldsymbol{R}\sum_{i}^{n_{w}}w_{i}{}^{\Gamma}\boldsymbol{p}_{r,i}-\sum_{i}^{n_{w}}w_{i}\boldsymbol{p}_{r,i} \text{ .}
\end{equation} 

Using the hypothesis that the weighted average of the contact rest positions is known, the centroid frame's position in the world is also observable. 

We can conclude that with the assumption of unbiased gyrometer measurements, the disturbance wrench and contact rest poses (thus contact slippage) with respect to the centroid frame are observable. By assuming known weighted averages of the contact frame rest positions and yaw angle, the full kinematics of the centroid frame in the world are observable.  
It is worth noting that these hypotheses would not be necessary anymore by adding exteroceptive measurements of the robot's pose, providing the observability of kinematics in the global frame.

\subsection{MEKF with Constrained Process Covariances}\label{sec:Implementation}
\noindent A Multiplicative Extended Kalman Filter (MEKF)~\citep{markley2004MEKF} was used to solve our estimation problem. Its structure, along with the main variables it relies on, are summarized in Figure~\ref{fig:KineticsObserver-structure}. 

In this section, we introduce a way to relax the constraint of fixed contacts in Kalman Filters for proprioceptive odometry, while preventing free drift. 

By appending the contact's reference pose in the state vector, it can be associated with an initial state covariance, allowing for its correction upon the contact creation. A process covariance on the model of fixed contacts allows for the long-term correction of their pose, mitigating potential slippage~\citep{bloesch2013FusionLegKineAndImu}. However, this also causes the robot to drift freely over time, undermining the primary anchor role of the contacts. For this reason, we introduce a specially constrained process noise covariance matrix: one for the sub-state of contact rest positions and another for the sub-state of contact rest orientations. These matrices are designed so that, at each iteration, the \emph{weighted average} of the process noise added across all contacts is zero for the relevant variables, each contact being weighted by the load it carries. By doing so, we continuously correct the contact's rest pose throughout its lifetime, while ensuring that the process noise does not cause this average rest pose to drift. We will refer to this feature as the \emph{constrained contact process covariances}. This constraint is applied to the non-observable components of the contact rest poses, namely their position and yaw angle.

Let us first consider the substate consisting only of the rest positions $\boldsymbol{X}$ of the $n$ current contacts. Its dynamics are:
\begin{equation}
  \left(\begin{array}{c} \boldsymbol{X}_{1}\\ \vdots\\ \boldsymbol{X}_{n} \end{array}\right)_{k+1}
  = \left(\begin{array}{c} \boldsymbol{X}_{1}\\ \vdots\\ \boldsymbol{X}_{n} \end{array}\right)_{k} + \boldsymbol{M}\boldsymbol{v} \text{ ,}
\end{equation}
with $\boldsymbol{M}\boldsymbol{v}$ the noise on our state-transition model and $n$ the number of contacts. $\boldsymbol{v} \in \mathbb{R}^{3n}$ is the process noise random variable. $\boldsymbol{M} \in \mathbb{R}^{3n\times 3n}$ is the matrix that maps this noise over the sub-state. We want a constraint that imposes a constant average position of the contacts in our state-transition model. We use a weighted average, in which each contact is weighted by the estimated force it carries: 
\begin{equation}
    w_i =
    \frac{\left\|{}^{\mathcal{C}}\boldsymbol{F}_i\right\| + \varepsilon}
    {\displaystyle\sum_{j=1}^{n}
    \left(\left\|{}^{\mathcal{C}}\boldsymbol{F}_j\right\|+\varepsilon\right)},
    \qquad
    \sum_{i=1}^{n} w_i = 1, 
    \label{eq:loadweights}
\end{equation}
where $\varepsilon>0$ is a small regularization term that prevents division by zero.
A contact carrying little load is either being created or removed, or is the most likely to slip: the weighting thus gives more influence to the contacts we can rely on.  

To define the constraint, we determine $\boldsymbol{M}$ such that the covariance of the weighted sum of this noise over each axis is zero. 
This can be written as
\begin{equation}
  \text{cov}\left(\boldsymbol{W}^{\top}\boldsymbol{M}\boldsymbol{v}\right)=\boldsymbol{0} \text{,} \label{eq:covstmv}
\end{equation}
where 
\begin{equation}
  \boldsymbol{W}^{\top} =\left(\begin{array}{ccc}
    w_{1}\mathbb{I}_{3\times3} & \cdots & w_{n}\mathbb{I}_{3\times3}
  \end{array}\right) \in \mathbb{R}^{3 \times 3n}
\end{equation}
is the matrix such that $\boldsymbol{W}^{\top}\boldsymbol{M}\boldsymbol{v} = \sum_{i=1}^{n}w_{i}\boldsymbol{M}_{i}\boldsymbol{v}$, with $\boldsymbol{M}_{i}$ the block of rows of $\boldsymbol{M}$ corresponding to the $i$-th contact.\\
We can also write
\begin{equation}\label{eq:cov_ST_Mv}
  \text{cov}\left(\boldsymbol{W}^{\top}\boldsymbol{M}\boldsymbol{v}\right)=\boldsymbol{W}^{\top}\text{cov}\left(\boldsymbol{M}\boldsymbol{v}\right)\boldsymbol{W} \text{ ,} 
\end{equation}
and 
\begin{equation} \label{eq:cov_Mv}
  \text{cov}\left(\boldsymbol{M}\boldsymbol{v}\right)=\boldsymbol{M}\text{cov}\left(\boldsymbol{v}\right)\boldsymbol{M}^{\top} \text{ .} 
\end{equation}
From \eqref{eq:covstmv}, \eqref{eq:cov_ST_Mv}, and \eqref{eq:cov_Mv}, we obtain
\begin{equation}
  \boldsymbol{W}^{\top}\boldsymbol{M}\text{cov}\left(\boldsymbol{v}\right)\boldsymbol{M}^{\top}\boldsymbol{W}=\boldsymbol{0} \text{ ,}
\end{equation}
which must be true for any vector $\boldsymbol{v}$. This is the case only if 
\begin{equation}
  \boldsymbol{W}^{\top}\boldsymbol{M}=\boldsymbol{0}. \label{eq:MW0}
\end{equation}
Using the fact that the weights sum up to one, we can define the matrix $\boldsymbol{S} =\left(\begin{array}{ccc} \mathbb{I}_{3\times3} & \cdots & \mathbb{I}_{3\times3}\end{array}\right)^{\top} \in \mathbb{R}^{3n \times 3}$, giving $\boldsymbol{W}^{\top}\boldsymbol{S}=\mathbb{I}_{3\times3}$. \\
We can  now express the matrix $\boldsymbol{M}$
\begin{equation}
  \boldsymbol{M}=\mathbb{I}_{3n\times3n}-\boldsymbol{S}\boldsymbol{W}^{\top} \text{ ,}
\end{equation}
which satisfies \eqref{eq:MW0}, and thus the condition \eqref{eq:covstmv}.
From \eqref{eq:cov_Mv}, we finally obtain the process noise covariance matrix on the state-transition model of our substate, which we use to replace the corresponding parts of the process noise covariance matrix. We then apply the same methodology to the rest yaw of the contacts, where the two first rows of the $\boldsymbol{W}^{\top}$ matrix, and the two first columns of $\boldsymbol{S}$, are zero.

We note that this feature is independent of the viscoelastic model of contacts and thus can be used in any Kalman Filter performing Leg odometry. Also, it allows us to constrain the Kalman Filter to restrict its correction to observable modes of the system, in a similar fashion to the Observability-Constrained Extended Kalman Filter (OC-EKF)~\citep{huang2010observabilityConstrainedEKF}. A core difference with this method is that rather than  modifying the EKF's structure and linearization point, we apply a constraint only to the propagation noise of key state variables.

We provide further explanation in Appendix~\ref{sec:appendix-implementation} on initializing contact rest poses and tuning the estimator parameters.

\section{Experimental Results}\label{sec:Experiments}

\newcommand{\getErrorResult}[5]{\csname#1#2#3#4#5\endcsname}
\newcommand{\FlatodometryHartleyVelerrorEstimateNormMeanabs}{0.026}
\newcommand{\FlatodometryHartleyVelerrorEstimateNormStd}{0.026}
\newcommand{\FlatodometryHartleyVelerrorEstimateXyMeanabs}{0.017}
\newcommand{\FlatodometryHartleyVelerrorEstimateXyStd}{0.019}
\newcommand{\FlatodometryHartleyVelerrorEstimateZMeanabs}{0.010}
\newcommand{\FlatodometryHartleyVelerrorEstimateZStd}{0.014}
\newcommand{\FlatodometryKowithoutwrenchsensorsVelerrorEstimateNormMeanabs}{0.029}
\newcommand{\FlatodometryKowithoutwrenchsensorsVelerrorEstimateNormStd}{0.031}
\newcommand{\FlatodometryKowithoutwrenchsensorsVelerrorEstimateXyMeanabs}{0.024}
\newcommand{\FlatodometryKowithoutwrenchsensorsVelerrorEstimateXyStd}{0.028}
\newcommand{\FlatodometryKowithoutwrenchsensorsVelerrorEstimateZMeanabs}{0.013}
\newcommand{\FlatodometryKowithoutwrenchsensorsVelerrorEstimateZStd}{0.018} 
\newcommand{\FlatodometryKoZpcVelerrorEstimateNormMeanabs}{0.028}
\newcommand{\FlatodometryKoZpcVelerrorEstimateNormStd}{0.029}
\newcommand{\FlatodometryKoZpcVelerrorEstimateXyMeanabs}{0.025}
\newcommand{\FlatodometryKoZpcVelerrorEstimateXyStd}{0.030}
\newcommand{\FlatodometryKoZpcVelerrorEstimateZMeanabs}{0.012}
\newcommand{\FlatodometryKoZpcVelerrorEstimateZStd}{0.016} 
\newcommand{\FlatodometryKineticsobserverVelerrorEstimateNormMeanabs}{0.027}
\newcommand{\FlatodometryKineticsobserverVelerrorEstimateNormStd}{0.030}
\newcommand{\FlatodometryKineticsobserverVelerrorEstimateXyMeanabs}{0.023}
\newcommand{\FlatodometryKineticsobserverVelerrorEstimateXyStd}{0.027}
\newcommand{\FlatodometryKineticsobserverVelerrorEstimateZMeanabs}{0.012}
\newcommand{\FlatodometryKineticsobserverVelerrorEstimateZStd}{0.016}
\newcommand{\FlatodometryHartleyRelerrorTiltMeanabs}{0.50}
\newcommand{\FlatodometryHartleyRelerrorTiltStd}{0.32}
\newcommand{\FlatodometryHartleyRelerrorTransxyMeanabs}{0.025}
\newcommand{\FlatodometryHartleyRelerrorTransxyStd}{0.017}
\newcommand{\FlatodometryHartleyRelerrorTranszMeanabs}{0.005}
\newcommand{\FlatodometryHartleyRelerrorTranszStd}{0.004}
\newcommand{\FlatodometryHartleyRelerrorYawMeanabs}{0.32}
\newcommand{\FlatodometryHartleyRelerrorYawStd}{0.28}
\newcommand{\FlatodometryKowithoutwrenchsensorsRelerrorTiltMeanabs}{0.50}
\newcommand{\FlatodometryKowithoutwrenchsensorsRelerrorTiltStd}{0.31}
\newcommand{\FlatodometryKowithoutwrenchsensorsRelerrorTransxyMeanabs}{0.018}
\newcommand{\FlatodometryKowithoutwrenchsensorsRelerrorTransxyStd}{0.014}
\newcommand{\FlatodometryKowithoutwrenchsensorsRelerrorTranszMeanabs}{0.004}
\newcommand{\FlatodometryKowithoutwrenchsensorsRelerrorTranszStd}{0.004}
\newcommand{\FlatodometryKowithoutwrenchsensorsRelerrorYawMeanabs}{0.35}
\newcommand{\FlatodometryKowithoutwrenchsensorsRelerrorYawStd}{0.28}
\newcommand{\FlatodometryKoZpcRelerrorTiltMeanabs}{0.51}
\newcommand{\FlatodometryKoZpcRelerrorTiltStd}{0.31}
\newcommand{\FlatodometryKoZpcRelerrorTransxyMeanabs}{0.023}
\newcommand{\FlatodometryKoZpcRelerrorTransxyStd}{0.015}
\newcommand{\FlatodometryKoZpcRelerrorTranszMeanabs}{0.006}
\newcommand{\FlatodometryKoZpcRelerrorTranszStd}{0.005}
\newcommand{\FlatodometryKoZpcRelerrorYawMeanabs}{0.32}
\newcommand{\FlatodometryKoZpcRelerrorYawStd}{0.26}
\newcommand{\FlatodometryKineticsobserverRelerrorTiltMeanabs}{0.50}
\newcommand{\FlatodometryKineticsobserverRelerrorTiltStd}{0.31}
\newcommand{\FlatodometryKineticsobserverRelerrorTransxyMeanabs}{0.015}
\newcommand{\FlatodometryKineticsobserverRelerrorTransxyStd}{0.011}
\newcommand{\FlatodometryKineticsobserverRelerrorTranszMeanabs}{0.005}
\newcommand{\FlatodometryKineticsobserverRelerrorTranszStd}{0.004}
\newcommand{\FlatodometryKineticsobserverRelerrorYawMeanabs}{0.31}
\newcommand{\FlatodometryKineticsobserverRelerrorYawStd}{0.26}
\newcommand{\LongwalkTiltRelerrorTiltMeanabs}{0.73}
\newcommand{\LongwalkTiltRelerrorTiltStd}{0.39}
\newcommand{\LongwalkTiltRelerrorTransxyMeanabs}{0.302}
\newcommand{\LongwalkTiltRelerrorTransxyStd}{0.218}
\newcommand{\LongwalkTiltRelerrorTranszMeanabs}{0.070}
\newcommand{\LongwalkTiltRelerrorTranszStd}{0.042}
\newcommand{\LongwalkTiltRelerrorYawMeanabs}{9.35}
\newcommand{\LongwalkTiltRelerrorYawStd}{6.76}
\newcommand{\LongwalkHartleyVelerrorEstimateNormMeanabs}{0.025}
\newcommand{\LongwalkHartleyVelerrorEstimateNormStd}{0.015}
\newcommand{\LongwalkHartleyVelerrorEstimateXyMeanabs}{0.091}
\newcommand{\LongwalkHartleyVelerrorEstimateXyStd}{0.065}
\newcommand{\LongwalkHartleyVelerrorEstimateZMeanabs}{0.020}
\newcommand{\LongwalkHartleyVelerrorEstimateZStd}{0.018} 
\newcommand{\LongwalkTiltVelerrorEstimateNormMeanabs}{0.035}
\newcommand{\LongwalkTiltVelerrorEstimateNormStd}{0.019}
\newcommand{\LongwalkTiltVelerrorEstimateXyMeanabs}{0.094}
\newcommand{\LongwalkTiltVelerrorEstimateXyStd}{0.066}
\newcommand{\LongwalkTiltVelerrorEstimateZMeanabs}{0.020}
\newcommand{\LongwalkTiltVelerrorEstimateZStd}{0.018} 
\newcommand{\LongwalkKineticsobserverVelerrorEstimateNormMeanabs}{0.034}
\newcommand{\LongwalkKineticsobserverVelerrorEstimateNormStd}{0.019}
\newcommand{\LongwalkKineticsobserverVelerrorEstimateXyMeanabs}{0.087}
\newcommand{\LongwalkKineticsobserverVelerrorEstimateXyStd}{0.061}
\newcommand{\LongwalkKineticsobserverVelerrorEstimateZMeanabs}{0.020}
\newcommand{\LongwalkKineticsobserverVelerrorEstimateZStd}{0.015} 
\newcommand{\LongwalkKowithoutwrenchsensorsVelerrorEstimateNormMeanabs}{0.034}
\newcommand{\LongwalkKowithoutwrenchsensorsVelerrorEstimateNormStd}{0.019}
\newcommand{\LongwalkKowithoutwrenchsensorsVelerrorEstimateXyMeanabs}{0.085}
\newcommand{\LongwalkKowithoutwrenchsensorsVelerrorEstimateXyStd}{0.059}
\newcommand{\LongwalkKowithoutwrenchsensorsVelerrorEstimateZMeanabs}{0.020}
\newcommand{\LongwalkKowithoutwrenchsensorsVelerrorEstimateZStd}{0.016} 
\newcommand{\LongwalkKoZpcVelerrorEstimateNormMeanabs}{0.034}
\newcommand{\LongwalkKoZpcVelerrorEstimateNormStd}{0.019}
\newcommand{\LongwalkKoZpcVelerrorEstimateXyMeanabs}{0.087}
\newcommand{\LongwalkKoZpcVelerrorEstimateXyStd}{0.061}
\newcommand{\LongwalkKoZpcVelerrorEstimateZMeanabs}{0.020}
\newcommand{\LongwalkKoZpcVelerrorEstimateZStd}{0.015}
\newcommand{\LongwalkHartleyRelerrorTiltMeanabs}{0.90}
\newcommand{\LongwalkHartleyRelerrorTiltStd}{0.39}
\newcommand{\LongwalkHartleyRelerrorTransxyMeanabs}{0.188}
\newcommand{\LongwalkHartleyRelerrorTransxyStd}{0.087}
\newcommand{\LongwalkHartleyRelerrorTranszMeanabs}{0.036}
\newcommand{\LongwalkHartleyRelerrorTranszStd}{0.025}
\newcommand{\LongwalkHartleyRelerrorYawMeanabs}{0.37}
\newcommand{\LongwalkHartleyRelerrorYawStd}{0.33}

\newcommand{\LongwalkKineticsobserverRelerrorTiltMeanabs}{0.57}
\newcommand{\LongwalkKineticsobserverRelerrorTiltStd}{0.32}
\newcommand{\LongwalkKineticsobserverRelerrorTransxyMeanabs}{0.186}
\newcommand{\LongwalkKineticsobserverRelerrorTransxyStd}{0.094}
\newcommand{\LongwalkKineticsobserverRelerrorTranszMeanabs}{0.037}
\newcommand{\LongwalkKineticsobserverRelerrorTranszStd}{0.024}
\newcommand{\LongwalkKineticsobserverRelerrorYawMeanabs}{0.19}
\newcommand{\LongwalkKineticsobserverRelerrorYawStd}{0.15}
\newcommand{\LongwalkKowithoutwrenchsensorsRelerrorTiltMeanabs}{0.56}
\newcommand{\LongwalkKowithoutwrenchsensorsRelerrorTiltStd}{0.32}
\newcommand{\LongwalkKowithoutwrenchsensorsRelerrorTransxyMeanabs}{0.186}
\newcommand{\LongwalkKowithoutwrenchsensorsRelerrorTransxyStd}{0.100}
\newcommand{\LongwalkKowithoutwrenchsensorsRelerrorTranszMeanabs}{0.066}
\newcommand{\LongwalkKowithoutwrenchsensorsRelerrorTranszStd}{0.029}
\newcommand{\LongwalkKowithoutwrenchsensorsRelerrorYawMeanabs}{0.26}
\newcommand{\LongwalkKowithoutwrenchsensorsRelerrorYawStd}{0.19}
\newcommand{\LongwalkKoZpcRelerrorTiltMeanabs}{0.56}
\newcommand{\LongwalkKoZpcRelerrorTiltStd}{0.32}
\newcommand{\LongwalkKoZpcRelerrorTransxyMeanabs}{0.177}
\newcommand{\LongwalkKoZpcRelerrorTransxyStd}{0.092}
\newcommand{\LongwalkKoZpcRelerrorTranszMeanabs}{0.031}
\newcommand{\LongwalkKoZpcRelerrorTranszStd}{0.020}
\newcommand{\LongwalkKoZpcRelerrorYawMeanabs}{0.22}
\newcommand{\LongwalkKoZpcRelerrorYawStd}{0.17}

\newcommand{\SlippingodometryHartleyVelerrorEstimateNormMeanabs}{0.030}
\newcommand{\SlippingodometryHartleyVelerrorEstimateNormStd}{0.032}
\newcommand{\SlippingodometryHartleyVelerrorEstimateXyMeanabs}{0.026}
\newcommand{\SlippingodometryHartleyVelerrorEstimateXyStd}{0.031}
\newcommand{\SlippingodometryHartleyVelerrorEstimateZMeanabs}{0.013}
\newcommand{\SlippingodometryHartleyVelerrorEstimateZStd}{0.017}
\newcommand{\SlippingodometryTiltVelerrorEstimateNormMeanabs}{0.042}
\newcommand{\SlippingodometryTiltVelerrorEstimateNormStd}{0.036}
\newcommand{\SlippingodometryTiltVelerrorEstimateXyMeanabs}{0.038}
\newcommand{\SlippingodometryTiltVelerrorEstimateXyStd}{0.037}
\newcommand{\SlippingodometryTiltVelerrorEstimateZMeanabs}{0.014}
\newcommand{\SlippingodometryTiltVelerrorEstimateZStd}{0.018}
\newcommand{\SlippingodometryKowithoutwrenchsensorsVelerrorEstimateNormMeanabs}{0.044}
\newcommand{\SlippingodometryKowithoutwrenchsensorsVelerrorEstimateNormStd}{0.040}
\newcommand{\SlippingodometryKowithoutwrenchsensorsVelerrorEstimateXyMeanabs}{0.036}
\newcommand{\SlippingodometryKowithoutwrenchsensorsVelerrorEstimateXyStd}{0.036}
\newcommand{\SlippingodometryKowithoutwrenchsensorsVelerrorEstimateZMeanabs}{0.018}
\newcommand{\SlippingodometryKowithoutwrenchsensorsVelerrorEstimateZStd}{0.020} 
\newcommand{\SlippingodometryKoZpcVelerrorEstimateNormMeanabs}{0.043}
\newcommand{\SlippingodometryKoZpcVelerrorEstimateNormStd}{0.039}
\newcommand{\SlippingodometryKoZpcVelerrorEstimateXyMeanabs}{0.037}
\newcommand{\SlippingodometryKoZpcVelerrorEstimateXyStd}{0.037}
\newcommand{\SlippingodometryKoZpcVelerrorEstimateZMeanabs}{0.017}
\newcommand{\SlippingodometryKoZpcVelerrorEstimateZStd}{0.019} 
\newcommand{\SlippingodometryKineticsobserverVelerrorEstimateNormMeanabs}{0.043}
\newcommand{\SlippingodometryKineticsobserverVelerrorEstimateNormStd}{0.039}
\newcommand{\SlippingodometryKineticsobserverVelerrorEstimateXyMeanabs}{0.033}
\newcommand{\SlippingodometryKineticsobserverVelerrorEstimateXyStd}{0.035}
\newcommand{\SlippingodometryKineticsobserverVelerrorEstimateZMeanabs}{0.016}
\newcommand{\SlippingodometryKineticsobserverVelerrorEstimateZStd}{0.019}  
\newcommand{\SlippingodometryHartleyRelerrorTiltMeanabs}{1.24}
\newcommand{\SlippingodometryHartleyRelerrorTiltStd}{1.15}
\newcommand{\SlippingodometryHartleyRelerrorTransxyMeanabs}{0.030}
\newcommand{\SlippingodometryHartleyRelerrorTransxyStd}{0.010}
\newcommand{\SlippingodometryHartleyRelerrorTranszMeanabs}{0.009}
\newcommand{\SlippingodometryHartleyRelerrorTranszStd}{0.007}
\newcommand{\SlippingodometryHartleyRelerrorYawMeanabs}{0.36}
\newcommand{\SlippingodometryHartleyRelerrorYawStd}{0.44}
\newcommand{\SlippingodometryTiltRelerrorTiltMeanabs}{1.53}
\newcommand{\SlippingodometryTiltRelerrorTiltStd}{1.21}
\newcommand{\SlippingodometryTiltRelerrorTransxyMeanabs}{0.042}
\newcommand{\SlippingodometryTiltRelerrorTransxyStd}{0.016}
\newcommand{\SlippingodometryTiltRelerrorTranszMeanabs}{0.007}
\newcommand{\SlippingodometryTiltRelerrorTranszStd}{0.006}
\newcommand{\SlippingodometryTiltRelerrorYawMeanabs}{2.54}
\newcommand{\SlippingodometryTiltRelerrorYawStd}{2.48}
\newcommand{\SlippingodometryKowithoutwrenchsensorsRelerrorTiltMeanabs}{1.22}
\newcommand{\SlippingodometryKowithoutwrenchsensorsRelerrorTiltStd}{1.15}
\newcommand{\SlippingodometryKowithoutwrenchsensorsRelerrorTransxyMeanabs}{0.022}
\newcommand{\SlippingodometryKowithoutwrenchsensorsRelerrorTransxyStd}{0.011}
\newcommand{\SlippingodometryKowithoutwrenchsensorsRelerrorTranszMeanabs}{0.006}
\newcommand{\SlippingodometryKowithoutwrenchsensorsRelerrorTranszStd}{0.006}
\newcommand{\SlippingodometryKowithoutwrenchsensorsRelerrorYawMeanabs}{0.38}
\newcommand{\SlippingodometryKowithoutwrenchsensorsRelerrorYawStd}{0.45}
\newcommand{\SlippingodometryKoZpcRelerrorTiltMeanabs}{1.26}
\newcommand{\SlippingodometryKoZpcRelerrorTiltStd}{1.15}
\newcommand{\SlippingodometryKoZpcRelerrorTransxyMeanabs}{0.027}
\newcommand{\SlippingodometryKoZpcRelerrorTransxyStd}{0.010}
\newcommand{\SlippingodometryKoZpcRelerrorTranszMeanabs}{0.007}
\newcommand{\SlippingodometryKoZpcRelerrorTranszStd}{0.007}
\newcommand{\SlippingodometryKoZpcRelerrorYawMeanabs}{0.38}
\newcommand{\SlippingodometryKoZpcRelerrorYawStd}{0.44}
\newcommand{\SlippingodometryKineticsobserverRelerrorTiltMeanabs}{1.25}
\newcommand{\SlippingodometryKineticsobserverRelerrorTiltStd}{1.15}
\newcommand{\SlippingodometryKineticsobserverRelerrorTransxyMeanabs}{0.018}
\newcommand{\SlippingodometryKineticsobserverRelerrorTransxyStd}{0.009}
\newcommand{\SlippingodometryKineticsobserverRelerrorTranszMeanabs}{0.007}
\newcommand{\SlippingodometryKineticsobserverRelerrorTranszStd}{0.007}
\newcommand{\SlippingodometryKineticsobserverRelerrorYawMeanabs}{0.38}
\newcommand{\SlippingodometryKineticsobserverRelerrorYawStd}{0.44}

\newcommand{\MulticontactHartleyVelerrorEstimateNormMeanabs}{0.009}
\newcommand{\MulticontactHartleyVelerrorEstimateNormStd}{0.009}
\newcommand{\MulticontactHartleyVelerrorEstimateXyMeanabs}{0.010}
\newcommand{\MulticontactHartleyVelerrorEstimateXyStd}{0.011}
\newcommand{\MulticontactHartleyVelerrorEstimateZMeanabs}{0.005}
\newcommand{\MulticontactHartleyVelerrorEstimateZStd}{0.009} 
\newcommand{\MulticontactTiltVelerrorEstimateNormMeanabs}{0.011}
\newcommand{\MulticontactTiltVelerrorEstimateNormStd}{0.011}
\newcommand{\MulticontactTiltVelerrorEstimateXyMeanabs}{0.009}
\newcommand{\MulticontactTiltVelerrorEstimateXyStd}{0.010}
\newcommand{\MulticontactTiltVelerrorEstimateZMeanabs}{0.004}
\newcommand{\MulticontactTiltVelerrorEstimateZStd}{0.007} 
\newcommand{\MulticontactKowithoutwrenchsensorsVelerrorEstimateNormMeanabs}{0.011}
\newcommand{\MulticontactKowithoutwrenchsensorsVelerrorEstimateNormStd}{0.011}
\newcommand{\MulticontactKowithoutwrenchsensorsVelerrorEstimateXyMeanabs}{0.011}
\newcommand{\MulticontactKowithoutwrenchsensorsVelerrorEstimateXyStd}{0.012}
\newcommand{\MulticontactKowithoutwrenchsensorsVelerrorEstimateZMeanabs}{0.005}
\newcommand{\MulticontactKowithoutwrenchsensorsVelerrorEstimateZStd}{0.008} 
\newcommand{\MulticontactKoZpcVelerrorEstimateNormMeanabs}{0.011}
\newcommand{\MulticontactKoZpcVelerrorEstimateNormStd}{0.012}
\newcommand{\MulticontactKoZpcVelerrorEstimateXyMeanabs}{0.012}
\newcommand{\MulticontactKoZpcVelerrorEstimateXyStd}{0.011}
\newcommand{\MulticontactKoZpcVelerrorEstimateZMeanabs}{0.005}
\newcommand{\MulticontactKoZpcVelerrorEstimateZStd}{0.007}
\newcommand{\MulticontactKineticsobserverVelerrorEstimateNormMeanabs}{0.011}
\newcommand{\MulticontactKineticsobserverVelerrorEstimateNormStd}{0.012}
\newcommand{\MulticontactKineticsobserverVelerrorEstimateXyMeanabs}{0.010}
\newcommand{\MulticontactKineticsobserverVelerrorEstimateXyStd}{0.011}
\newcommand{\MulticontactKineticsobserverVelerrorEstimateZMeanabs}{0.005}
\newcommand{\MulticontactKineticsobserverVelerrorEstimateZStd}{0.007} 
\newcommand{\MulticontactHartleyRelerrorTiltMeanabs}{0.22}
\newcommand{\MulticontactHartleyRelerrorTiltStd}{0.16}
\newcommand{\MulticontactHartleyRelerrorTransxyMeanabs}{0.006}
\newcommand{\MulticontactHartleyRelerrorTransxyStd}{0.004}
\newcommand{\MulticontactHartleyRelerrorTranszMeanabs}{0.002}
\newcommand{\MulticontactHartleyRelerrorTranszStd}{0.001}
\newcommand{\MulticontactHartleyRelerrorYawMeanabs}{0.12}
\newcommand{\MulticontactHartleyRelerrorYawStd}{0.11}
\newcommand{\MulticontactKowithoutwrenchsensorsRelerrorTiltMeanabs}{0.22}
\newcommand{\MulticontactKowithoutwrenchsensorsRelerrorTiltStd}{0.16}
\newcommand{\MulticontactKowithoutwrenchsensorsRelerrorTransxyMeanabs}{0.008}
\newcommand{\MulticontactKowithoutwrenchsensorsRelerrorTransxyStd}{0.010}
\newcommand{\MulticontactKowithoutwrenchsensorsRelerrorTranszMeanabs}{0.001}
\newcommand{\MulticontactKowithoutwrenchsensorsRelerrorTranszStd}{0.001}
\newcommand{\MulticontactKowithoutwrenchsensorsRelerrorYawMeanabs}{0.12}
\newcommand{\MulticontactKowithoutwrenchsensorsRelerrorYawStd}{0.11} 
\newcommand{\MulticontactKoZpcRelerrorTiltMeanabs}{0.33}
\newcommand{\MulticontactKoZpcRelerrorTiltStd}{0.20}
\newcommand{\MulticontactKoZpcRelerrorTransxyMeanabs}{0.006}
\newcommand{\MulticontactKoZpcRelerrorTransxyStd}{0.004}
\newcommand{\MulticontactKoZpcRelerrorTranszMeanabs}{0.002}
\newcommand{\MulticontactKoZpcRelerrorTranszStd}{0.001}
\newcommand{\MulticontactKoZpcRelerrorYawMeanabs}{0.36}
\newcommand{\MulticontactKoZpcRelerrorYawStd}{0.25}
\newcommand{\MulticontactKineticsobserverRelerrorTiltMeanabs}{0.22}
\newcommand{\MulticontactKineticsobserverRelerrorTiltStd}{0.15}
\newcommand{\MulticontactKineticsobserverRelerrorTransxyMeanabs}{0.005}
\newcommand{\MulticontactKineticsobserverRelerrorTransxyStd}{0.004}
\newcommand{\MulticontactKineticsobserverRelerrorTranszMeanabs}{0.001}
\newcommand{\MulticontactKineticsobserverRelerrorTranszStd}{0.001}
\newcommand{\MulticontactKineticsobserverRelerrorYawMeanabs}{0.12}
\newcommand{\MulticontactKineticsobserverRelerrorYawStd}{0.11}   
\newcommand{\MulticontactTiltRelerrorTiltMeanabs}{0.33}
\newcommand{\MulticontactTiltRelerrorTiltStd}{0.24}
\newcommand{\MulticontactTiltRelerrorTransxyMeanabs}{0.007}
\newcommand{\MulticontactTiltRelerrorTransxyStd}{0.005}
\newcommand{\MulticontactTiltRelerrorTranszMeanabs}{0.001}
\newcommand{\MulticontactTiltRelerrorTranszStd}{0.001}
\newcommand{\MulticontactTiltRelerrorYawMeanabs}{0.42}
\newcommand{\MulticontactTiltRelerrorYawStd}{0.31}

\newcommand{\MulticontactbKineticsobserverVelerrorEstimateNormMeanabs}{0.013}
\newcommand{\MulticontactbKineticsobserverVelerrorEstimateNormStd}{0.015}
\newcommand{\MulticontactbKineticsobserverVelerrorEstimateXyMeanabs}{0.013}
\newcommand{\MulticontactbKineticsobserverVelerrorEstimateXyStd}{0.016}
\newcommand{\MulticontactbKineticsobserverVelerrorEstimateZMeanabs}{0.009}
\newcommand{\MulticontactbKineticsobserverVelerrorEstimateZStd}{0.013}
\newcommand{\MulticontactbKineticsobserverVelerrorLlveNormMeanabs}{0.015}
\newcommand{\MulticontactbKineticsobserverVelerrorLlveNormStd}{0.025}
\newcommand{\MulticontactbKineticsobserverVelerrorLlveXyMeanabs}{0.010}
\newcommand{\MulticontactbKineticsobserverVelerrorLlveXyStd}{0.015}
\newcommand{\MulticontactbKineticsobserverVelerrorLlveZMeanabs}{0.009}
\newcommand{\MulticontactbKineticsobserverVelerrorLlveZStd}{0.021}
\newcommand{\MulticontactbKineticsobserverRelerrorTiltMeanabs}{0.22}
\newcommand{\MulticontactbKineticsobserverRelerrorTiltStd}{0.15}
\newcommand{\MulticontactbKineticsobserverRelerrorTransxyMeanabs}{0.006}
\newcommand{\MulticontactbKineticsobserverRelerrorTransxyStd}{0.005}
\newcommand{\MulticontactbKineticsobserverRelerrorTranszMeanabs}{0.004}
\newcommand{\MulticontactbKineticsobserverRelerrorTranszStd}{0.004}
\newcommand{\MulticontactbKineticsobserverRelerrorYawMeanabs}{0.12}
\newcommand{\MulticontactbKineticsobserverRelerrorYawStd}{0.10}

\newcommand{\MulticontactcKineticsobserverVelerrorEstimateNormMeanabs}{0.011}
\newcommand{\MulticontactcKineticsobserverVelerrorEstimateNormStd}{0.013}
\newcommand{\MulticontactcKineticsobserverVelerrorEstimateXyMeanabs}{0.012}
\newcommand{\MulticontactcKineticsobserverVelerrorEstimateXyStd}{0.011}
\newcommand{\MulticontactcKineticsobserverVelerrorEstimateZMeanabs}{0.005}
\newcommand{\MulticontactcKineticsobserverVelerrorEstimateZStd}{0.007}
\newcommand{\MulticontactcKineticsobserverVelerrorLlveNormMeanabs}{0.012}
\newcommand{\MulticontactcKineticsobserverVelerrorLlveNormStd}{0.014}
\newcommand{\MulticontactcKineticsobserverVelerrorLlveXyMeanabs}{0.011}
\newcommand{\MulticontactcKineticsobserverVelerrorLlveXyStd}{0.013}
\newcommand{\MulticontactcKineticsobserverVelerrorLlveZMeanabs}{0.003}
\newcommand{\MulticontactcKineticsobserverVelerrorLlveZStd}{0.004}
\newcommand{\MulticontactcKineticsobserverRelerrorTiltMeanabs}{0.41}
\newcommand{\MulticontactcKineticsobserverRelerrorTiltStd}{0.39}
\newcommand{\MulticontactcKineticsobserverRelerrorTransxyMeanabs}{0.007}
\newcommand{\MulticontactcKineticsobserverRelerrorTransxyStd}{0.005}
\newcommand{\MulticontactcKineticsobserverRelerrorTranszMeanabs}{0.002}
\newcommand{\MulticontactcKineticsobserverRelerrorTranszStd}{0.001}
\newcommand{\MulticontactcKineticsobserverRelerrorYawMeanabs}{0.13}
\newcommand{\MulticontactcKineticsobserverRelerrorYawStd}{0.11}

\newcommand{\SlippingodometrybKineticsobserverVelerrorEstimateNormMeanabs}{0.098}
\newcommand{\SlippingodometrybKineticsobserverVelerrorEstimateNormStd}{0.068}
\newcommand{\SlippingodometrybKineticsobserverVelerrorEstimateXyMeanabs}{0.037}
\newcommand{\SlippingodometrybKineticsobserverVelerrorEstimateXyStd}{0.035}
\newcommand{\SlippingodometrybKineticsobserverVelerrorEstimateZMeanabs}{0.080}
\newcommand{\SlippingodometrybKineticsobserverVelerrorEstimateZStd}{0.053}
\newcommand{\SlippingodometrybKineticsobserverVelerrorLlveNormMeanabs}{0.183}
\newcommand{\SlippingodometrybKineticsobserverVelerrorLlveNormStd}{0.139}
\newcommand{\SlippingodometrybKineticsobserverVelerrorLlveXyMeanabs}{0.060}
\newcommand{\SlippingodometrybKineticsobserverVelerrorLlveXyStd}{0.049}
\newcommand{\SlippingodometrybKineticsobserverVelerrorLlveZMeanabs}{0.162}
\newcommand{\SlippingodometrybKineticsobserverVelerrorLlveZStd}{0.143} 
\newcommand{\SlippingodometrybKineticsobserverRelerrorTiltMeanabs}{1.25}
\newcommand{\SlippingodometrybKineticsobserverRelerrorTiltStd}{1.14}
\newcommand{\SlippingodometrybKineticsobserverRelerrorTransxyMeanabs}{0.040}
\newcommand{\SlippingodometrybKineticsobserverRelerrorTransxyStd}{0.024}
\newcommand{\SlippingodometrybKineticsobserverRelerrorTranszMeanabs}{0.652}
\newcommand{\SlippingodometrybKineticsobserverRelerrorTranszStd}{0.167}
\newcommand{\SlippingodometrybKineticsobserverRelerrorYawMeanabs}{0.37}
\newcommand{\SlippingodometrybKineticsobserverRelerrorYawStd}{0.44}

\newcommand{\SlippingodometrycKineticsobserverVelerrorEstimateNormMeanabs}{0.047}
\newcommand{\SlippingodometrycKineticsobserverVelerrorEstimateNormStd}{0.040}
\newcommand{\SlippingodometrycKineticsobserverVelerrorEstimateXyMeanabs}{0.039}
\newcommand{\SlippingodometrycKineticsobserverVelerrorEstimateXyStd}{0.038}
\newcommand{\SlippingodometrycKineticsobserverVelerrorEstimateZMeanabs}{0.014}
\newcommand{\SlippingodometrycKineticsobserverVelerrorEstimateZStd}{0.017}
\newcommand{\SlippingodometrycKineticsobserverVelerrorLlveNormMeanabs}{0.080}
\newcommand{\SlippingodometrycKineticsobserverVelerrorLlveNormStd}{0.061}
\newcommand{\SlippingodometrycKineticsobserverVelerrorLlveXyMeanabs}{0.077}
\newcommand{\SlippingodometrycKineticsobserverVelerrorLlveXyStd}{0.061}
\newcommand{\SlippingodometrycKineticsobserverVelerrorLlveZMeanabs}{0.016}
\newcommand{\SlippingodometrycKineticsobserverVelerrorLlveZStd}{0.018} 
\newcommand{\SlippingodometrycKineticsobserverRelerrorTiltMeanabs}{1.78}
\newcommand{\SlippingodometrycKineticsobserverRelerrorTiltStd}{1.21}
\newcommand{\SlippingodometrycKineticsobserverRelerrorTransxyMeanabs}{0.031}
\newcommand{\SlippingodometrycKineticsobserverRelerrorTransxyStd}{0.014}
\newcommand{\SlippingodometrycKineticsobserverRelerrorTranszMeanabs}{0.011}
\newcommand{\SlippingodometrycKineticsobserverRelerrorTranszStd}{0.009}
\newcommand{\SlippingodometrycKineticsobserverRelerrorYawMeanabs}{0.73}
\newcommand{\SlippingodometrycKineticsobserverRelerrorYawStd}{0.71} 

% Measured on the retained configuration, not present in the previous file.
\newcommand{\ExtwrenchFxMeanabs}{6.5}
\newcommand{\ExtwrenchFxStd}{6.6}
\newcommand{\ExtwrenchFyMeanabs}{9.7}
\newcommand{\ExtwrenchFyStd}{10.9}
\newcommand{\ExtwrenchFzMeanabs}{9.2}
\newcommand{\ExtwrenchFzStd}{13.1}
\newcommand{\ExtwrenchTxMeanabs}{10.3}
\newcommand{\ExtwrenchTxStd}{11.2}
\newcommand{\ExtwrenchTyMeanabs}{4.4}
\newcommand{\ExtwrenchTyStd}{5.9}
\newcommand{\ExtwrenchTzMeanabs}{3.6}
\newcommand{\ExtwrenchTzStd}{4.1}

% Measured on the retained configuration, not present in the previous file.
\newcommand{\FlatodometryTiltRelerrorTiltMeanabs}{0.83}
\newcommand{\FlatodometryTiltRelerrorTiltStd}{0.52}
\newcommand{\FlatodometryTiltRelerrorTransxyMeanabs}{0.034}
\newcommand{\FlatodometryTiltRelerrorTransxyStd}{0.024}
\newcommand{\FlatodometryTiltRelerrorTranszMeanabs}{0.005}
\newcommand{\FlatodometryTiltRelerrorTranszStd}{0.005}
\newcommand{\FlatodometryTiltRelerrorYawMeanabs}{1.19}
\newcommand{\FlatodometryTiltRelerrorYawStd}{1.24}
\newcommand{\FlatodometryTiltVelerrorEstimateXyMeanabs}{0.025}
\newcommand{\FlatodometryTiltVelerrorEstimateXyStd}{0.029}
\newcommand{\FlatodometryTiltVelerrorEstimateZMeanabs}{0.010}
\newcommand{\FlatodometryTiltVelerrorEstimateZStd}{0.015}

% Measured on the retained configuration, not present in the previous file.
\newcommand{\FlatodometryKolinearRelerrorTiltMeanabs}{0.51}
\newcommand{\FlatodometryKolinearRelerrorTiltStd}{0.32}
\newcommand{\FlatodometryKolinearRelerrorTransxyMeanabs}{0.018}
\newcommand{\FlatodometryKolinearRelerrorTransxyStd}{0.018}
\newcommand{\FlatodometryKolinearRelerrorTranszMeanabs}{0.005}
\newcommand{\FlatodometryKolinearRelerrorTranszStd}{0.004}
\newcommand{\FlatodometryKolinearRelerrorYawMeanabs}{1.12}
\newcommand{\FlatodometryKolinearRelerrorYawStd}{2.63}
\newcommand{\FlatodometryKolinearVelerrorEstimateXyMeanabs}{0.023}
\newcommand{\FlatodometryKolinearVelerrorEstimateXyStd}{0.027}
\newcommand{\FlatodometryKolinearVelerrorEstimateZMeanabs}{0.012}
\newcommand{\FlatodometryKolinearVelerrorEstimateZStd}{0.016}
\newcommand{\LongwalkKolinearRelerrorTiltMeanabs}{0.57}
\newcommand{\LongwalkKolinearRelerrorTiltStd}{0.32}
\newcommand{\LongwalkKolinearRelerrorTransxyMeanabs}{0.183}
\newcommand{\LongwalkKolinearRelerrorTransxyStd}{0.091}
\newcommand{\LongwalkKolinearRelerrorTranszMeanabs}{0.037}
\newcommand{\LongwalkKolinearRelerrorTranszStd}{0.024}
\newcommand{\LongwalkKolinearRelerrorYawMeanabs}{0.28}
\newcommand{\LongwalkKolinearRelerrorYawStd}{0.52}
\newcommand{\LongwalkKolinearVelerrorEstimateXyMeanabs}{0.087}
\newcommand{\LongwalkKolinearVelerrorEstimateXyStd}{0.061}
\newcommand{\LongwalkKolinearVelerrorEstimateZMeanabs}{0.020}
\newcommand{\LongwalkKolinearVelerrorEstimateZStd}{0.015}
\newcommand{\MulticontactKolinearRelerrorTiltMeanabs}{0.22}
\newcommand{\MulticontactKolinearRelerrorTiltStd}{0.15}
\newcommand{\MulticontactKolinearRelerrorTransxyMeanabs}{0.005}
\newcommand{\MulticontactKolinearRelerrorTransxyStd}{0.004}
\newcommand{\MulticontactKolinearRelerrorTranszMeanabs}{0.001}
\newcommand{\MulticontactKolinearRelerrorTranszStd}{0.001}
\newcommand{\MulticontactKolinearRelerrorYawMeanabs}{0.12}
\newcommand{\MulticontactKolinearRelerrorYawStd}{0.11}
\newcommand{\MulticontactKolinearVelerrorEstimateXyMeanabs}{0.010}
\newcommand{\MulticontactKolinearVelerrorEstimateXyStd}{0.012}
\newcommand{\MulticontactKolinearVelerrorEstimateZMeanabs}{0.005}
\newcommand{\MulticontactKolinearVelerrorEstimateZStd}{0.007}
\newcommand{\SlippingodometryKolinearRelerrorTiltMeanabs}{1.25}
\newcommand{\SlippingodometryKolinearRelerrorTiltStd}{1.15}
\newcommand{\SlippingodometryKolinearRelerrorTransxyMeanabs}{0.017}
\newcommand{\SlippingodometryKolinearRelerrorTransxyStd}{0.009}
\newcommand{\SlippingodometryKolinearRelerrorTranszMeanabs}{0.007}
\newcommand{\SlippingodometryKolinearRelerrorTranszStd}{0.007}
\newcommand{\SlippingodometryKolinearRelerrorYawMeanabs}{0.43}
\newcommand{\SlippingodometryKolinearRelerrorYawStd}{0.47}
\newcommand{\SlippingodometryKolinearVelerrorEstimateXyMeanabs}{0.033}
\newcommand{\SlippingodometryKolinearVelerrorEstimateXyStd}{0.035}
\newcommand{\SlippingodometryKolinearVelerrorEstimateZMeanabs}{0.016}
\newcommand{\SlippingodometryKolinearVelerrorEstimateZStd}{0.019}

\subsection{Experimental Framework}
\subsubsection{Setup, baseline and ground truth}
The experiments were conducted on two different humanoid robots: RHP Friends (Figure~\ref{fig:rhp-friends}) and HRP-5P (Figure~\ref{fig:hrp-5p}). Both robots are equipped with wrench sensors on the contact limbs and an IMU located in the upper body. In all the scenarios, the robots were controlled via the mc\_rtc framework\endnote{\url{https://jrl-umi3218.github.io/mc\_rtc/}}. 

Our odometry is compared to that performed by the state-of-the-art right-invariant EKF (RI-EKF) by \citet{Hartley2020RIEKF}. Both estimators were tuned using identical noise models, for the IMU measurements, and for the process associated with the gyrometer bias. This ensures that both methods utilize the sensor data equally.  

As an additional baseline, we include VALINOR in the quantitative comparison. This estimator, based on a complementary filter, was introduced in our previous work~\citep{demont2025ValinorIJCAS} as an alternative to the RI-EKF. It is lightweight and provides highly accurate estimates of tilt and local linear velocity, with global asymptotic convergence that has been mathematically proven. 

All estimators are designed to receive the contact states from any contact detection framework. For a fair comparison, they receive the same information from a Schmitt Trigger on the forces measured by the sensors. We set the lower threshold of the Schmitt Trigger to 10\% of the weight and its upper threshold to 12\% of the robot's weight.
Ground-truth data for the robot's pose are obtained by motion capture (Opti Track system with 16 Prime\textsuperscript{X} 13 cameras). The ground truth velocity is obtained via finite differences from the ground truth position, then filtered with a zero-phase low pass filter.

To the trajectory plots, we also include the robot-modeled trajectory, resulting from the integration of the commands at the output of the controller. This trajectory assumes perfect tracking of the references, and can thus be compared against the ground truth trajectory in order to assess the discrepancy between our ideal model of the system \{robot + environment\} and the actual system. We refer to this modeled trajectory as the \textquotedbl control\textquotedbl{} trajectory. Note that neither the Kinetics Observer nor the RI-EKF have access to the control trajectory: they rely solely on measurements.
\subsubsection{Experiments overview}
The estimation performance of the Kinetics Observer was evaluated in three scenarios across 13 sequences, totaling approximately 305 meters of walking. The scenarios are as follows: (1) multi-contact motion on tilted obstacles, aimed at evaluating estimation performance with three non-coplanar contacts; (2) two flat-ground walking experiments designed to assess long-term and long-distance estimation over trajectories involving many contact creation events; and (3) walking with slipping contacts, used to evaluate robustness to unreliable contacts. The first experiment was adapted to additionally evaluate the estimation of contact local geometry and disturbance wrenches. Finally, the long walk is replayed with a simulated gyrometer bias drift to evaluate the robustness of the estimators to this drift. A summary of all experiments and their key conclusions is provided in Table~\ref{tab:expSummary}.

\begin{table*}
    \centering
    \footnotesize
    \caption{Summary of the experimental scenarios and their key conclusions.}\label{tab:expSummary}
    \setlength{\tabcolsep}{4.5pt}
    \begin{tabularx}{\textwidth}{  l   c   c   l  }
            \toprule
            \multirow{2}{*}{Experiment}      &  \multirow{2}{*}{Nb. trials}    & Total  &   \multirow{2}{*}{Conclusion on the Kinetics Observer}  \\
                                &                                 & Dist. [m] & \\
            \midrule
            Walk on flat ground~(\ref{subsec:flat-odometry}) 
            & \makecell{5 \\ \\ 1}
            & \makecell{90 \\ \\ 192}
            & \gape{\makecell[l]{Comparison to the RI-EKF: \\
            \tabitem Better tracking of lateral translations and yaw. }} \\
            \midrule
             Walking with slipping contacts~(\ref{subsec:Slipping-odometry})    &       3    &  15  &       \gape{\makecell[l]{Comparison to the RI-EKF:       \\
                                                                        \tabitem Better robustness of lateral translations to slippage.}}          \\
             \midrule
             Robustness to a drifting gyrometer bias~(\ref{subsec:gyro-bias-drift})    &       1    &     &   \gape{\makecell[l]{Comparison to the RI-EKF:       \\ \tabitem Ability to track yaw-axis gyrometer drift in static double support phases. }}           \\
             \midrule
             Multi-contact motion with tilted obstacles~(\ref{subsec:Multi-contact-with-tilted})               &       4       &  8  &   Improved odometry thanks to the \emph{constrained contact process covariances}.          \\
             \midrule
             Est. of the Contact’s Local Geometry~(\ref{subsec:contact-char-estimation})            &       1   &    &       Reconstruction of the contact's orientation despite considerable initial error.  \\
             \midrule
             Disturbance wrench estimation~(\ref{subsec:ext-wrench-estimation})         &       4    &   &       Accurate and reactive estimation of a significant disturbance wrench.    \\
             \bottomrule

        \end{tabularx}
    \end{table*}

\subsubsection{Ablation study}
To demonstrate the improvements achieved by the \emph{constrained contact process covariances} feature introduced in Section~\ref{sec:Implementation}, we add to the comparison a variant of the Kinetics Observer that does not implement it. Using KO as a diminutive for the Kinetics Observer, this variant is referred to as KO-ZPC (Zero Process Covariance). 

We also introduce the KO-Lin variant of the Kinetics Observer, which removes the contact orientations and the angular part of the viscoelastic model. This variant shows that the angular part remains necessary to obtain the best estimation, notably of the yaw, but that the Kinetics Observer already performs very well with the linear part alone. It preserves the estimates of the gyrometer bias and of the disturbance wrench, and all the covariances tuned for the Kinetics Observer. 

Also, the Kinetics Observer is able to estimate the robot's pose even in the absence of contact orientations and contact wrench sensor measurements. To prove this, we include the evaluation of the trajectory estimated by this variant of the Kinetics Observer to our results. We refer to this variant as the KO-PC (Pin Contacts). We solely consider a damping on the angular velocity of the contact limb around the yaw axis to consider some friction when turning, since these ablations remove most of the feed-forward  and correction of rotations in the model, especially in single support phases. Since the RI-EKF exploits the gyrometer measurement directly as a feed-forward, we can expect lower performances of the KO-PC for the yaw estimation by bipedal locomotion. 
For a fair comparison of the trajectory estimation performance, we decided to preserve the same tuning of the filter across all variants of the Kinetics Observer and across all experiments. The information redundancy coming from the wrench sensor being lost in the KO-PC, we removed the disturbance wrench and the gyrometer bias from its state\endnote{Note that there was no further adaptation of this variant. This means that we may expect better results with extensive tuning, especially in terms of covariance of the viscoelastic model.}. Note that the analysis of the results will focus on the Kinetics Observer, the KO-ZPC, and the KO-Lin.

\subsubsection{Metrics for odometry and velocity estimation evaluation}\label{par:metrics}
The odometry performances are evaluated against the ground truth using the Relative Error (RE) on the robot's floating base trajectory, as described in~\citep{Zhang2018QuantitativeTrajectoryEvaluation}. The Relative Error allows to assess local drift of the odometry over a defined distance. It is particularly suitable for evaluating proprioceptive odometry, since it is not affected by a global drift of the estimated trajectory.  
We note that lateral (along the $\boldsymbol{x}$ and $\boldsymbol{y}$ axes) and vertical (along the $\boldsymbol{z}$ axis) translation errors are evaluated independently, as their errors originate from different sources and they can represent distinct targets for odometries. Indeed, the accurate estimation of lateral translations relies on both accurate estimates of local displacements and the robot's orientation in the world relative to the vertical axis (yaw angle). In contrast, the quality of vertical translation estimation depends on the accurate estimation of the robot's tilt , i.e. its inclination~\citep{allgeuer2015fusedAnglesTilt} in the world, as well as the estimator's ability to account for potential errors in new contact heights, notably due to material deformations. Accordingly, the errors in yaw and in tilt angle estimations are also distinguished. The error $e_{\ell}$ in the tilt estimate is computed with:
\begin{equation}
    e_{\ell} = \text{arccos}\left(\boldsymbol{\ell}_{gt}^{\top} \hat{\boldsymbol{\ell}} \right),
\end{equation}
with $\boldsymbol{\ell}_{gt}$ the ground truth tilt, and $\hat{\boldsymbol{\ell}} = \hat{\boldsymbol{R}}^{\top}\boldsymbol{e}_{z}$ the estimated one.  
The linear velocity estimation is also evaluated. We express it in the IMU's frame such that it is independent of the robot's orientation. We then provide the norm of the error on the lateral and vertical velocities.

\subsection{Multi-contact Motion with Tilted Obstacles}\label{subsec:Multi-contact-with-tilted}
\begin{figure*}
    \centering
    \begin{subfigure}{0.36\textwidth}
        \centering
        \includegraphics[height=6.5cm]{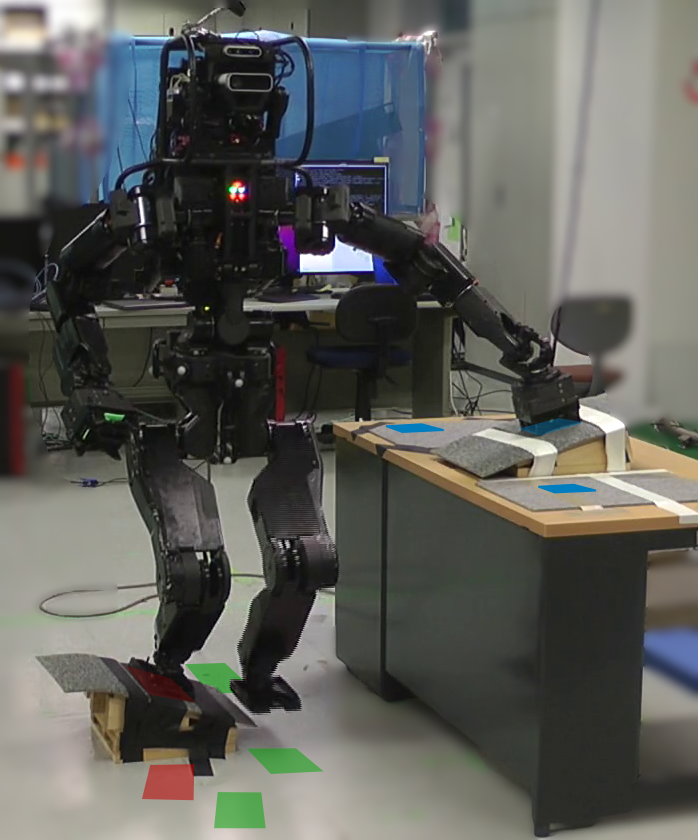}
        \caption{}
        \label{fig:hrp-5p}
    \end{subfigure}
    \hfil
    \begin{subfigure}{0.60\textwidth}
        \centering
        \includegraphics[height=6.5cm]{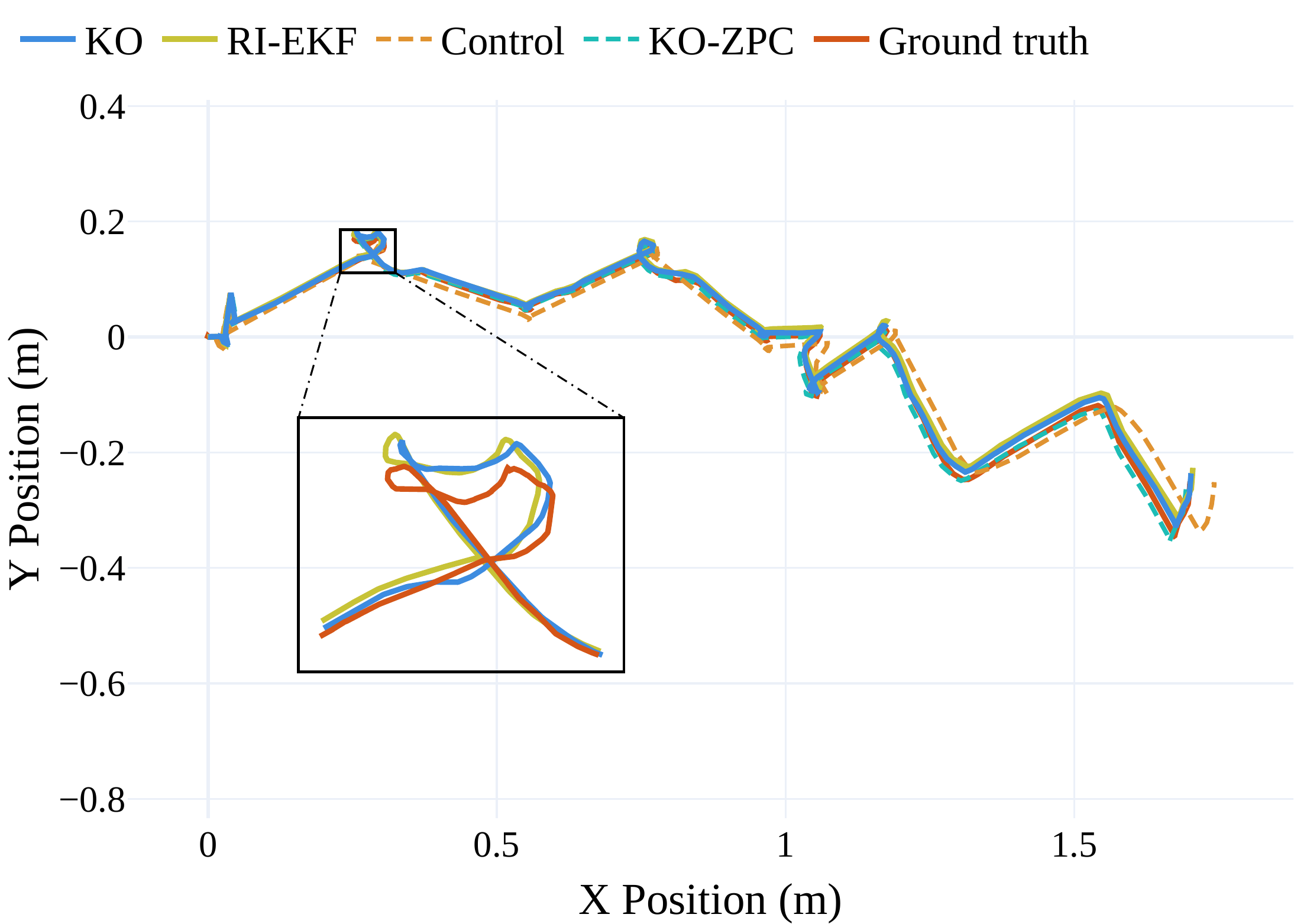}
        \caption{}
        \label{fig:multiContact-odom-traj}
    \end{subfigure}
    \caption{(a) HRP-5P walking over tilted obstacles during the multi-contact experiment. The successive imprints of the main contacts of the motion are represented by colored trapezoids. In \textit{red}: Right foot. In \textit{green}: Left foot. In \textit{blue}: Left hand. (b) Top view of the ground truth and estimated trajectories of the robot's floating base during the first multi-contact motion. The robot starts at the point (0,0) and walks towards the $\boldsymbol{x}$ direction. }
    \label{fig:multi-contract-odom}
\end{figure*}

\noindent To evaluate the contact pose estimation and the odometry performance of the Kinetics Observer over multi-contact motion, it was tested within a non-coplanar contact scenario using a multi-contact controller{\endnote{\url{https://github.com/isri-aist/MultiContactController}}}. This trajectory, represented in Figure~\ref{fig:multiContact-odom-traj}, involves stepping and pushing with the left hand on oriented and elevated obstacles, over a distance of about 2 meters.  This experiment was performed four times on the humanoid robot HRP-5P. 

Obstacles are placed at controlled positions and orientations, and thin carpet tiles are placed on their surfaces to increase contact friction, ensuring the correct execution of the motion planned offline by the multi-contact controller. These tiles introduce negligible compliance, which is not accounted for in the viscoelastic model. As explained in Appendix~\ref{subsec:tuning-of-KO}, the Kinetics Observer is robust to such approximations in the flexibility tuning.

\begin{table}
    \footnotesize
    \centering
    \caption{Mean and standard deviation (in parentheses) of errors computed during multi-contact motions. The 0.3 m Relative Error is represented. The best results for each metric are highlighted in bold.} \label{tab:multicontact-odometry-results}
        \begin{tabularx}{\columnwidth}{ C c c c c c c  }
            \toprule
            \multirow{4}{*}{} 
            & \multicolumn{2}{c}{RE Trans.} 
            & \multicolumn{2}{c}{RE Ori.} 
            & \multicolumn{2}{c}{Lin. vel.} \\

            & \multicolumn{2}{c}{[m]} 
            & \multicolumn{2}{c}{[deg]} 
            & \multicolumn{2}{c}{$[\text{m.s}^{-1}]$} \\ 
             
            \addlinespace[0.2em]

            \cline{2-7}

            \addlinespace[0.2em]

                        &       $\left\{\boldsymbol{x}, \boldsymbol{y}\right\}$   &         $\boldsymbol{z}$    & $\text{Tilt}$ &    $\text{Yaw}$  &   $\left\{\boldsymbol{x}, \boldsymbol{y}\right\}$  &  $\boldsymbol{z}$  \\
                        \midrule     
            Kinetics      &  \textbf{\getErrorResult{Multicontact}{Kineticsobserver}{Relerror}{Transxy}{Meanabs}}   &  \textbf{\getErrorResult{Multicontact}{Kineticsobserver}{Relerror}{Transz}{Meanabs}}      &  \textbf{\getErrorResult{Multicontact}{Kineticsobserver}{Relerror}{Tilt}{Meanabs}}      &    \textbf{\getErrorResult{Multicontact}{Kineticsobserver}{Relerror}{Yaw}{Meanabs}}    &  \getErrorResult{Multicontact}{Kineticsobserver}{Velerror}{EstimateXy}{Meanabs}    &   \getErrorResult{Multicontact}{Kineticsobserver}{Velerror}{EstimateZ}{Meanabs}  \\ 
            Observer        &    \scalebox{0.9}{\scriptsize(\getErrorResult{Multicontact}{Kineticsobserver}{Relerror}{Transxy}{Std})}    &     \scalebox{0.9}{\scriptsize(\getErrorResult{Multicontact}{Kineticsobserver}{Relerror}{Transz}{Std})}        &   \scalebox{0.9}{\scriptsize(\getErrorResult{Multicontact}{Kineticsobserver}{Relerror}{Tilt}{Std})}      &    \scalebox{0.9}{\scriptsize(\getErrorResult{Multicontact}{Kineticsobserver}{Relerror}{Yaw}{Std})}    &  \scalebox{0.9}{\scriptsize(\getErrorResult{Multicontact}{Kineticsobserver}{Velerror}{EstimateXy}{Std})}    &   \scalebox{0.9}{\scriptsize(\getErrorResult{Multicontact}{Kineticsobserver}{Velerror}{EstimateZ}{Std})}  \\ 
            \midrule

            KO-ZPC      &  \getErrorResult{Multicontact}{KoZpc}{Relerror}{Transxy}{Meanabs}  &   \getErrorResult{Multicontact}{KoZpc}{Relerror}{Transz}{Meanabs}   &      \getErrorResult{Multicontact}{KoZpc}{Relerror}{Tilt}{Meanabs}      &    \getErrorResult{Multicontact}{KoZpc}{Relerror}{Yaw}{Meanabs}     &  \getErrorResult{Multicontact}{KoZpc}{Velerror}{EstimateXy}{Meanabs}    &   \getErrorResult{Multicontact}{KoZpc}{Velerror}{EstimateZ}{Meanabs}  \\ 
            (Ablation) &   \scalebox{0.9}{\scriptsize(\getErrorResult{Multicontact}{KoZpc}{Relerror}{Transxy}{Std})}   &   \scalebox{0.9}{\scriptsize(\getErrorResult{Multicontact}{KoZpc}{Relerror}{Transz}{Std})}    &     \scalebox{0.9}{\scriptsize(\getErrorResult{Multicontact}{KoZpc}{Relerror}{Tilt}{Std})}  &    \scalebox{0.9}{\scriptsize(\getErrorResult{Multicontact}{KoZpc}{Relerror}{Yaw}{Std})}   &  \scalebox{0.9}{\scriptsize(\getErrorResult{Multicontact}{KoZpc}{Velerror}{EstimateXy}{Std})}   &   \scalebox{0.9}{\scriptsize(\getErrorResult{Multicontact}{KoZpc}{Velerror}{EstimateZ}{Std})}  \\ 
            \midrule

            KO-PC       &     \getErrorResult{Multicontact}{Kowithoutwrenchsensors}{Relerror}{Transxy}{Meanabs}  &   \textbf{\getErrorResult{Multicontact}{Kowithoutwrenchsensors}{Relerror}{Transz}{Meanabs}}      &   \textbf{\getErrorResult{Multicontact}{Kowithoutwrenchsensors}{Relerror}{Tilt}{Meanabs}}     &      \textbf{\getErrorResult{Multicontact}{Kowithoutwrenchsensors}{Relerror}{Yaw}{Meanabs}}     &  \getErrorResult{Multicontact}{Kowithoutwrenchsensors}{Velerror}{EstimateXy}{Meanabs}    &   \getErrorResult{Multicontact}{Kowithoutwrenchsensors}{Velerror}{EstimateZ}{Meanabs}  \\ 
            (Ablation) &    \scalebox{0.9}{\scriptsize(\getErrorResult{Multicontact}{Kowithoutwrenchsensors}{Relerror}{Transxy}{Std})}   &     \scalebox{0.9}{\scriptsize(\getErrorResult{Multicontact}{Kowithoutwrenchsensors}{Relerror}{Transz}{Std})}     &   \scalebox{0.9}{\scriptsize(\getErrorResult{Multicontact}{Kowithoutwrenchsensors}{Relerror}{Tilt}{Std})}   &      \scalebox{0.9}{\scriptsize(\getErrorResult{Multicontact}{Kowithoutwrenchsensors}{Relerror}{Yaw}{Std})}   &  \scalebox{0.9}{\scriptsize(\getErrorResult{Multicontact}{Kowithoutwrenchsensors}{Velerror}{EstimateXy}{Std})}   &   \scalebox{0.9}{\scriptsize(\getErrorResult{Multicontact}{Kowithoutwrenchsensors}{Velerror}{EstimateZ}{Std})}  \\
            \midrule

            KO-Lin      &  \textbf{\getErrorResult{Multicontact}{Kolinear}{Relerror}{Transxy}{Meanabs}}  &   \textbf{\getErrorResult{Multicontact}{Kolinear}{Relerror}{Transz}{Meanabs}}   &   \textbf{\getErrorResult{Multicontact}{Kolinear}{Relerror}{Tilt}{Meanabs}}   &    \textbf{\getErrorResult{Multicontact}{Kolinear}{Relerror}{Yaw}{Meanabs}}   &  \getErrorResult{Multicontact}{Kolinear}{Velerror}{EstimateXy}{Meanabs}   &   \getErrorResult{Multicontact}{Kolinear}{Velerror}{EstimateZ}{Meanabs}  \\
            (Ablation) &   \scalebox{0.9}{\scriptsize(\getErrorResult{Multicontact}{Kolinear}{Relerror}{Transxy}{Std})}   &   \scalebox{0.9}{\scriptsize(\getErrorResult{Multicontact}{Kolinear}{Relerror}{Transz}{Std})}   &   \scalebox{0.9}{\scriptsize(\getErrorResult{Multicontact}{Kolinear}{Relerror}{Tilt}{Std})}   &   \scalebox{0.9}{\scriptsize(\getErrorResult{Multicontact}{Kolinear}{Relerror}{Yaw}{Std})}   &  \scalebox{0.9}{\scriptsize(\getErrorResult{Multicontact}{Kolinear}{Velerror}{EstimateXy}{Std})}   &   \scalebox{0.9}{\scriptsize(\getErrorResult{Multicontact}{Kolinear}{Velerror}{EstimateZ}{Std})}  \\ 
            \midrule

            \multirow{2}{*}{RI-EKF}     &  \getErrorResult{Multicontact}{Hartley}{Relerror}{Transxy}{Meanabs}  &    \getErrorResult{Multicontact}{Hartley}{Relerror}{Transz}{Meanabs}   &   \textbf{\getErrorResult{Multicontact}{Hartley}{Relerror}{Tilt}{Meanabs}}    &  \textbf{\getErrorResult{Multicontact}{Hartley}{Relerror}{Yaw}{Meanabs}}   &  \getErrorResult{Multicontact}{Hartley}{Velerror}{EstimateXy}{Meanabs}    &   \getErrorResult{Multicontact}{Hartley}{Velerror}{EstimateZ}{Meanabs}  \\ 
                        &     \scalebox{0.9}{\scriptsize(\getErrorResult{Multicontact}{Hartley}{Relerror}{Transxy}{Std})}  &     \scalebox{0.9}{\scriptsize(\getErrorResult{Multicontact}{Hartley}{Relerror}{Transz}{Std})}    &    \scalebox{0.9}{\scriptsize(\getErrorResult{Multicontact}{Hartley}{Relerror}{Tilt}{Std})}   &     \scalebox{0.9}{\scriptsize(\getErrorResult{Multicontact}{Hartley}{Relerror}{Yaw}{Std})}    &  \scalebox{0.9}{\scriptsize(\getErrorResult{Multicontact}{Hartley}{Velerror}{EstimateXy}{Std})}    &   \scalebox{0.9}{\scriptsize(\getErrorResult{Multicontact}{Hartley}{Velerror}{EstimateZ}{Std})}  \\ 
            \midrule

            \multirow{2}{*}{VALINOR}     &  \getErrorResult{Multicontact}{Tilt}{Relerror}{Transxy}{Meanabs}  &    \textbf{\getErrorResult{Multicontact}{Tilt}{Relerror}{Transz}{Meanabs}}   &   \getErrorResult{Multicontact}{Tilt}{Relerror}{Tilt}{Meanabs}    &  \getErrorResult{Multicontact}{Tilt}{Relerror}{Yaw}{Meanabs}   &  \textbf{\getErrorResult{Multicontact}{Tilt}{Velerror}{EstimateXy}{Meanabs}}     &   \textbf{\getErrorResult{Multicontact}{Tilt}{Velerror}{EstimateZ}{Meanabs}}  \\ 
                        &     \scalebox{0.9}{\scriptsize(\getErrorResult{Multicontact}{Tilt}{Relerror}{Transxy}{Std})}  &     \scalebox{0.9}{\scriptsize(\getErrorResult{Multicontact}{Tilt}{Relerror}{Transz}{Std})}    &    \scalebox{0.9}{\scriptsize(\getErrorResult{Multicontact}{Tilt}{Relerror}{Tilt}{Std})}   &     \scalebox{0.9}{\scriptsize(\getErrorResult{Multicontact}{Tilt}{Relerror}{Yaw}{Std})}    &  \scalebox{0.9}{\scriptsize(\getErrorResult{Multicontact}{Tilt}{Velerror}{EstimateXy}{Std})}    &   \scalebox{0.9}{\scriptsize(\getErrorResult{Multicontact}{Tilt}{Velerror}{EstimateZ}{Std})}  \\ 
            \bottomrule
        \end{tabularx}
\end{table}

\begin{figure}
    \centering
        \includegraphics[width=1\columnwidth]{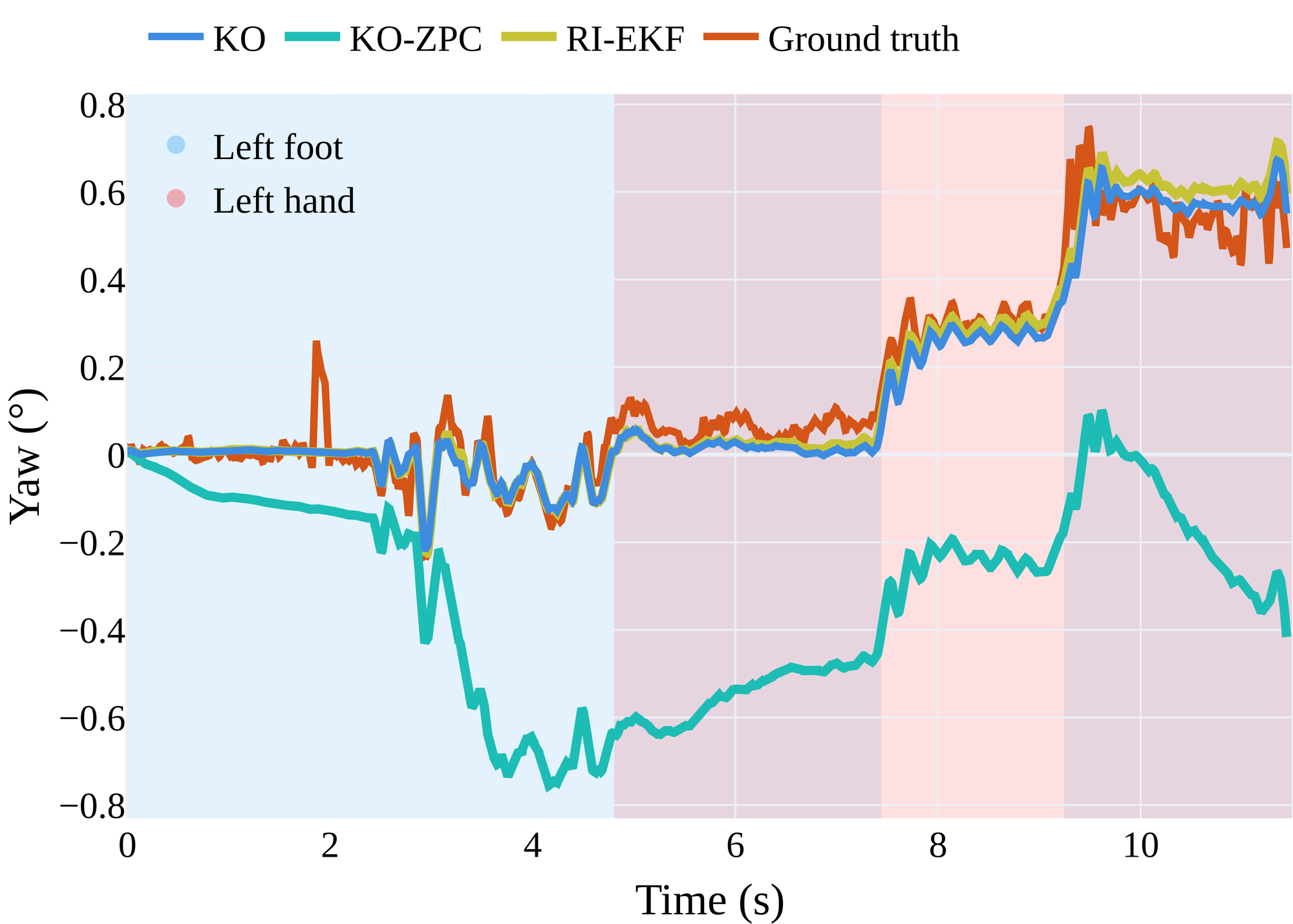}
        \caption{Yaw of the right foot of HRP-5P at the beginning of the multi-contact experiment. The colored areas represent the times the left foot and / or the left hand are in contact with the environment.}
        \label{fig:rightFootYaw}
\end{figure}

From the evaluation metrics presented in Table~\ref{tab:multicontact-odometry-results}, the Kinetics Observer and the RI-EKF are of comparable accuracy on this short-stride motion: over 0.3 m sub-trajectories all their errors stay within half a millimetre and a few hundredths of a degree of each other. But this experiment is particularly interesting because it showcases the improvements brought by the \emph{constrained contact process covariances} feature we introduced in Section~\ref{sec:Implementation}, as shown by the comparison between the Kinetics Observer and KO-ZPC, the variant that does not implement this feature. As a reminder, in the case of simultaneous multiple contacts, this feature enables long-term correction of the contact rest poses while ensuring that this correction cannot cause the robot to drift freely. In particular, the average error per 0.3 meter walked in its yaw estimate is significantly lower (by 66\%), and its tilt error is lower by 33\%. Figure~\ref{fig:rightFootYaw}, which shows the evolution of the right foot's yaw estimates and ground truth at the beginning of the experiment, allows to understand this better performance. This yaw is obtained by forward kinematics from the pose of the floating base, allowing to visualize how accurately the robot's yaw estimate reflects the actual behavior of the contacts.
From the beginning of the experiment, it can be observed that the yaw estimate of the KO-ZPC  starts to decrease and follow a similar pattern that quickly differs from the actual one. This mis-estimation is actually due to the slippage of the left foot and the right foot, that occurred at that time along the $\boldsymbol{x}$ axis, causing a change in their relative position to each other. In the absence of covariance on the process noise for the contact positions, the KO-ZPC was unable to correct them, and thus explained the change in relative position as a rotation of the floating base around the yaw axis.  
On the contrary, we observe that thanks to the \emph{constrained contact process covariances}, the Kinetics Observer succeeded in accurately tracking the pose of the contacts and thus the pose of the robot accurately, over the entire motion.

\subsection{Walking on Flat Ground}\label{subsec:flat-odometry}
\begin{figure}[ht]
    \centering
    \includegraphics[width=0.6\columnwidth]{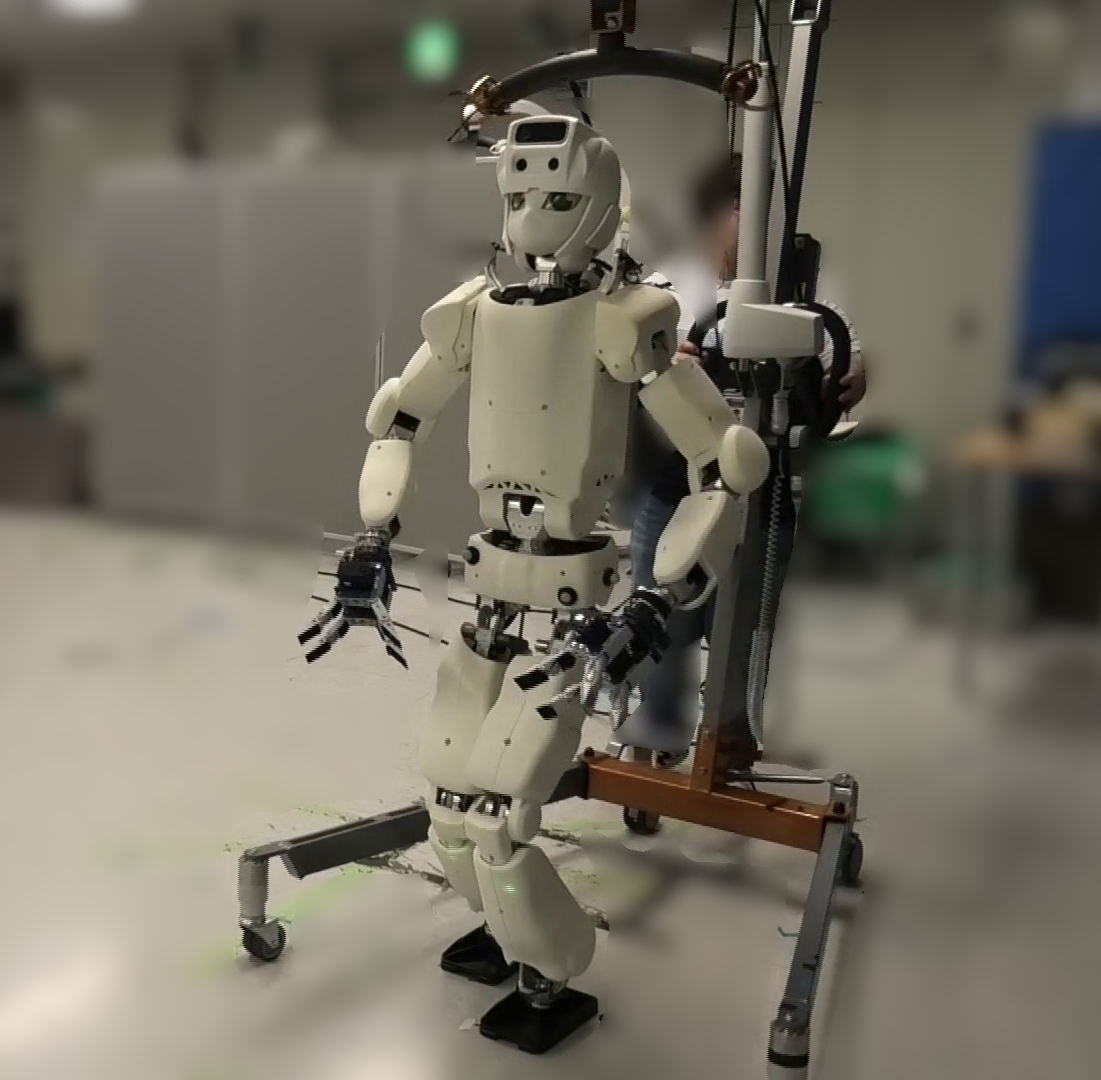}    
    \caption{RHP Friends.}
    \label{fig:rhp-friends}
\end{figure}

\begin{figure*}
    \centering
    \begin{subfigure}{0.47\textwidth}
        \centering
        \includegraphics[width=\linewidth]{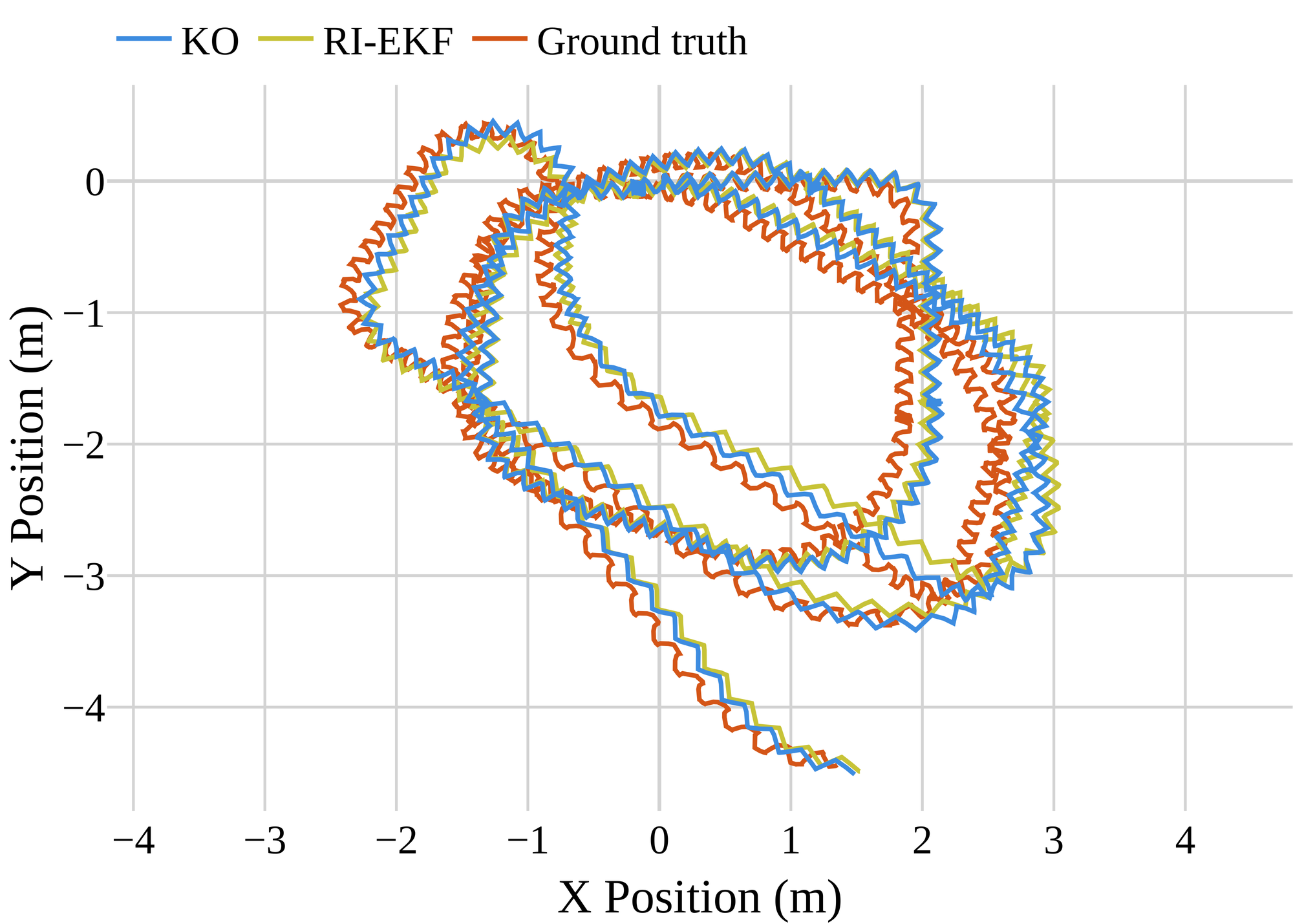}
        \caption{}
        \label{fig:hrp5_long_traj}
    \end{subfigure}
    \hfil
    \begin{subfigure}{0.47\textwidth}
        \centering
        \includegraphics[width=\linewidth]{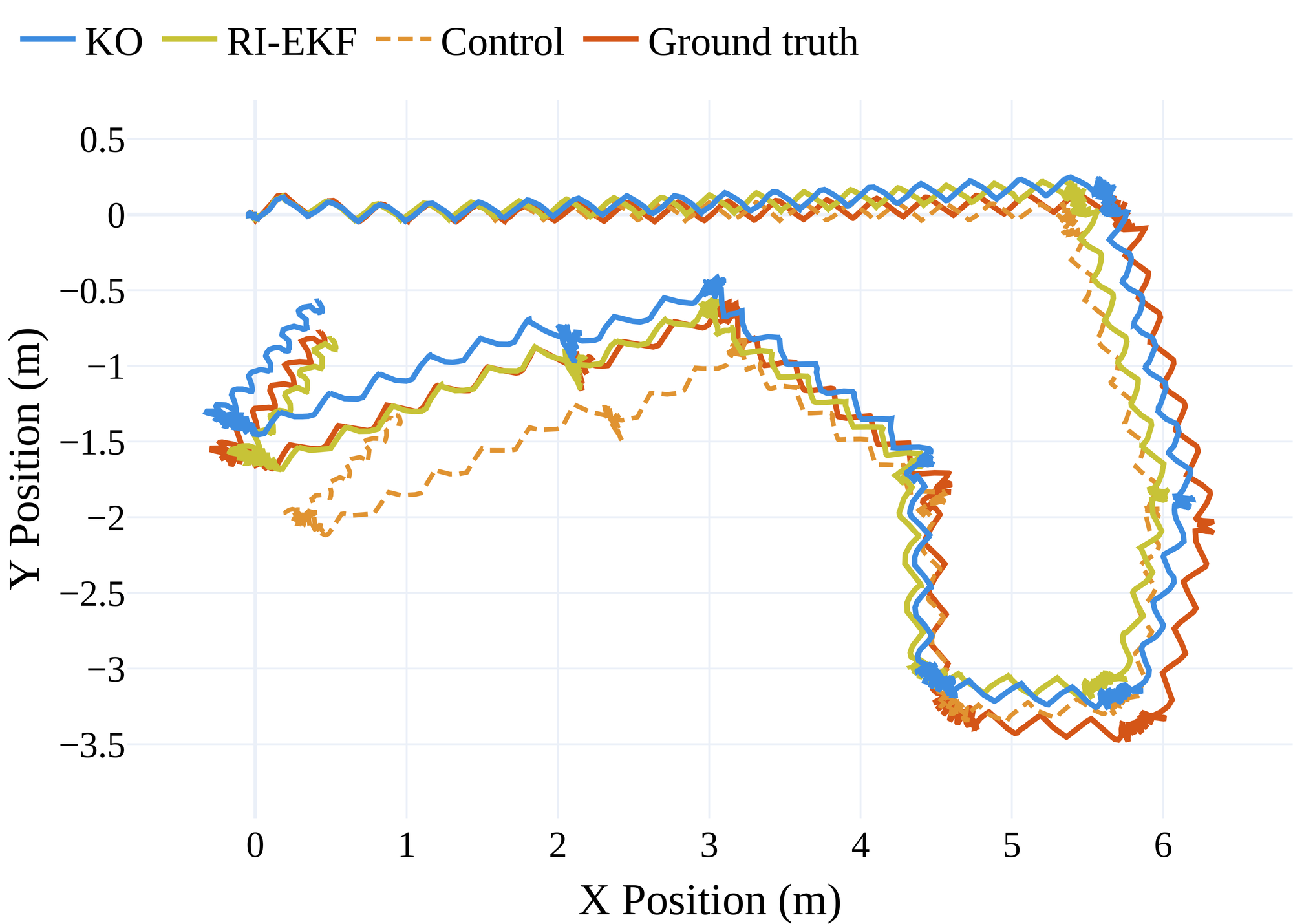}
        \caption{}
        \label{fig:flatOdom-traj}
    \end{subfigure}
    \caption{Top view of the ground truth and estimated trajectories of the robot's floating base during (a) the first 500 seconds of the long walk with HRP-5P and (b) the first walk with RHP Friends.  The robot starts at the point (0,0) and walks towards the $\boldsymbol{x}$ direction.  }
    \label{fig:flat-odom-rphs1}
\end{figure*}

\begin{figure*}[ht]
    \centering
    \includegraphics[width=1.0\textwidth]{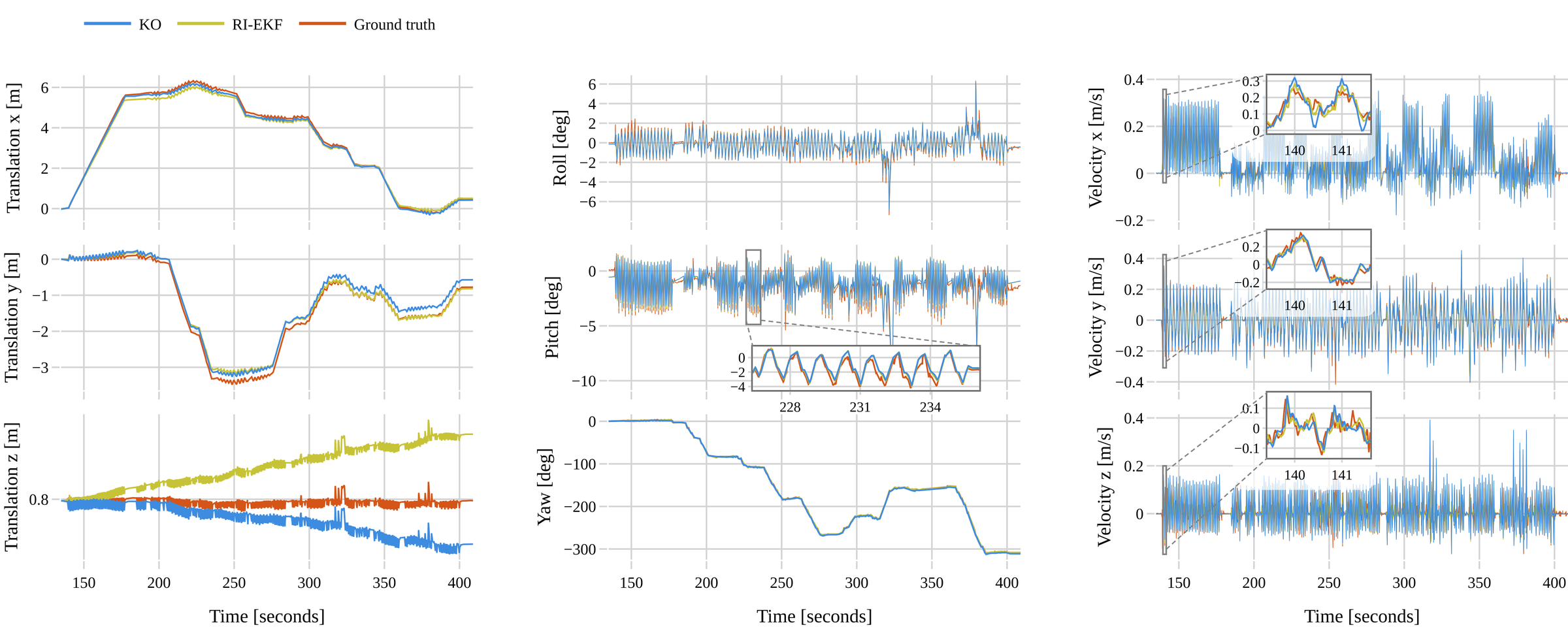}    
    \caption{Pose and linear velocity of RHP Friends estimated by the Kinetics Observer and the RI-EKF during the first walk on flat ground, compared to the ground truth obtained by motion capture. Areas of interest are highlighted by insets.}
    \label{fig:pose-vel-estimates-flat-odom}
\end{figure*}

\noindent Walking on flat ground was evaluated through two experiments. The first consisted of a 192-meter walk performed by the humanoid robot HRP-5P using a LIPM walking controller with reference footstep generation. The robot was remotely controlled to perform forward walking at 0.3 m/s, lateral motion, in-place turning, and curved turns over a duration of approximately 30 minutes. Since drift in legged robot proprioceptive odometry mainly occurs during contact creation and removal, the robot was commanded to keep stepping even when remaining in place, resulting in approximately 4180 contact detections. 
The second experiment was conducted on the humanoid robot RHP Friends (Figure~\ref{fig:rhp-friends}) across five sequences. Each trial involved walking and turning over approximately 18 meters (about 200 steps). Figs.~\ref{fig:hrp5_long_traj} and~\ref{fig:flatOdom-traj} illustrate the trajectories followed by the robot in both experiments, and Figure~\ref{fig:pose-vel-estimates-flat-odom} shows the estimated and ground-truth kinematics during the first trial of the RHP Friends experiment.
Evaluation metrics for both experiments are also summarized in Table~\ref{tab:flat-odometry-results}.

\begin{table}
    \footnotesize
    \centering
    \caption{Mean and standard deviation (in parentheses) of errors computed during the long walk with HRP-5P and the five trials of the walk with RHP Friends. The 10 m and the 1 m Relative Errors are represented, respectively. The best results for each metric are highlighted in bold.} \label{tab:flat-odometry-results}
    \begin{tabularx}{\columnwidth}{  C c c c c c c  }
            \toprule
   
            \multicolumn{7}{c}{\textbf{Long walk with HRP-5P}}  \\
            
            \midrule  

            \multirow{4}{*}{} 
            & \multicolumn{2}{c}{RE Trans.} 
            & \multicolumn{2}{c}{RE Ori.} 
            & \multicolumn{2}{c}{Lin. vel.} \\

            & \multicolumn{2}{c}{[m]} 
            & \multicolumn{2}{c}{[deg]} 
            & \multicolumn{2}{c}{$[\text{m.s}^{-1}]$} \\
           
            \addlinespace[0.2em]

            \cline{2-7}

            \addlinespace[0.2em]

                        &       $\left\{\boldsymbol{x}, \boldsymbol{y}\right\}$   &         $\boldsymbol{z}$    & $\text{Tilt}$ &    $\text{Yaw}$  &   $\left\{\boldsymbol{x}, \boldsymbol{y}\right\}$  &  $\boldsymbol{z}$  \\
            
            \midrule

            Kinetics       &  \getErrorResult{Longwalk}{Kineticsobserver}{Relerror}{Transxy}{Meanabs}   &   \getErrorResult{Longwalk}{Kineticsobserver}{Relerror}{Transz}{Meanabs}      &  \getErrorResult{Longwalk}{Kineticsobserver}{Relerror}{Tilt}{Meanabs}     &     \textbf{\getErrorResult{Longwalk}{Kineticsobserver}{Relerror}{Yaw}{Meanabs}}    &  \getErrorResult{Longwalk}{Kineticsobserver}{Velerror}{EstimateXy}{Meanabs}    &   \textbf{\getErrorResult{Longwalk}{Kineticsobserver}{Velerror}{EstimateZ}{Meanabs}}  \\ 

            Observer   &     \scalebox{0.9}{\scriptsize(\getErrorResult{Longwalk}{Kineticsobserver}{Relerror}{Transxy}{Std})}   &     \scalebox{0.9}{\scriptsize(\getErrorResult{Longwalk}{Kineticsobserver}{Relerror}{Transz}{Std})}    &       \scalebox{0.9}{\scriptsize(\getErrorResult{Longwalk}{Kineticsobserver}{Relerror}{Tilt}{Std})}   &       \scalebox{0.9}{\scriptsize(\getErrorResult{Longwalk}{Kineticsobserver}{Relerror}{Yaw}{Std})}    &  \scalebox{0.9}{\scriptsize(\getErrorResult{Longwalk}{Kineticsobserver}{Velerror}{EstimateXy}{Std})}    &   \scalebox{0.9}{\scriptsize(\getErrorResult{Longwalk}{Kineticsobserver}{Velerror}{EstimateZ}{Std})}     \\
            \midrule

            KO-ZPC       &   \textbf{\getErrorResult{Longwalk}{KoZpc}{Relerror}{Transxy}{Meanabs}}   &   \textbf{\getErrorResult{Longwalk}{KoZpc}{Relerror}{Transz}{Meanabs}}   &    \textbf{\getErrorResult{Longwalk}{KoZpc}{Relerror}{Tilt}{Meanabs}}     &       \getErrorResult{Longwalk}{KoZpc}{Relerror}{Yaw}{Meanabs}    &  \getErrorResult{Longwalk}{KoZpc}{Velerror}{EstimateXy}{Meanabs}    &   \textbf{\getErrorResult{Longwalk}{KoZpc}{Velerror}{EstimateZ}{Meanabs}} \\ 

            (Ablation) &     \scalebox{0.9}{\scriptsize(\getErrorResult{Longwalk}{KoZpc}{Relerror}{Transxy}{Std})}   &      \scalebox{0.9}{\scriptsize(\getErrorResult{Longwalk}{KoZpc}{Relerror}{Transz}{Std})}    &       \scalebox{0.9}{\scriptsize(\getErrorResult{Longwalk}{KoZpc}{Relerror}{Tilt}{Std})}    &    \scalebox{0.9}{\scriptsize(\getErrorResult{Longwalk}{KoZpc}{Relerror}{Yaw}{Std})}     &  \scalebox{0.9}{\scriptsize(\getErrorResult{Longwalk}{KoZpc}{Velerror}{EstimateXy}{Std})}    &   \scalebox{0.9}{\scriptsize(\getErrorResult{Longwalk}{KoZpc}{Velerror}{EstimateZ}{Std})}     \\
            \midrule

            KO-PC       &    \getErrorResult{Longwalk}{Kowithoutwrenchsensors}{Relerror}{Transxy}{Meanabs}   &    \getErrorResult{Longwalk}{Kowithoutwrenchsensors}{Relerror}{Transz}{Meanabs}   &    \textbf{\getErrorResult{Longwalk}{Kowithoutwrenchsensors}{Relerror}{Tilt}{Meanabs}}    &      \getErrorResult{Longwalk}{Kowithoutwrenchsensors}{Relerror}{Yaw}{Meanabs}  &  \textbf{\getErrorResult{Longwalk}{Kowithoutwrenchsensors}{Velerror}{EstimateXy}{Meanabs}}    &   \textbf{\getErrorResult{Longwalk}{Kowithoutwrenchsensors}{Velerror}{EstimateZ}{Meanabs}}  \\ 

            (Ablation) &    \scalebox{0.9}{\scriptsize(\getErrorResult{Longwalk}{Kowithoutwrenchsensors}{Relerror}{Transxy}{Std})}   &      \scalebox{0.9}{\scriptsize(\getErrorResult{Longwalk}{Kowithoutwrenchsensors}{Relerror}{Transz}{Std})}    &        \scalebox{0.9}{\scriptsize(\getErrorResult{Longwalk}{Kowithoutwrenchsensors}{Relerror}{Tilt}{Std})}   &   \scalebox{0.9}{\scriptsize(\getErrorResult{Longwalk}{Kowithoutwrenchsensors}{Relerror}{Yaw}{Std})}   &  \scalebox{0.9}{\scriptsize(\getErrorResult{Longwalk}{Kowithoutwrenchsensors}{Velerror}{EstimateXy}{Std})}   &   \scalebox{0.9}{\scriptsize(\getErrorResult{Longwalk}{Kowithoutwrenchsensors}{Velerror}{EstimateZ}{Std})}  \\
            \midrule

            KO-Lin      &  \getErrorResult{Longwalk}{Kolinear}{Relerror}{Transxy}{Meanabs}  &   \getErrorResult{Longwalk}{Kolinear}{Relerror}{Transz}{Meanabs}   &   \getErrorResult{Longwalk}{Kolinear}{Relerror}{Tilt}{Meanabs}   &    \getErrorResult{Longwalk}{Kolinear}{Relerror}{Yaw}{Meanabs}   &  \getErrorResult{Longwalk}{Kolinear}{Velerror}{EstimateXy}{Meanabs}   &   \textbf{\getErrorResult{Longwalk}{Kolinear}{Velerror}{EstimateZ}{Meanabs}}  \\
            (Ablation) &   \scalebox{0.9}{\scriptsize(\getErrorResult{Longwalk}{Kolinear}{Relerror}{Transxy}{Std})}   &   \scalebox{0.9}{\scriptsize(\getErrorResult{Longwalk}{Kolinear}{Relerror}{Transz}{Std})}   &   \scalebox{0.9}{\scriptsize(\getErrorResult{Longwalk}{Kolinear}{Relerror}{Tilt}{Std})}   &   \scalebox{0.9}{\scriptsize(\getErrorResult{Longwalk}{Kolinear}{Relerror}{Yaw}{Std})}   &  \scalebox{0.9}{\scriptsize(\getErrorResult{Longwalk}{Kolinear}{Velerror}{EstimateXy}{Std})}   &   \scalebox{0.9}{\scriptsize(\getErrorResult{Longwalk}{Kolinear}{Velerror}{EstimateZ}{Std})}  \\ 
            \midrule

            \multirow{2}{*}{RI-EKF}    &    \getErrorResult{Longwalk}{Hartley}{Relerror}{Transxy}{Meanabs}  &    \getErrorResult{Longwalk}{Hartley}{Relerror}{Transz}{Meanabs}   &     \getErrorResult{Longwalk}{Hartley}{Relerror}{Tilt}{Meanabs}     &     \getErrorResult{Longwalk}{Hartley}{Relerror}{Yaw}{Meanabs}   &  \getErrorResult{Longwalk}{Hartley}{Velerror}{EstimateXy}{Meanabs}    &   \textbf{\getErrorResult{Longwalk}{Hartley}{Velerror}{EstimateZ}{Meanabs}}  \\ 

            &    \scalebox{0.9}{\scriptsize(\getErrorResult{Longwalk}{Hartley}{Relerror}{Transxy}{Std})}   &      \scalebox{0.9}{\scriptsize(\getErrorResult{Longwalk}{Hartley}{Relerror}{Transz}{Std})}    &      \scalebox{0.9}{\scriptsize(\getErrorResult{Longwalk}{Hartley}{Relerror}{Tilt}{Std})}   &       \scalebox{0.9}{\scriptsize(\getErrorResult{Longwalk}{Hartley}{Relerror}{Yaw}{Std})}    &  \scalebox{0.9}{\scriptsize(\getErrorResult{Longwalk}{Hartley}{Velerror}{EstimateXy}{Std})}    &   \scalebox{0.9}{\scriptsize(\getErrorResult{Longwalk}{Hartley}{Velerror}{EstimateZ}{Std})}     \\

            \midrule

            \multirow{2}{*}{VALINOR}    &    \getErrorResult{Longwalk}{Tilt}{Relerror}{Transxy}{Meanabs}  &    \getErrorResult{Longwalk}{Tilt}{Relerror}{Transz}{Meanabs}   &     \getErrorResult{Longwalk}{Tilt}{Relerror}{Tilt}{Meanabs}     &     \getErrorResult{Longwalk}{Tilt}{Relerror}{Yaw}{Meanabs}   &   \getErrorResult{Longwalk}{Tilt}{Velerror}{EstimateXy}{Meanabs}     &   \textbf{\getErrorResult{Longwalk}{Tilt}{Velerror}{EstimateZ}{Meanabs}}  \\ 

            &    \scalebox{0.9}{\scriptsize(\getErrorResult{Longwalk}{Tilt}{Relerror}{Transxy}{Std})}   &      \scalebox{0.9}{\scriptsize(\getErrorResult{Longwalk}{Tilt}{Relerror}{Transz}{Std})}    &      \scalebox{0.9}{\scriptsize(\getErrorResult{Longwalk}{Tilt}{Relerror}{Tilt}{Std})}   &       \scalebox{0.9}{\scriptsize(\getErrorResult{Longwalk}{Tilt}{Relerror}{Yaw}{Std})}    &  \scalebox{0.9}{\scriptsize(\getErrorResult{Longwalk}{Tilt}{Velerror}{EstimateXy}{Std})}    &   \scalebox{0.9}{\scriptsize(\getErrorResult{Longwalk}{Tilt}{Velerror}{EstimateZ}{Std})}     \\
            
            \midrule
            \multicolumn{7}{c}{\textbf{Walking with RHP Friends}} \\
            
            \midrule
                        
            Kinetics       &  \textbf{\getErrorResult{Flatodometry}{Kineticsobserver}{Relerror}{Transxy}{Meanabs}}   &  \getErrorResult{Flatodometry}{Kineticsobserver}{Relerror}{Transz}{Meanabs}      &  \textbf{\getErrorResult{Flatodometry}{Kineticsobserver}{Relerror}{Tilt}{Meanabs}}    &      \textbf{\getErrorResult{Flatodometry}{Kineticsobserver}{Relerror}{Yaw}{Meanabs}}     &  \getErrorResult{Flatodometry}{Kineticsobserver}{Velerror}{EstimateXy}{Meanabs}    &   \getErrorResult{Flatodometry}{Kineticsobserver}{Velerror}{EstimateZ}{Meanabs}  \\ 

            Observer   &     \scalebox{0.9}{\scriptsize(\getErrorResult{Flatodometry}{Kineticsobserver}{Relerror}{Transxy}{Std})}   &     \scalebox{0.9}{\scriptsize(\getErrorResult{Flatodometry}{Kineticsobserver}{Relerror}{Transz}{Std})}    &       \scalebox{0.9}{\scriptsize(\getErrorResult{Flatodometry}{Kineticsobserver}{Relerror}{Tilt}{Std})}   &       \scalebox{0.9}{\scriptsize(\getErrorResult{Flatodometry}{Kineticsobserver}{Relerror}{Yaw}{Std})}    &  \scalebox{0.9}{\scriptsize(\getErrorResult{Flatodometry}{Kineticsobserver}{Velerror}{EstimateXy}{Std})}    &   \scalebox{0.9}{\scriptsize(\getErrorResult{Flatodometry}{Kineticsobserver}{Velerror}{EstimateZ}{Std})}     \\
            \midrule

            KO-ZPC       &   \getErrorResult{Flatodometry}{KoZpc}{Relerror}{Transxy}{Meanabs}  &   \getErrorResult{Flatodometry}{KoZpc}{Relerror}{Transz}{Meanabs}   &    \getErrorResult{Flatodometry}{KoZpc}{Relerror}{Tilt}{Meanabs}    &       \getErrorResult{Flatodometry}{KoZpc}{Relerror}{Yaw}{Meanabs}    &  \getErrorResult{Flatodometry}{KoZpc}{Velerror}{EstimateXy}{Meanabs}    &   \getErrorResult{Flatodometry}{KoZpc}{Velerror}{EstimateZ}{Meanabs} \\ 

            (Ablation) &     \scalebox{0.9}{\scriptsize(\getErrorResult{Flatodometry}{KoZpc}{Relerror}{Transxy}{Std})}   &      \scalebox{0.9}{\scriptsize(\getErrorResult{Flatodometry}{KoZpc}{Relerror}{Transz}{Std})}    &       \scalebox{0.9}{\scriptsize(\getErrorResult{Flatodometry}{KoZpc}{Relerror}{Tilt}{Std})}    &    \scalebox{0.9}{\scriptsize(\getErrorResult{Flatodometry}{KoZpc}{Relerror}{Yaw}{Std})}     &  \scalebox{0.9}{\scriptsize(\getErrorResult{Flatodometry}{KoZpc}{Velerror}{EstimateXy}{Std})}    &   \scalebox{0.9}{\scriptsize(\getErrorResult{Flatodometry}{KoZpc}{Velerror}{EstimateZ}{Std})}     \\
            \midrule

            KO-PC       &   \getErrorResult{Flatodometry}{Kowithoutwrenchsensors}{Relerror}{Transxy}{Meanabs}  &    \textbf{\getErrorResult{Flatodometry}{Kowithoutwrenchsensors}{Relerror}{Transz}{Meanabs}}   &    \textbf{\getErrorResult{Flatodometry}{Kowithoutwrenchsensors}{Relerror}{Tilt}{Meanabs}}    &      \getErrorResult{Flatodometry}{Kowithoutwrenchsensors}{Relerror}{Yaw}{Meanabs}  &  \getErrorResult{Flatodometry}{Kowithoutwrenchsensors}{Velerror}{EstimateXy}{Meanabs}    &   \getErrorResult{Flatodometry}{Kowithoutwrenchsensors}{Velerror}{EstimateZ}{Meanabs}  \\ 

            (Ablation) &    \scalebox{0.9}{\scriptsize(\getErrorResult{Flatodometry}{Kowithoutwrenchsensors}{Relerror}{Transxy}{Std})}   &      \scalebox{0.9}{\scriptsize(\getErrorResult{Flatodometry}{Kowithoutwrenchsensors}{Relerror}{Transz}{Std})}    &        \scalebox{0.9}{\scriptsize(\getErrorResult{Flatodometry}{Kowithoutwrenchsensors}{Relerror}{Tilt}{Std})}   &   \scalebox{0.9}{\scriptsize(\getErrorResult{Flatodometry}{Kowithoutwrenchsensors}{Relerror}{Yaw}{Std})}   &  \scalebox{0.9}{\scriptsize(\getErrorResult{Flatodometry}{Kowithoutwrenchsensors}{Velerror}{EstimateXy}{Std})}   &   \scalebox{0.9}{\scriptsize(\getErrorResult{Flatodometry}{Kowithoutwrenchsensors}{Velerror}{EstimateZ}{Std})}  \\
            \midrule

            KO-Lin      &  \getErrorResult{Flatodometry}{Kolinear}{Relerror}{Transxy}{Meanabs}  &   \getErrorResult{Flatodometry}{Kolinear}{Relerror}{Transz}{Meanabs}   &   \getErrorResult{Flatodometry}{Kolinear}{Relerror}{Tilt}{Meanabs}   &    \getErrorResult{Flatodometry}{Kolinear}{Relerror}{Yaw}{Meanabs}   &  \getErrorResult{Flatodometry}{Kolinear}{Velerror}{EstimateXy}{Meanabs}   &   \getErrorResult{Flatodometry}{Kolinear}{Velerror}{EstimateZ}{Meanabs}  \\
            (Ablation) &   \scalebox{0.9}{\scriptsize(\getErrorResult{Flatodometry}{Kolinear}{Relerror}{Transxy}{Std})}   &   \scalebox{0.9}{\scriptsize(\getErrorResult{Flatodometry}{Kolinear}{Relerror}{Transz}{Std})}   &   \scalebox{0.9}{\scriptsize(\getErrorResult{Flatodometry}{Kolinear}{Relerror}{Tilt}{Std})}   &   \scalebox{0.9}{\scriptsize(\getErrorResult{Flatodometry}{Kolinear}{Relerror}{Yaw}{Std})}   &  \scalebox{0.9}{\scriptsize(\getErrorResult{Flatodometry}{Kolinear}{Velerror}{EstimateXy}{Std})}   &   \scalebox{0.9}{\scriptsize(\getErrorResult{Flatodometry}{Kolinear}{Velerror}{EstimateZ}{Std})}  \\ 
            \midrule

            \multirow{2}{*}{RI-EKF}    &    \getErrorResult{Flatodometry}{Hartley}{Relerror}{Transxy}{Meanabs}  &    \getErrorResult{Flatodometry}{Hartley}{Relerror}{Transz}{Meanabs}   &     \textbf{\getErrorResult{Flatodometry}{Hartley}{Relerror}{Tilt}{Meanabs}}    &     \getErrorResult{Flatodometry}{Hartley}{Relerror}{Yaw}{Meanabs}   &  \textbf{\getErrorResult{Flatodometry}{Hartley}{Velerror}{EstimateXy}{Meanabs}}    &   \textbf{\getErrorResult{Flatodometry}{Hartley}{Velerror}{EstimateZ}{Meanabs}}  \\ 

            &    \scalebox{0.9}{\scriptsize(\getErrorResult{Flatodometry}{Hartley}{Relerror}{Transxy}{Std})}   &      \scalebox{0.9}{\scriptsize(\getErrorResult{Flatodometry}{Hartley}{Relerror}{Transz}{Std})}    &      \scalebox{0.9}{\scriptsize(\getErrorResult{Flatodometry}{Hartley}{Relerror}{Tilt}{Std})}   &       \scalebox{0.9}{\scriptsize(\getErrorResult{Flatodometry}{Hartley}{Relerror}{Yaw}{Std})}    &  \scalebox{0.9}{\scriptsize(\getErrorResult{Flatodometry}{Hartley}{Velerror}{EstimateXy}{Std})}    &   \scalebox{0.9}{\scriptsize(\getErrorResult{Flatodometry}{Hartley}{Velerror}{EstimateZ}{Std})}     \\ 
            \bottomrule   
        \end{tabularx}
\end{table}

We can first conclude from these results that the Kinetics Observer improved proprioceptive odometry compared with the state-of-the-art RI-EKF. It indeed reduced the drift in lateral translations by 40\% throughout the walk with RHP Friends, while during the long walk with HRP-5P the two estimators drifted by the same amount. In the latter experiment, it reduced the drift in yaw by 49\%. We attribute this better accuracy in part to the additional correction of the odometry by contact wrench measurements through the visco-elastic model.  

In both experiments, the Kinetics Observer and the RI-EKF showed drift in vertical translation of similar magnitude (3.7 cm vs. 3.6 cm on average over 10 m in the first experiment, and about 0.5 cm per meter for both in the second). Additionally, they showed comparable performance for tilt estimation, with an advantage for the Kinetics Observer over the long walk with HRP-5P, and for velocity estimation, with an advantage for the RI-EKF over the walk with RHP Friends.

\subsection{Walking with Slipping Contacts}\label{subsec:Slipping-odometry}

\begin{figure*}
    \centering
    \begin{subfigure}{0.36\textwidth}
        \centering
        \includegraphics[width=\linewidth]{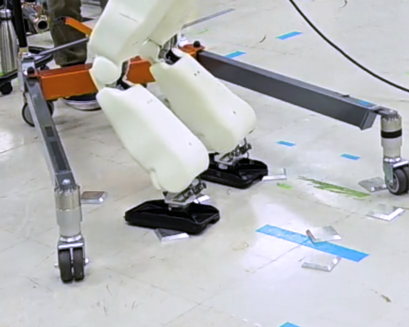}
        \caption{}
        \label{fig:slippery-tiles}
    \end{subfigure}
    \hfil
    \begin{subfigure}{0.47\textwidth}
        \centering
        \includegraphics[width=\linewidth]{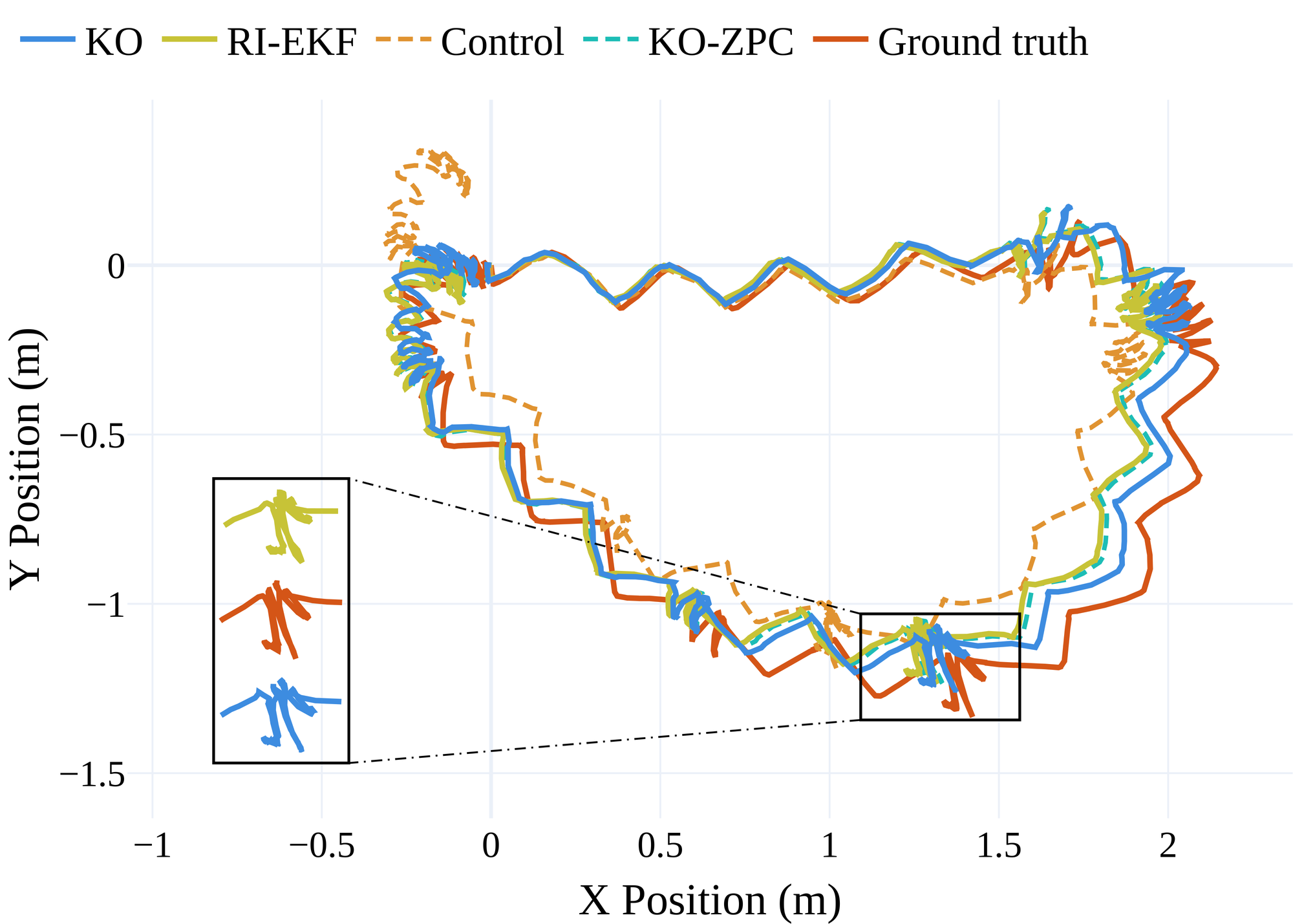}
        \caption{}
        \label{fig:slippingOdom-traj}
    \end{subfigure}
    \caption{(a) Slippery tiles RHP Friends walked over during the experiment. 
    (b) Top view of the ground truth and estimated trajectories of the robot's floating base during the first experiment of walking with slipping contacts. The robot starts at the point (0,0) and walks towards the x direction.  A zoomed-in view of the trajectory during slippage is shown in the inset. The trajectories in the inset are separated to improve visibility.}
    \label{fig:slipping-odom-rphs1}
\end{figure*}

\noindent The purpose of this scenario is to assess the robustness of our estimation against slipping contacts. To this end, tiles with dimensions 7.5~(l)~$\times$~7.5~(w)~$\times$~1~(h) cm, which have low friction coefficients with both the robot's feet and the ground, are placed along the robot's path, as shown in Figure~\ref{fig:slippery-tiles}. Over three experiments and a distance of about 5 meters, the robot walks and turns on the tiles, causing slippage in translation and rotation. The trajectory walked by the robot is shown in Figure~\ref{fig:slippingOdom-traj}. Note that we did not further increase the slipperiness of the environment for two reasons. First, maintaining the robot's balance would have become significantly more challenging to the walk controller. Second, the drift caused by more significant slippage is not observable and would have increased the resulting variability across experiments.

Based on the evaluation metrics summarized in Table~\ref{tab:slipping-odometry-results}, and as observed in the previous experiments of walking on flat ground (Section~\ref{subsec:flat-odometry}), the Kinetics Observer was able to track the robot's position more accurately than the RI-EKF, by 39\%. These results also highlight the proposed method's greater robustness
  to slipping contacts in lateral translation estimation compared to the RI-EKF. Indeed, when compared to the results of the walk with RHP Friends, the average Relative Error in the lateral translation estimate over a 1 m translation increased by 0.23 cm for the Kinetics Observer, against 0.36 cm for the RI-EKF. We note that in this experiment, the RI-EKF estimated the yaw slightly better by 4\%. Nevertheless, the orientation estimates of both methods, are barely affected by the slippage: the mean Relative Error on the yaw estimate increases by less than 0.07$^{\circ}$ per meter walked for either of them.

\begin{table}
    \footnotesize
    \centering 
    \caption{Mean and standard deviation (in parentheses) of errors computed during walking with slipping contacts. The 1 m Relative Error is represented. The best results for each metric are highlighted in bold.} 
    \label{tab:slipping-odometry-results}
        \begin{tabularx}{\columnwidth}{ C c c c c c c  }
            \toprule
            \multirow{4}{*}{} 
            & \multicolumn{2}{c}{RE Trans.} 
            & \multicolumn{2}{c}{RE Ori.} 
            & \multicolumn{2}{c}{Lin. vel.} \\

            & \multicolumn{2}{c}{[m]} 
            & \multicolumn{2}{c}{[deg]} 
            & \multicolumn{2}{c}{$[\text{m.s}^{-1}]$} \\   

            \addlinespace[0.2em]

            \cline{2-7}

            \addlinespace[0.2em]

            &       $\left\{\boldsymbol{x}, \boldsymbol{y}\right\}$   &         $\boldsymbol{z}$    & $\text{Tilt}$ &    $\text{Yaw}$  &   $\left\{\boldsymbol{x}, \boldsymbol{y}\right\}$  &  $\boldsymbol{z}$  \\
            \midrule 
            Kinetics      &    \getErrorResult{Slippingodometry}{Kineticsobserver}{Relerror}{Transxy}{Meanabs}   &  \getErrorResult{Slippingodometry}{Kineticsobserver}{Relerror}{Transz}{Meanabs}   &  \getErrorResult{Slippingodometry}{Kineticsobserver}{Relerror}{Tilt}{Meanabs}    &       \getErrorResult{Slippingodometry}{Kineticsobserver}{Relerror}{Yaw}{Meanabs}     &  \getErrorResult{Slippingodometry}{Kineticsobserver}{Velerror}{EstimateXy}{Meanabs}    &   \getErrorResult{Slippingodometry}{Kineticsobserver}{Velerror}{EstimateZ}{Meanabs}  \\ 

            Observer    &     \scalebox{0.9}{\scriptsize(\getErrorResult{Slippingodometry}{Kineticsobserver}{Relerror}{Transxy}{Std})}   &      \scalebox{0.9}{\scriptsize(\getErrorResult{Slippingodometry}{Kineticsobserver}{Relerror}{Transz}{Std})}    &    \scalebox{0.9}{\scriptsize(\getErrorResult{Slippingodometry}{Kineticsobserver}{Relerror}{Tilt}{Std})}   &        \scalebox{0.9}{\scriptsize(\getErrorResult{Slippingodometry}{Kineticsobserver}{Relerror}{Yaw}{Std})}     &  \scalebox{0.9}{\scriptsize(\getErrorResult{Slippingodometry}{Kineticsobserver}{Velerror}{EstimateXy}{Std})}    &   \scalebox{0.9}{\scriptsize(\getErrorResult{Slippingodometry}{Kineticsobserver}{Velerror}{EstimateZ}{Std})}     \\
            \midrule 

            KO-ZPC      &    \getErrorResult{Slippingodometry}{KoZpc}{Relerror}{Transxy}{Meanabs}   &  \getErrorResult{Slippingodometry}{KoZpc}{Relerror}{Transz}{Meanabs}   &   \getErrorResult{Slippingodometry}{KoZpc}{Relerror}{Tilt}{Meanabs}    &      \getErrorResult{Slippingodometry}{KoZpc}{Relerror}{Yaw}{Meanabs}  &  \getErrorResult{Slippingodometry}{KoZpc}{Velerror}{EstimateXy}{Meanabs}    &   \getErrorResult{Slippingodometry}{KoZpc}{Velerror}{EstimateZ}{Meanabs}  \\ 
            (Ablation) &     \scalebox{0.9}{\scriptsize(\getErrorResult{Slippingodometry}{KoZpc}{Relerror}{Transxy}{Std})}   &     \scalebox{0.9}{\scriptsize(\getErrorResult{Slippingodometry}{KoZpc}{Relerror}{Transz}{Std})}    &    \scalebox{0.9}{\scriptsize(\getErrorResult{Slippingodometry}{KoZpc}{Relerror}{Tilt}{Std})}   &       \scalebox{0.9}{\scriptsize(\getErrorResult{Slippingodometry}{KoZpc}{Relerror}{Yaw}{Std})}  &  \scalebox{0.9}{\scriptsize(\getErrorResult{Slippingodometry}{KoZpc}{Velerror}{EstimateXy}{Std})}    &   \scalebox{0.9}{\scriptsize(\getErrorResult{Slippingodometry}{KoZpc}{Velerror}{EstimateZ}{Std})}     \\
            \midrule 

            KO-PC       &     \getErrorResult{Slippingodometry}{Kowithoutwrenchsensors}{Relerror}{Transxy}{Meanabs}  &     \textbf{\getErrorResult{Slippingodometry}{Kowithoutwrenchsensors}{Relerror}{Transz}{Meanabs}}   &   \textbf{\getErrorResult{Slippingodometry}{Kowithoutwrenchsensors}{Relerror}{Tilt}{Meanabs}}    &     \getErrorResult{Slippingodometry}{Kowithoutwrenchsensors}{Relerror}{Yaw}{Meanabs}  &  \getErrorResult{Slippingodometry}{Kowithoutwrenchsensors}{Velerror}{EstimateXy}{Meanabs}    &   \getErrorResult{Slippingodometry}{Kowithoutwrenchsensors}{Velerror}{EstimateZ}{Meanabs}  \\ 
            (Ablation) &     \scalebox{0.9}{\scriptsize(\getErrorResult{Slippingodometry}{Kowithoutwrenchsensors}{Relerror}{Transxy}{Std})}   &     \scalebox{0.9}{\scriptsize(\getErrorResult{Slippingodometry}{Kowithoutwrenchsensors}{Relerror}{Transz}{Std})}    &      \scalebox{0.9}{\scriptsize(\getErrorResult{Slippingodometry}{Kowithoutwrenchsensors}{Relerror}{Tilt}{Std})}   &     \scalebox{0.9}{\scriptsize(\getErrorResult{Slippingodometry}{Kowithoutwrenchsensors}{Relerror}{Yaw}{Std})}   &  \scalebox{0.9}{\scriptsize(\getErrorResult{Slippingodometry}{Kowithoutwrenchsensors}{Velerror}{EstimateXy}{Std})}   &   \scalebox{0.9}{\scriptsize(\getErrorResult{Slippingodometry}{Kowithoutwrenchsensors}{Velerror}{EstimateZ}{Std})}  \\
            \midrule

            KO-Lin      &  \textbf{\getErrorResult{Slippingodometry}{Kolinear}{Relerror}{Transxy}{Meanabs}}  &   \getErrorResult{Slippingodometry}{Kolinear}{Relerror}{Transz}{Meanabs}   &   \getErrorResult{Slippingodometry}{Kolinear}{Relerror}{Tilt}{Meanabs}   &    \getErrorResult{Slippingodometry}{Kolinear}{Relerror}{Yaw}{Meanabs}   &  \getErrorResult{Slippingodometry}{Kolinear}{Velerror}{EstimateXy}{Meanabs}   &   \getErrorResult{Slippingodometry}{Kolinear}{Velerror}{EstimateZ}{Meanabs}  \\
            (Ablation) &   \scalebox{0.9}{\scriptsize(\getErrorResult{Slippingodometry}{Kolinear}{Relerror}{Transxy}{Std})}   &   \scalebox{0.9}{\scriptsize(\getErrorResult{Slippingodometry}{Kolinear}{Relerror}{Transz}{Std})}   &   \scalebox{0.9}{\scriptsize(\getErrorResult{Slippingodometry}{Kolinear}{Relerror}{Tilt}{Std})}   &   \scalebox{0.9}{\scriptsize(\getErrorResult{Slippingodometry}{Kolinear}{Relerror}{Yaw}{Std})}   &  \scalebox{0.9}{\scriptsize(\getErrorResult{Slippingodometry}{Kolinear}{Velerror}{EstimateXy}{Std})}   &   \scalebox{0.9}{\scriptsize(\getErrorResult{Slippingodometry}{Kolinear}{Velerror}{EstimateZ}{Std})}  \\ 
            \midrule 
                        
            \multirow{2}{*}{RI-EKF}    &    \getErrorResult{Slippingodometry}{Hartley}{Relerror}{Transxy}{Meanabs}   &   \getErrorResult{Slippingodometry}{Hartley}{Relerror}{Transz}{Meanabs}    &    \getErrorResult{Slippingodometry}{Hartley}{Relerror}{Tilt}{Meanabs}     &     \textbf{\getErrorResult{Slippingodometry}{Hartley}{Relerror}{Yaw}{Meanabs}}     &   \textbf{\getErrorResult{Slippingodometry}{Hartley}{Velerror}{EstimateXy}{Meanabs}}    &   \textbf{\getErrorResult{Slippingodometry}{Hartley}{Velerror}{EstimateZ}{Meanabs}}  \\ 
            &     \scalebox{0.9}{\scriptsize(\getErrorResult{Slippingodometry}{Hartley}{Relerror}{Transxy}{Std})}   &      \scalebox{0.9}{\scriptsize(\getErrorResult{Slippingodometry}{Hartley}{Relerror}{Transz}{Std})}    &       \scalebox{0.9}{\scriptsize(\getErrorResult{Slippingodometry}{Hartley}{Relerror}{Tilt}{Std})}   &     \scalebox{0.9}{\scriptsize(\getErrorResult{Slippingodometry}{Hartley}{Relerror}{Yaw}{Std})}     &   \scalebox{0.9}{\scriptsize(\getErrorResult{Slippingodometry}{Hartley}{Velerror}{EstimateXy}{Std})}    &   \scalebox{0.9}{\scriptsize(\getErrorResult{Slippingodometry}{Hartley}{Velerror}{EstimateZ}{Std})}     \\

            \midrule 
                        
            \multirow{2}{*}{VALINOR}    &    \getErrorResult{Slippingodometry}{Tilt}{Relerror}{Transxy}{Meanabs}   &   \getErrorResult{Slippingodometry}{Tilt}{Relerror}{Transz}{Meanabs}    &    \getErrorResult{Slippingodometry}{Tilt}{Relerror}{Tilt}{Meanabs}     &     \getErrorResult{Slippingodometry}{Tilt}{Relerror}{Yaw}{Meanabs}     &  \getErrorResult{Slippingodometry}{Tilt}{Velerror}{EstimateXy}{Meanabs}     &   \getErrorResult{Slippingodometry}{Tilt}{Velerror}{EstimateZ}{Meanabs}   \\ 
            &     \scalebox{0.9}{\scriptsize(\getErrorResult{Slippingodometry}{Tilt}{Relerror}{Transxy}{Std})}   &      \scalebox{0.9}{\scriptsize(\getErrorResult{Slippingodometry}{Tilt}{Relerror}{Transz}{Std})}    &       \scalebox{0.9}{\scriptsize(\getErrorResult{Slippingodometry}{Tilt}{Relerror}{Tilt}{Std})}   &     \scalebox{0.9}{\scriptsize(\getErrorResult{Slippingodometry}{Tilt}{Relerror}{Yaw}{Std})}     &   \scalebox{0.9}{\scriptsize(\getErrorResult{Slippingodometry}{Tilt}{Velerror}{EstimateXy}{Std})}    &   \scalebox{0.9}{\scriptsize(\getErrorResult{Slippingodometry}{Tilt}{Velerror}{EstimateZ}{Std})}     \\
            \bottomrule
        \end{tabularx}
\end{table}

\subsection{Robustness to a Drifting Gyrometer Bias}\label{subsec:gyro-bias-drift}

\noindent In this section, we evaluate the robustness of both the Kinetics Observer and the RI-EKF to a drift of the gyrometer bias. The experiments introduced above are conducted on robots with high-end gyrometers with low noise and bias instability. In particular, the gyrometer of HRP-5P is a fiber-optic unit, and despite running for about 50 minutes during the experiment of long walk, its bias remained nearly constant at about $3\cdot10^{-4}$\,deg/s. Most legged robots, however, carry MEMS inertial units whose bias drifts by orders of magnitude more over a run of this duration. 
  
To evaluate the robustness of the estimators to the drift in the bias of a mid-range MEMS unit, we simulated the bias evolution in a Bosch BMI088 IMU based on its datasheet characteristics. The drift is composed of a thermal ramp, assuming a 12 K warm-up and a temperature compensation of 85\%, a bias instability of 2 deg/h and a rate random walk of 1 deg/h/$\sqrt{\text{h}}$. We added this drift to the gyrometer signal measured by HRP-5P before feeding it to the estimators, which kept the tuning used throughout this paper.   The injected yaw bias grows to 0.021 deg/s over the walk.  We also evaluate two variants of the RI-EKF, to give it the best chance of tracking the drift. The first has its bias process covariance matched to the injected drift. The second, which we name RI-EKF-ZPC, raises the initial yaw bias variance  from $10^{-8}$ to $10^{-6}$ so that it can absorb the bias better from the start, and  implements the constrained contact process covariances of Section~\ref{sec:Implementation}. It therefore features the same constraint on the contact positions as the KO-Lin variant, which highlights the contribution of the linear viscoelastic model and the dynamic model of the Kinetics Observer to the bias estimation, especially during the long initial double support phase.

\begin{table}
    \footnotesize
    \centering
    \caption{Mean of the 10 m Relative Errors  computed during the long walk with a simulated BMI088 gyrometer drift. }
    \label{tab:gyro-bias-drift}
        \newcolumntype{Y}{>{\centering\arraybackslash}X}
        \begin{tabularx}{\columnwidth}{ Y c c c c }
            \toprule
            \multirow{4}{*}{}
            & \multicolumn{2}{c}{RE Trans.}
            & \multicolumn{2}{c}{RE Ori.} \\

            & \multicolumn{2}{c}{[m]}
            & \multicolumn{2}{c}{[deg]} \\

            \addlinespace[0.2em]

            \cline{2-5}

            \addlinespace[0.2em]

            &       $\left\{\boldsymbol{x}, \boldsymbol{y}\right\}$   &         $\boldsymbol{z}$    & $\text{Tilt}$ &    $\text{Yaw}$  \\
            \midrule
            Kinetics Observer   &  0.189  &  0.039  &   0.556  &   0.498   \\
            \midrule

            KO-Lin (Ablation)   &  0.187  &  0.039  &  0.554  &  0.622  \\
            \midrule

            RI-EKF   &  0.190  &  0.033  &  0.969  &  0.935  \\
            \midrule

            RI-EKF-ZPC   &  0.786  &  0.037  &  1.209  &  23.33  \\
            \midrule

            RI-EKF, adapted bias process   &   0.184   &  0.035  &  0.787  &  1.972  \\
            \bottomrule
        \end{tabularx}
\end{table}

\begin{figure}[h]
    \centering
    \includegraphics[width=1\columnwidth]{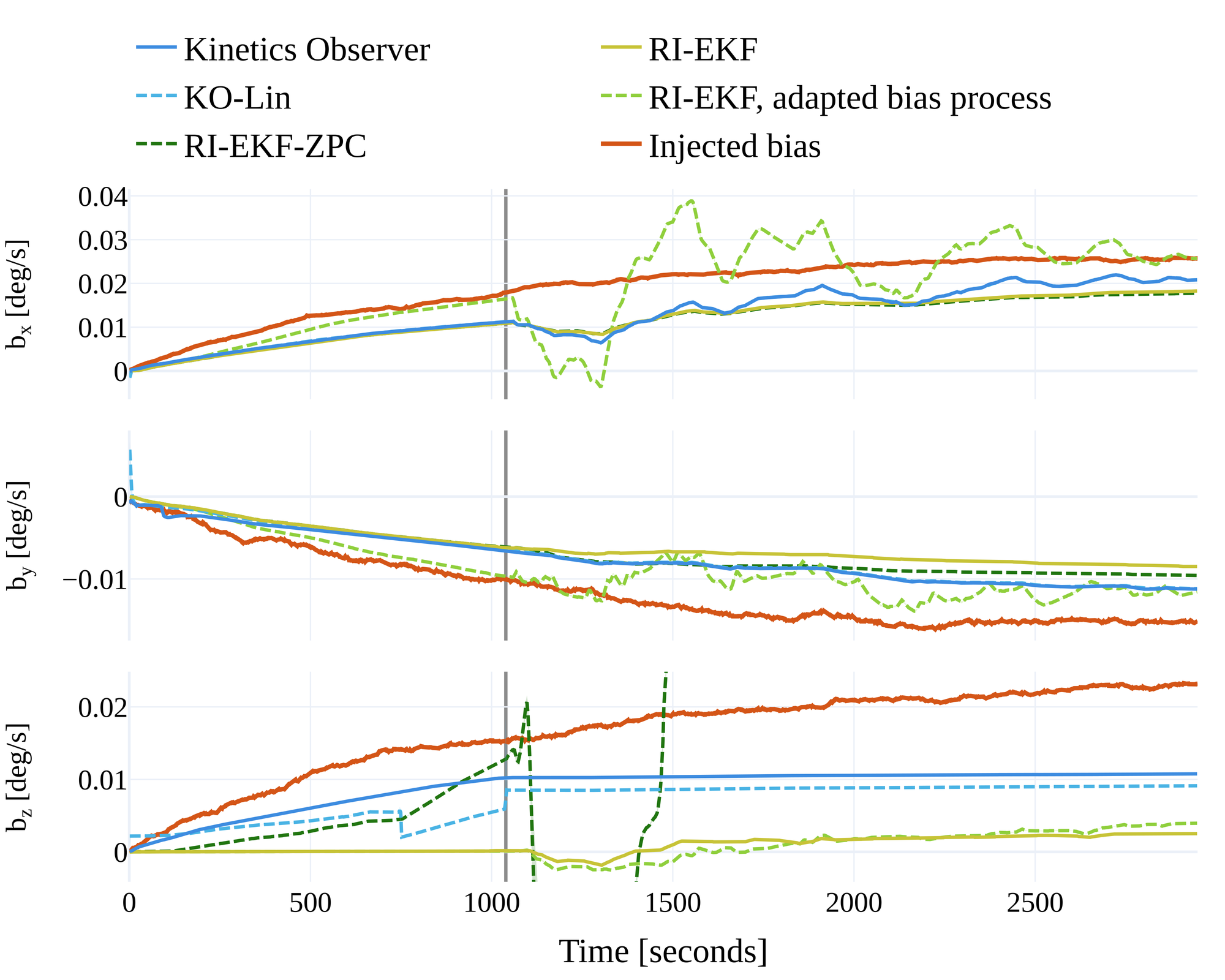}
    \caption{Injected and estimated gyrometer bias during the long walk with HRP-5P. The vertical grey line marks the first foot lift-off at the start of the walk.  The bias along the yaw axis estimated by the RI-EKF-ZPC erroneously fluctuates to much higher values than the estimates of the other estimators. The range of the bias value axis is nevertheless adapted to the latter so they are not compressed and the plot remains readable. }
    \label{fig:injectedGyroBias}
\end{figure}

Figure~\ref{fig:injectedGyroBias} compares the evolution of the estimated biases with the injected bias. All estimators track the horizontal bias   components relatively well, as expected from the observability of tilt. However, their yaw-axis bias estimates behave differently. Despite the initial 1000~s standing phase, the RI-EKF is unable to track the yaw-axis bias: its estimate remains close to zero throughout this phase. Instead, it interprets the injected bias as a yaw rotation of the base accompanied by translations of the feet. This behavior stems from the invariance of its forward-kinematics measurement under a common yaw rotation of the base and feet. The measurement therefore cannot distinguish a gyrometer bias from a yaw rotation of the base with the feet slipping around it.
Meanwhile, the Kinetics Observer uses the additional information provided by its dynamics, viscoelastic contact model, and wrench measurements to better track the injected bias and limit yaw drift as shown in Table~\ref{tab:gyro-bias-drift}. One can also observe that the KO-Lin variant, which, like the RI-EKF, uses neither contact orientations nor moment measurements, also tracks the bias drift better and displays a lower drift in yaw.  Its better robustness therefore highlights the benefit of using the robot dynamics and the correction provided by contact force   measurements through the linear viscoelastic model. Interestingly, the KO-Lin yaw-axis bias estimate changes abruptly at   two events during and at the end of the initial double-support phase, when the contacts change for the first time after the long double-support phase.   These changes appear to coincide with sudden changes in the   forward-kinematics measurement: one during standing, and the other   when the foot is released at the start of the walk. The Kinetics   Observer itself is barely affected by these events, highlighting   the additional robustness provided by the angular viscoelastic   model and contact moment measurements.

The RI-EKF-ZPC variant increases the initial variance on the yaw-axis bias and constrains the foot positions as in the Kinetics Observer. This allows it to estimate the bias more effectively during the double-support phase. However, the larger variance allows the bias estimate to explain and absorb too much of the error in the system. This occurs at the same events as for KO-Lin, but becomes much more problematic  during walking, when the bias estimate becomes highly erratic and consequently degrades the yaw estimate.

Increasing the initial bias variance therefore degrades the estimation. As shown by the second RI-EKF variant, adapting the bias process variance to the injected drift has the same adverse effect and does not improve yaw bias recovery. We conclude that, although low initial and process variances may limit the accuracy of gyrometer bias tracking, they are necessary to prevent the overall estimation from deteriorating. Despite extensive tuning, we could not find RI-EKF parameters that achieved both accurate bias tracking and good odometry performance. The same trade-off applies to the Kinetics Observer and explains why its yaw-axis bias estimate plateaus during walking, as the estimated disturbance wrench then absorbs most of the model error.

To conclude this section, we have shown that the additional redundancy   provided by the viscoelastic model, robot dynamics, and wrench measurements improves the Kinetics Observer's robustness to modeling errors and   gyrometer bias drift. It even allows the estimator to track the yaw-axis   bias reasonably well during double-support phases, which the RI-EKF cannot do without degrading its overall estimation.
 
\subsection{Contact Geometry Estimation under Initialization Errors} \label{subsec:contact-char-estimation}
\noindent As explained in Section~\ref{subsec:Viscoelastic-model}, the Kinetics Observer is able to estimate the pose of the contact \emph{rest frame}. The rest frame is attached to the environment and corresponds to its local geometry at the contact when no force is applied to it (and at any time when considering a perfectly rigid environment). It thus differs from the \emph{contact frame} and from the usual definition of contacts used in conventional Leg-Inertial odometry (e.g.~\citet{Rotella2014StateEstimationForHumanoid}). The latter is attached to the robot, and its kinematics are obtained by forward kinematics from those of the base. These kinematics are therefore directly affected by the robot's internal flexibilities (see top of Figure~\ref{fig:Viscoelastic-model}) and may not accurately match those of the environment. In addition, the leg kinematic measurement is a core component of conventional Leg-Inertial odometry methods, making them highly sensitive to errors in the initialization of the contact pose. Thanks to the viscoelastic model, the Kinetics Observer can correct the contact rest pose using contact wrench measurements, greatly enhancing robustness to initialization errors.

To demonstrate this robustness, we show that the Kinetics Observer is able to recover the actual contact rest orientation during the multi-contact motion of Section~\ref{subsec:Multi-contact-with-tilted}, despite a considerable initial error. Each contact created during the motion has its initial rest orientation rotated by $30^\circ$ about a randomly drawn axis. The estimation is compared against the ground-truth contact orientation measured using a precision instrument. We also compare it with the orientation obtained by forward kinematics from the ground-truth orientation of the floating base. This provides a baseline to assess the error in the ground orientation when the contact frame obtained from floating base estimators is used as an estimate of the environment geometry. 

\begin{figure}
    \centering
    \includegraphics[width=1\columnwidth]{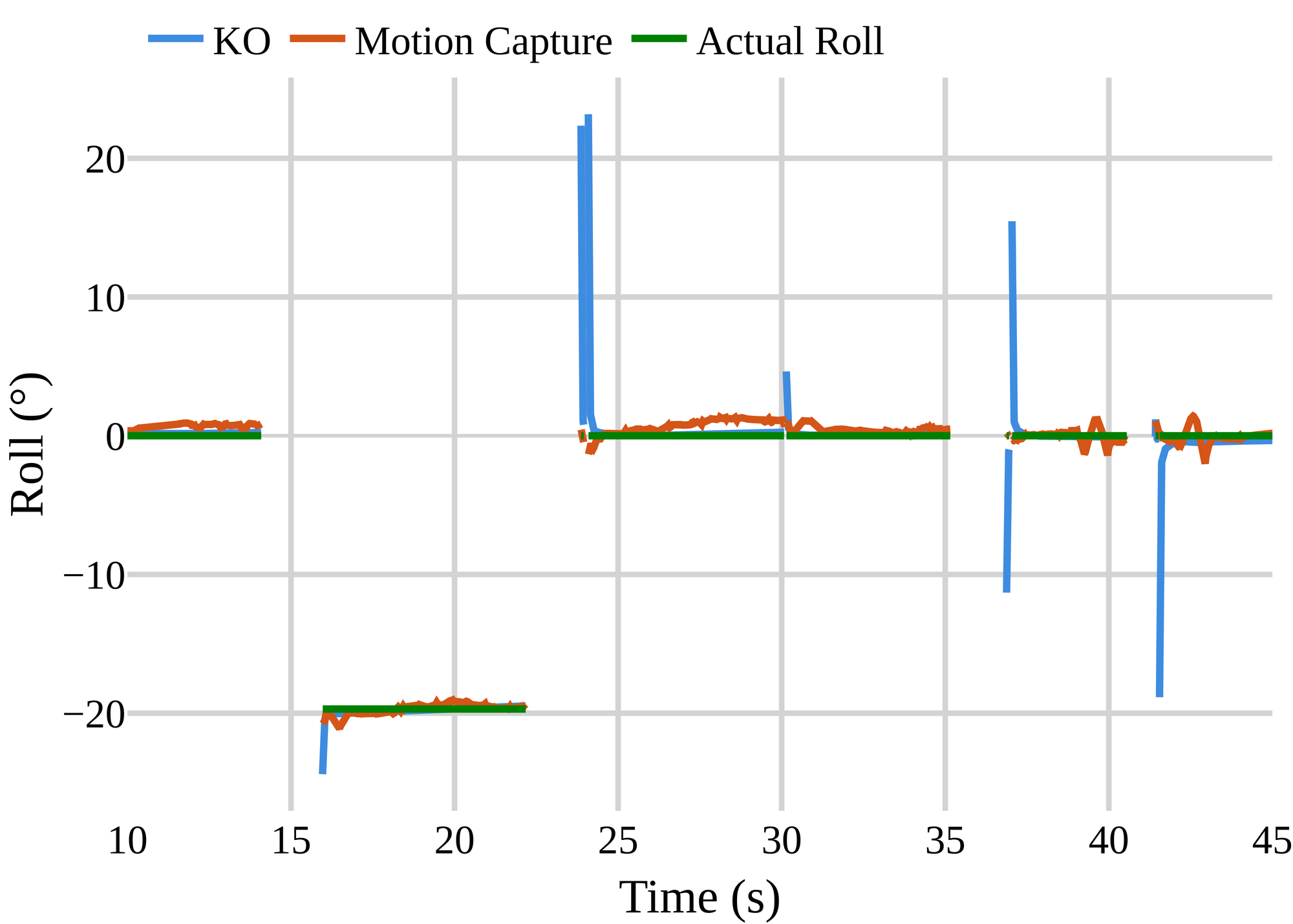}
    \caption{Right foot roll estimated by the Kinetics Observer during the multi-contact experiment, compared to the measured ground truth and the roll obtained by forward kinematics from the ground truth pose of the floating base (motion capture). The plot is discontinuous since the rest pose is not associated with the foot itself, but only exists when the foot is in contact with the environment. The contact happens to be detected over very short periods of time due to the bouncing of the foot when it touches the ground.}
    \label{fig:RightFootRoll}
    \end{figure}

Figure~\ref{fig:RightFootRoll} shows the estimation results of the roll angle of the right foot during the multi-contact motion. During the motion, the right foot steps on the flat ground, and on an inclined obstacle, tilted by -19.7$^{\circ}$ as measured with the precision instrument.

We observe that despite the considerable initial error, the estimated roll quickly converges within a range of $\pm$0.5$^{\circ}$ when walking on the flat ground. When walking on the tilted obstacle, it starts at -24.4$^{\circ}$ and converges within one second to -19.7$^{\circ}$, an error of 0.03$^{\circ}$ to the ground truth. It then remains within $\text{[-19.9$^{\circ}$, -19.5$^{\circ}$]}$ for the rest of the step. The roll obtained by forward kinematics from the ground truth of the floating base's pose varies within a range of $\text{[-20.1$^{\circ}$, -19.0$^{\circ}$]}$ during this step, which is generally a worse error than our estimation. Note that we didn't modify the tuning of the Kalman Filter's covariances, which greatly underestimate the introduced initial error.

\subsection{Disturbance Wrench Estimation} \label{subsec:ext-wrench-estimation}
\begin{figure}
    \centering
    \includegraphics[width=1\columnwidth]{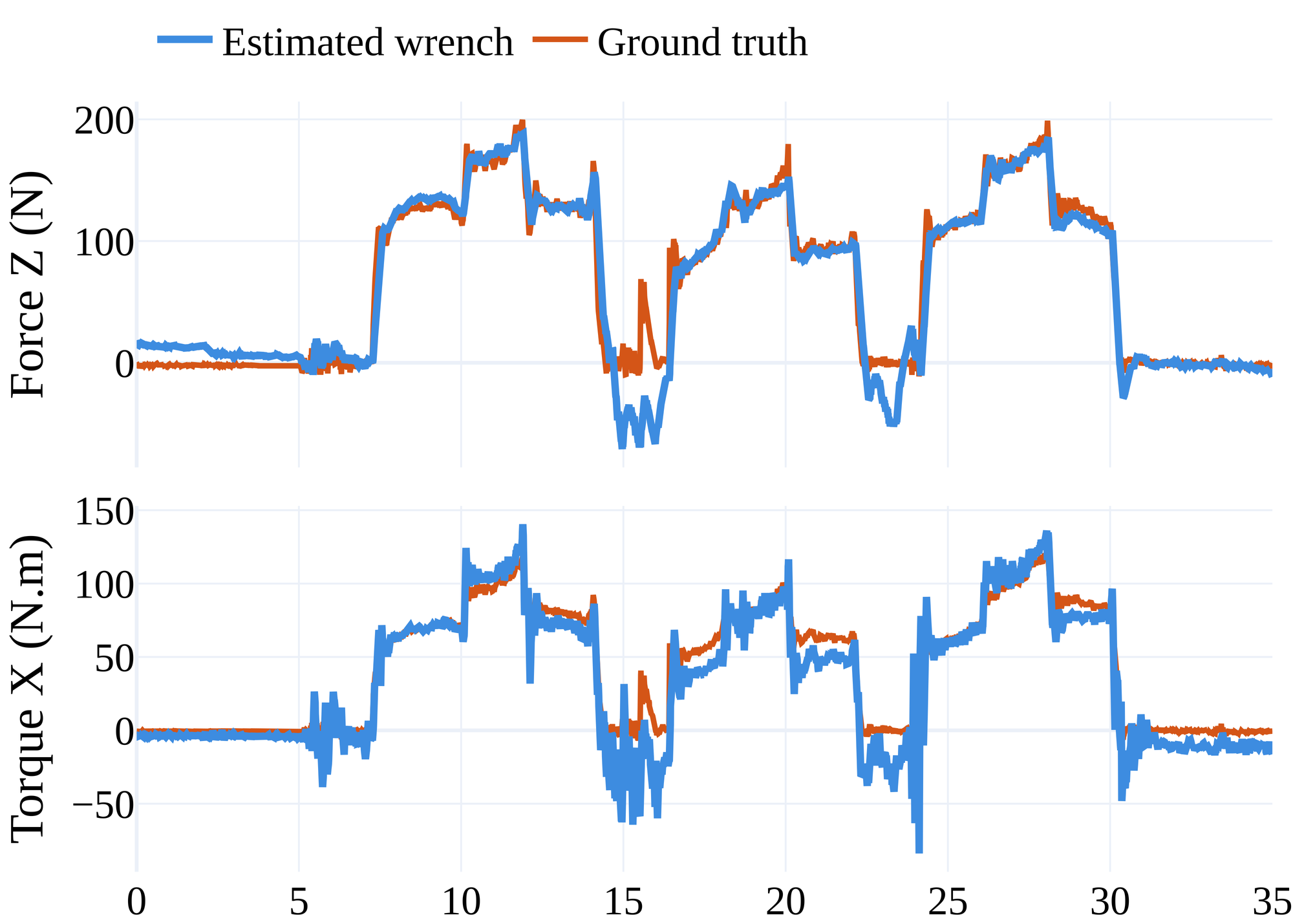}
    \caption{Estimated disturbance force along the $\boldsymbol{z}$ axis and moment along the $\boldsymbol{x}$ axis compared to the ground truth.}
    \label{fig:extForces}
\end{figure}

\noindent This section aims to evaluate the estimation of the disturbance wrench applied to the robot, made by the Kinetics Observer. For this purpose, the scenario used to evaluate the odometry over multi-contact motions (Section~\ref{subsec:Multi-contact-with-tilted}) was adapted through the following change. The Kinetics Observer was made blind to all the information coming from the robot's left hand. This contact is not detected and therefore its state is not estimated. Accordingly, the measurements of its wrench sensor are not used. Doing so, the contact at the left hand becomes a disturbance for the Kinetics Observer. We then evaluate how closely this disturbance wrench is estimated using the force sensor measurements as a ground truth. The mean absolute error (MAE) and standard deviation of the estimated disturbance wrench from the ground truth, across the four multi-contact experiments, are regrouped in Table~\ref{tab:ExtWrenchErrors}.
Fig.\ref{fig:extForces} also shows the wrench estimates on the axes with the highest variations during the first experiment. We can see that the estimator can provide an accurate and reactive estimation of the left hand's wrench. 
We observe an offset on the force and moment estimates on the order of 10 N and 10 N.m, respectively, between the estimates and the ground truth. This offset shows that in our model the disturbance wrench also serves as a slack variable to compensate for modeling errors, especially for identification errors in the dynamic parameters involved in the Newton-Euler dynamics (Eqs.~\eqref{eq:newtonLinAcc} and~\eqref{eq:eulerAngAcc}) (e.g., an error in a body mass and CoM position), and drifts in wrench sensor measurements. This slack variable cannot be separated from the actual disturbance wrench. Regardless, the Kinetics Observer shows satisfactory estimation results, for wrenches which can reach more than 150 N and 100 N.m, respectively. The reasonable standard deviation of the error from its mean also reflects the consistency of our estimate. 

\begin{table}
    \centering
    \small
    \setlength{\tabcolsep}{3pt}
    \caption{Mean error and standard deviation of the estimated disturbance wrench from the ground truth, across the four multi-contact experiments.}\label{tab:ExtWrenchErrors}
        \begin{tabular}{ c  c  c  c  c  c  c }
            \toprule
                      &   ${}^{\Gamma}\boldsymbol{F}_{e,x}$   &   ${}^{\Gamma}\boldsymbol{F}_{e,y}$   &   ${}^{\Gamma}\boldsymbol{F}_{e,z}$   &   ${}^{\Gamma}\boldsymbol{T}_{e,x}$   &   ${}^{\Gamma}\boldsymbol{T}_{e,y}$   &   ${}^{\Gamma}\boldsymbol{T}_{e,z}$   \\
             \midrule
             MAE &           6.4 N            &           9.7 N           &           9.2 N       &           10.3 N.m            &           4.4 N.m            &           3.6 N.m                       \\
             \midrule
             Std Dev. &           6.6 N             &           10.9 N           &           13.1 N       &           11.2 N.m             &           5.9 N.m             &           4.1 N.m                       \\
             \bottomrule

        \end{tabular}
    \end{table}

The estimation of disturbance wrenches using a viscoelastic model of contacts was already partially addressed by \citet{Benallegue2018ExtForceEstimationWithNoTorqueMeas}, but only for static cases. Here, the Kinetics Observer was able to estimate them during dynamic motions.

\subsection{Computational Performance}

\begin{figure}
    \centering
    \includegraphics[width=1\columnwidth]{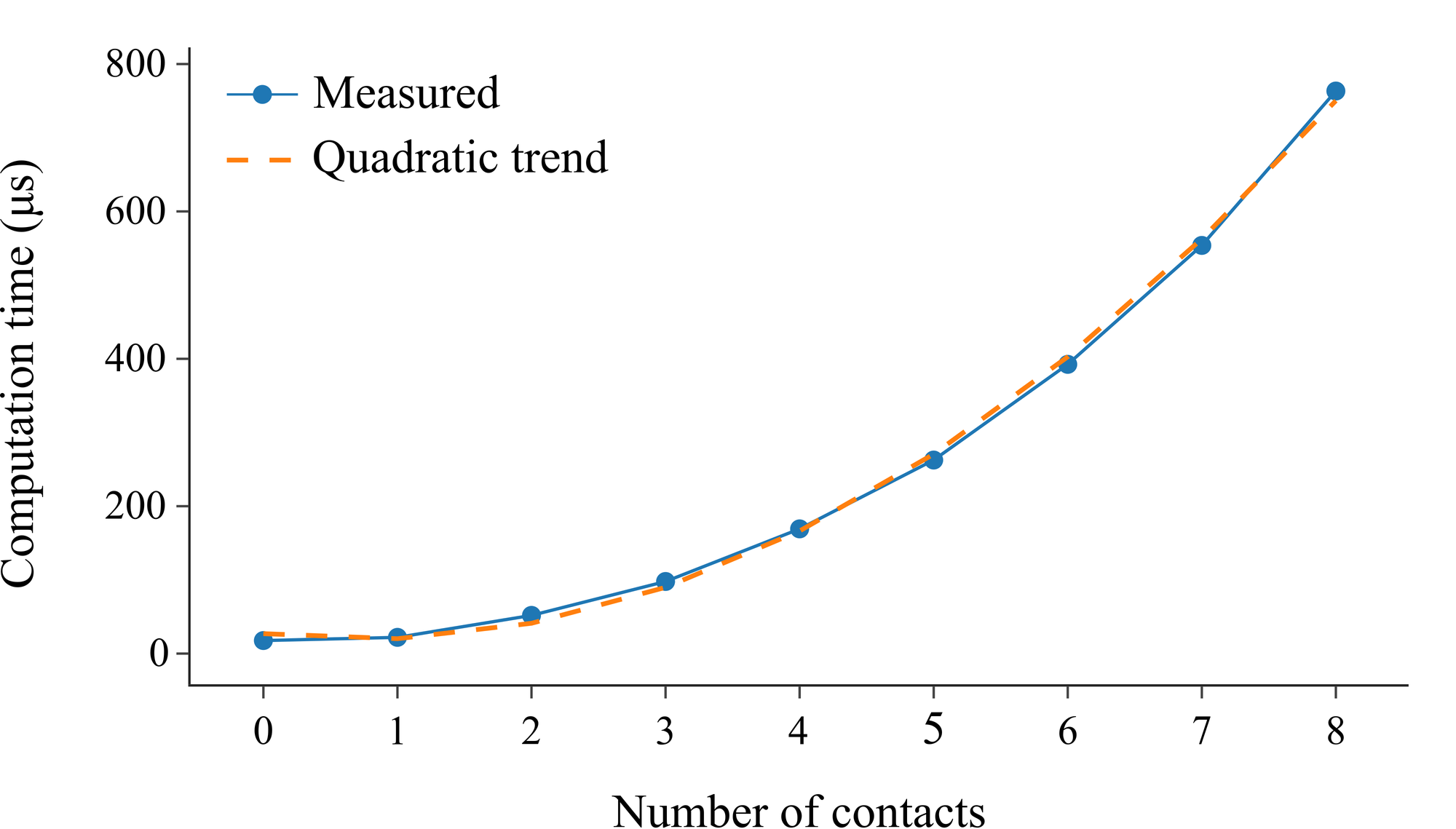}
    \caption{Average computation time as a function of the number of contacts. The dashed curve shows a quadratic trend for reference.}
    \label{fig:compute_time}
\end{figure}

Since the contact rest poses and wrenches are included in the Kinetics Observer's state vector, its computational cost increases with the number of contacts. In this section, we evaluate this scalability by computing the average time per iteration over 10,000 iterations for a number of contacts ranging from 0 to 8. Over the tested range, the runtime appears to scale approximately quadratically with the number of contacts, as shown in Figure~\ref{fig:compute_time}. 
Nevertheless, the computation time remains below one millisecond even for eight contacts, making the estimator suitable for real-time feedback in most control loops. This computational efficiency is partly due to the exploitation of the symmetry of the state covariance matrix, as well as to the analytical computation of the state-transition and observation Jacobian matrices, which avoids the costly evaluation of these matrices using finite differences. The analytical expressions of these matrices are available in~\citet{demont2025Thesis}. Further improvements in computational efficiency could be achieved by exploiting the sparsity of the state-transition and observation Jacobian matrices, which contain numerous zero elements.

\section{Conclusion and Future Works} \label{sec:conclusion}
In this paper, we presented the Kinetics Observer, a state-estimation framework that leverages the robot's centroidal dynamics and a contact dynamics model to achieve consistent fusion of IMU data, leg kinematics, and contact wrench measurements. This enables real-time joint estimation of the robot's kinematics, contact rest poses, and disturbance wrenches, with improved robustness to modeling errors, sensor noise, and contact slippage. It thereby provides feedback suited to versatile control of legged robots able to interact physically with the environment during complex locomotion.

Real-world experiments on humanoid robots showed that the Kinetics Observer provides accurate proprioceptive odometry over long distances, with improved robustness to contact slippage compared to the state of the art. It is also more robust than the RI-EKF to a drift of the gyrometer bias, such as the one of MEMS inertial units. It also provides accurate and reactive estimation of external disturbance wrenches. Finally, we showed that it can recover contact rest orientations   despite substantial initialization errors, demonstrating the   robustness of this contact-pose correction.

As an important remark, centroidal dynamics are at the core of the Kinetics Observer, notably enabling the integration of the viscoelastic contact model, the use of contact wrench measurements, and the filtering of IMU signals. While the use of dynamics introduces modeling uncertainty, notably due to identification errors in limb masses and inertia parameters, the Kinetics Observer includes mechanisms that mitigate their impact on the estimation. High-frequency dynamic-model errors, which may notably arise from sensor noise, are attenuated by the steady-state complementary-filter-like behavior of the Kalman filter~\citep{higgins1975comparison}. Low-frequency errors, such as inaccuracies in the identified inertial parameters or constant sensor biases, appear as consistent offsets in the Newton-Euler equations. These offsets are largely absorbed into the estimated disturbance wrench, which acts as a slack variable within our framework, thereby limiting their effect on the estimation of the other state variables. Overall, these mechanisms allow the Kinetics Observer to benefit from the centroidal dynamics model while remaining robust to its inaccuracies.

We also note that estimation accuracy might be improved for some variables by more extensive tuning of the covariances, but we noticed that improving the estimation of a variable may result in slightly decreasing accuracy for another one. This is because of the modeling ``tension'' that we create by integrating all these models: the kinematic, the dynamic and the viscoelastic one, within a single prediction loop.

Regarding the identified points for improvement, we showed that the computation time scales almost quadratically with the number of contacts due to the augmentation of the state vector and associated matrices. While the computation time remains satisfactory for real-time performances for most configurations, future work could explore alternatives to the Kalman Filter to reduce this scaling of the computational complexity.

At the core of the Kinetics Observer, lies the viscoelastic model of contacts, which enables to predict contact wrenches solely from state kinematics and inputs, and thus to obtain a coupling between kinematics and the external wrench applied to the robot. As detailed in Section~\ref{subsec:Viscoelastic-model}, we have chosen this model for its relevance and simplicity, but other models could be used as long as they relate kinematics to the forces and are differentiable. 
In our implementation, we consider constant contact flexibilities, which is of course a simplification since the effect of internal flexibilities on the kinematic chain up to a contact depends on the joint configuration. While satisfactory results are obtained using this simplification, future work could investigate the integration of a more detailed modeling of the flexibilities in the kinematic chain to improve estimation even further. Finally, we evaluated the estimator solely on rigid ground, restricting flexibilities to that intrinsic to the robot. Future works would also evaluate performance on deformable ground.
 
We also note that due to control and safety constraints, the estimator remains to be evaluated in more critical scenarios with even more slippage, walks on obstacles, faster walking, etc.

Additionally, we acknowledge that potential biases in the contact wrench sensor measurements were not considered in this work. These biases could notably affect odometry through the viscoelastic model. A more thorough analysis of this effect is left for future work.

Finally, several additional directions for improvement are considered, including: (i) to study ways to better identify environment stiffness and damping, (ii) to include additional proprioceptive sensors such as joint torque sensors or contact-located IMUs, (iii) to simplify the tuning of the Kalman Filter, (iv) to allow for the addition of exteroceptive sensors to ensure observability, (v) to improve contact and slippage detection, and (vi) to perform mapping through the contact local geometry estimation.

\section*{ORCID iDs}

Arnaud Demont:  \orcid{0009-0006-8325-8331} \\ 
Mehdi Benallegue:  \orcid{0000-0001-7537-9498}  \\
Abdelaziz Benallegue:  \orcid{0000-0003-2316-7981} \\
Pierre Gergondet:  \orcid{0000-0003-4925-2519} \\
Antonin Dallard:  \orcid{0000-0002-2569-5852}  \\
Rafael Cisneros-Limón:  \orcid{0000-0002-0850-070X} \\
Masaki Murooka:  \orcid{0000-0002-1258-0431} \\
Fumio Kanehiro:  \orcid{0000-0002-0277-3467}

\section*{Statements and declarations}
\subsection*{Ethical considerations}

This article does not contain any studies with human or animal participants.

\subsection*{Consent to participate}

Not applicable.
 
\subsection*{Declaration of conflicting interest}

The authors declared no potential conflicts of interest with respect to the research, authorship, and/or publication of this article

\subsection*{Funding}

This research includes results obtained from the project ``Programs for Bridging the Gap between R\&D and the Ideal Society (Society 5.0) and Generating Economic and Social Value (BRIDGE)/Practical Global Research in the AI x Robotics Services'', implemented by the Cabinet Office, Government of Japan.

\appendix
   
\section{Appendix: Index to Multimedia Extensions}

\begin{table}[h]
\centering
\caption{Index to Multimedia Extensions}
\label{tab:multimedia-extensions}
\small
\begin{tabularx}{\columnwidth}{ll>{\raggedright\arraybackslash}X}
\hline
Extension & Media type & Description \\
\hline
1 & Video & Experiments conducted to evaluate the Kinetics Observer. \\
\hline
\end{tabularx}
\end{table}

    \section{Appendix: One Contact Observability Lemma}\label{sec:appendix-lemma-observability}

      \noindent We consider the dynamical system represented by a contact subject to viscoelastic forces. This system is equipped with an IMU providing the angular velocity in the body's frame $\boldsymbol{\omega}_{l}$ and the local acceleration field $\boldsymbol{y}_{a}$ that includes gravity, and the contact force $\boldsymbol{F}$.
      Let's define 
    \begin{equation}\label{eq:lemma-dyn-syst1}
      {\textstyle \boldsymbol{x}=\left(\tilde{\boldsymbol{p}},\tilde{\boldsymbol{v}},\boldsymbol{\psi}\right)} \text{;} \qquad  \boldsymbol{u}=\left(\boldsymbol{\omega}_{l},\boldsymbol{y}_{a}\right) \text{;} \qquad  \boldsymbol{y}=\left(\boldsymbol{F}\right)
    \end{equation}
the state, the input and the measurement vectors respectively. $\tilde{\boldsymbol{p}}$ is the deformation of the flexibility, i.e. the difference between the current frame and the rest frame, expressed in the local frame. $\tilde{\boldsymbol{v}}$ is the linear velocity expressed in the local frame and $\boldsymbol{\psi}=\boldsymbol{R}^{\top}\boldsymbol{e}_{z}$ is the variable expressing the tilt of the system (similar to roll and pitch angles). The state dynamics are the following:

    \begin{align}\label{eq:lemma-dyn-syst2}
        & \dot{\tilde{\boldsymbol{p}}}=\tilde{\boldsymbol{v}}-\left[\boldsymbol{\omega}_{l}\right]_{\times}\tilde{\boldsymbol{p}} \text{ ,} \\
        & \dot{\tilde{\boldsymbol{v}}}=\boldsymbol{y}_{a}-g_{0}\boldsymbol{\psi}-\left[\boldsymbol{\omega}_{l}\right]_{\times}\tilde{\boldsymbol{v}} \text{ ,} \\
        & \dot{\boldsymbol{\psi}}=-\left[\boldsymbol{\omega}_{l}\right]_{\times}\boldsymbol{\psi} \text{ .}
    \end{align}
    
    Finally we define the measurements models
    \begin{align}\label{eq:lemma-dyn-syst3}
        \boldsymbol{F}=-\boldsymbol{K}_{p} \tilde{\boldsymbol{p}}-\boldsymbol{K}_{d} \tilde{\boldsymbol{v}} \text{ ,}
   \end{align}
where $\boldsymbol{K}_{p}$ and $\boldsymbol{K}_{d}$ are the stiffness of damping of the viscoelastic contact.

\begin{lemma}\label{lemma:observability}
The system defined by Equations \eqref{eq:lemma-dyn-syst1}-\eqref{eq:lemma-dyn-syst3} is uniformly observable if any of the following two conditions is met
\begin{enumerate}
\item If $\boldsymbol{\omega}_{l}$ and $\dot{\boldsymbol{\omega}}_{l}$ are null. 
\item OR, if exists a scalar $\lambda$ such that $\boldsymbol{K}_{d}=\lambda\boldsymbol{K}_{p}$ 
\end{enumerate}
\end{lemma}
    
  \begin{conjecture}
The system defined by Equations \eqref{eq:lemma-dyn-syst1}-\eqref{eq:lemma-dyn-syst3} is uniformly observable.
\end{conjecture}

\begin{proof}{Proof of the Lemma:}

    \noindent The observability matrix of this system is
    \begin{equation} 
      \nonumber
      \boldsymbol{\mathcal{O}} = 
      \left[\begin{array}{ccc}
        -\boldsymbol{K}_{p} & -\boldsymbol{K}_{d} & \mathbb{O}\\
        \boldsymbol{K}_{p}\left[\boldsymbol{\omega}_{l}\right]_{\times} & \boldsymbol{K}_{d}\left[\boldsymbol{\omega}_{l}\right]_{\times}-\boldsymbol{K}_{p} & g_{0}\boldsymbol{K}_{d}\\
        \boldsymbol{K}_{p}\left(\left[\dot{\boldsymbol{\omega}}_{l}\right]_{\times}-\left[\boldsymbol{\omega}_{l}\right]^{2}_{\times}\right) & \boldsymbol{\mathcal{O}}_{3,2} & \boldsymbol{\mathcal{O}}_{3,3}
        \end{array}\right]
    \end{equation}
    \noindent with 
    \begin{align}
      & \boldsymbol{\mathcal{O}}_{3,2} = 2\boldsymbol{K}_{p} \left[\boldsymbol{\omega}_{l}\right]_{\times} +\boldsymbol{K}_{d}\left(\left[\dot{\boldsymbol{\omega}}_{l}\right]_{\times}-\left[\boldsymbol{\omega}_{l}\right]^{2}_{\times}\right) \text{ ,} \\
      & \boldsymbol{\mathcal{O}}_{3,3} = g_{0}\boldsymbol{K}_{p}-2g_{0}\boldsymbol{K}_{d}\left[\boldsymbol{\omega}_{l}\right]_{\times} \text{ .}
    \end{align}
    \noindent It can be shown that this observability matrix is full-rank in each of the two conditions of the lemma above 
    \begin{enumerate}
      \item If $\boldsymbol{\omega}_{l}$ and $\dot{\boldsymbol{\omega}}_{l}$ are null:
      \begin{equation} 
        \boldsymbol{\mathcal{O}} = 
        \left[\begin{array}{ccc}
          -\boldsymbol{K}_{p} & -\boldsymbol{K}_{d} & \mathbb{O}\\
          \mathbb{O} & -\boldsymbol{K}_{p} & g_{0}\boldsymbol{K}_{d}\\
          \mathbb{O} & \boldsymbol{\mathcal{O}}_{3,2} & \boldsymbol{\mathcal{O}}_{3,3}
          \end{array}\right] \text{ .}
      \end{equation}
      \item If exists a scalar $\lambda$ such that $\boldsymbol{K}_{d}=\lambda\boldsymbol{K}_{p}$ 
      \begin{equation}
        \boldsymbol{\mathcal{O}}= \boldsymbol{A}_{1} \boldsymbol{A}_{2} \boldsymbol{A}_{3} \text{ ,}
      \end{equation}
      with 

      \begin{align}
        \boldsymbol{A}_{1} & =
        \left[\begin{array}{ccc}
        \boldsymbol{K}_{p} & \mathbb{O} & \mathbb{O}\\
        \mathbb{O} & \boldsymbol{K}_{p} & \mathbb{O}\\
        \mathbb{O} & \mathbb{O} & \boldsymbol{K}_{p}
        \end{array} 
        \right] \text{ ,} \\
        \boldsymbol{A}_{2} & =
        \left[\begin{array}{ccc}
        -\mathbb{I} & \mathbb{O} & \mathbb{O}\\
        \left[\boldsymbol{\omega}_{l}\right]_{\times} & -\mathbb{I} & \mathbb{O}\\
        \left[\dot{\boldsymbol{\omega}}_{l}\right]_{\times}-\left[\boldsymbol{\omega}_{l}\right]_{\times}^{2} & 2\left[\boldsymbol{\omega}_{l}\right]_{\times} & g_{0}\mathbb{I}
        \end{array}
        \right] \text{ ,} \\
        \boldsymbol{A}_{3} & =
        \left[\begin{array}{ccc}
          \mathbb{I} & \lambda\mathbb{I} & \boldsymbol{0}\\
        \boldsymbol{0} & \mathbb{I} & -g_{0}\lambda\mathbb{I}\\
        \boldsymbol{0} & \boldsymbol{0} & \mathbb{I}
        \end{array}
        \right] \text{ .}
        \end{align}
        
    \end{enumerate}
    
    This provides the uniform observability of the system.

\end{proof}

\noindent Proving that this matrix is full rank in general is a tedious task that we will not develop in this paper, but we believe the occurrence of a singular observability matrix highly unlikely. Therefore, we leave it as a conjecture.

In conclusion, the visco-elastic model of contacts along with the contact wrench measurement and an IMU provide the observability of the position and velocity errors $\tilde{\boldsymbol{p}}$ and $\tilde{\boldsymbol{v}}$, and of the robot's tilt $\boldsymbol{R}^{\top}\boldsymbol{e}_{z}$.

\section{Appendix: Implementation of the MEKF}\label{sec:appendix-implementation}

\subsection{Contact Rest Pose Initialization}\label{subsec:contact-rest-pose-initialization}
\noindent In order to perform odometry, the Kinetics Observer requires the 'impulsional' input $\left\{\check{\boldsymbol{p}}_{r,i},\check{\boldsymbol{\Omega}}_{r,i}\right\}$, which is used to initialize in the state vector the rest pose $\left\{ \boldsymbol{p}_{r,i},\boldsymbol{\Omega}_{r,i}\right\}$ of the successive contacts in the world. As a reminder, the contact's rest pose is the pose of the contact frame when the material deformation is zero. In theory, it can be obtained by forward kinematics at the exact instant the robot's limb touches the environment. However, in practice contacts are detected as a reaction wrench is already applied and thus the deformation has already occurred. The rest pose is therefore not readily available, to recover it, it is thus necessary to identify and compensate for the deformation. This can be done through a reformulation of the viscoelastic model, which instead of giving a wrench from the deformation, gives the deformation from the the wrench measurement. 
    In other words, from \eqref{eq:ViscoElasticForce} we obtain the translational deformation and add it to $\boldsymbol{p}_{\mathcal{C}_{i}}$ to get the initial guess on the rest position:
    \begin{align}
    \check{\boldsymbol{p}}_{r,i} & =\boldsymbol{p}_{\mathcal{C}_{i}}+\boldsymbol{R}_{\mathcal{C}_{i}}\boldsymbol{K}_{p,t}^{-1}\left(\boldsymbol{y}_{F,i}+\boldsymbol{R}_{\mathcal{C}_{i}}^{\top}\boldsymbol{K}_{d,t}\dot{\boldsymbol{p}}_{\mathcal{C}_{i}}\right)\text{ .}
    \end{align}
    
    \noindent Similarly, we obtain the rotational deformation $\tilde{\boldsymbol{R}}_{i}$. Let us define the term $\boldsymbol{D}_{i}=\widetilde{\boldsymbol{R}}_{i}-\widetilde{\boldsymbol{R}}_{i}^{\top}$. From \eqref{eq:ViscoElasticTorque}, we have:
    \begin{align}
      \left[\boldsymbol{D}_{i}\right]_{\vee} & =-2\boldsymbol{R}_{\mathcal{C}_{i}}\boldsymbol{K}_{p,r}^{-1}\left(\boldsymbol{y}_{T,i}+\boldsymbol{R}_{\mathcal{C}_{i}}^{\top}\boldsymbol{K}_{d,r}\boldsymbol{\omega}_{\mathcal{C}_{i}}\right)\text{ .}\label{eq:vecD}
    \end{align}
    
    \noindent $\boldsymbol{D}_{i}$ also respects the property~\eqref{eq:r-rt-property}, which gives 
        \begin{equation}
    \phi_{i}   =\arcsin\left(\frac{\left\Vert \left[\boldsymbol{D}_{i}\right]_{\vee} \right\Vert }{2}\right)\text{ ,} \qquad
    \boldsymbol{u}_{i}   =\frac{\left[\boldsymbol{D}_{i}\right]_{\vee}}{\left\Vert \left[\boldsymbol{D}_{i}\right]_{\vee} \right\Vert }\text{ ,} 
    \end{equation}

    \noindent where $\phi_{i}$ and $\boldsymbol{u}{}_{i}$ are the rotation axis and angle corresponding to $\widetilde{\boldsymbol{R}}_{i}$. We finally obtain the initial guess on the contact rest orientation:
    \begin{equation}
    \check{\boldsymbol{R}}_{r,i}=\widetilde{\boldsymbol{R}}_{i}^{\top}\boldsymbol{R}_{\mathcal{C}_{i}}\text{ ,}\label{eq:restOrientationOdometry}
    \end{equation}
    which completes our input. Note that since the position along the vertical axis $\boldsymbol{z}$ is not observable, small displacements resulting from the IMU integration and the viscoelastic behavior of the contacts may accumulate at each step, and cannot be corrected by the observer. This would lead to significant drifts over long walks. To solve this problem when walking on a flat ground, one can constrain the robot to remain on the ground by initializing the rest position along the vertical axis and the associated initial state variance of all newly set contacts to zero.
    
    Finally, it is important to note that this estimator can be used without odometry. This is useful when one does not want the robot estimate to drift from the reference plan, but still needs to observe the local state of the robot, for example to control balance and locomotion. In such a case it is sufficient to initialize the contact rest pose with the reference provided by the contact planner.
    
    \subsection{Tuning of the Kinetics Observer}\label{subsec:tuning-of-KO}
    \noindent The main challenges in implementing the Kinetics Observer are identifying the flexibilities used in the viscoelastic model, and tuning the covariance matrices involved in the Kalman filter. Here we provide some insight into the process.

    Covariance matrices are constructed by combining the covariance submatrices corresponding to each state and measurement variable along the three axes. These submatrices are considered diagonal, which means that we define only the variance on each variable and assume that they are independent of each other. The \emph{measurement noise} variances can be obtained directly from the noise models on the sensors.
    The explanation of the tuning of the initial \emph{state covariance} matrix and the \emph{process noise} covariance matrices will focus on the submatrices associated with the viscoelastic model, since they play a crucial role in the behavior of our estimator. 
    \paragraph{contact rest pose} The initial state variance and the process noise variance were respectively set based on the expected orders of magnitude of the discrepancy between our model and reality at contact creation, and of potential drifts of contacts. The process noise submatrix is then designed to respect specific constraints, as we further develop in Section~\ref{sec:Implementation}.
    \paragraph{contact wrench} The associated state-transition model is the viscoelastic model defined in Section~\ref{subsec:Viscoelastic-model}. While the viscoelastic model is a useful approximation, it is evident that the wrench sensors are more reliable, we thus set process noise variances higher than the measurement noise variances.

    As an important note, the state covariances of new contact poses are initialized with a fixed value even in odometry mode. This is not consistent with reality since the position of every step is supposed to be more uncertain than the previous one, and thus, the covariances should add up. This imprecise choice has been made to prevent the covariance from building up and causing instability problems during long experiments. However, this care can be dropped if an additional absolute pose measurement (e.g., GPS or SLAM) is added to the estimator, since in this case the system would be observable, and the covariance would remain bounded.
    
    When walking on a rigid ground, contact flexibilities can be approximated by those of the robot, which can be empirically identified~\citep{Mikami2014IdentificationHRP2Dynamics, flayols2018These}. Identifying such flexibilities may nevertheless be challenging, and considering constant flexibilities is also an assumption as they would generally change depending on the robot's configuration. 
    Nevertheless, we argue that it is not critical to identify these flexibilities perfectly and that the estimator would provide satisfactory estimation despite identification approximations. 
    During experiments with RHP Friends, we used the flexibilities that were identified for the robot HRP2~\citep{Mikami2014IdentificationHRP2Dynamics} and still obtained satisfactory results.

    A reason for this robustness is that the Newton-Euler dynamics in Eqs.~\eqref{eq:newtonLinAcc} and~\eqref{eq:eulerAngAcc} are not significantly affected by such approximations. Indeed, as discussed above, contact wrench measurements are generally more accurate than the viscoelastic model predictions and are therefore associated with covariances that are lower by several orders of magnitude. Consequently, the contact wrench predicted by the viscoelastic model has only a minor influence on the final wrench estimate used in the kinematics prediction.
    This behavior is expected, since the viscoelastic model mainly serves to penalize drift between the robot and its contacts and to correct it using the wrench measurements, as explained in Section~\ref{subsec:Viscoelastic-model}. The flexibilities therefore act as gains in this correction process, which remains valid regardless of the gain values.
    
    To prove our claim, we provide in Table~\ref{tab:results_diff_flexs} the kinematics estimation errors computed across the four sequences of the multi-contact experiment with HRP-5P and the three trials of walking with slipping contacts with RHP Friends, for different values of flexibilities. We multiplied and divided the initial flexibilities by a factor of ten to show the impact of a considerable approximation of an order of magnitude. In Table~\ref{tab:dist_wrench_res_flexs}, we have also added the average estimation error in the disturbance wrench for these flexibilities. We can see that overall, the estimation accuracy remains satisfactory across all tunings. The only ``outlier'' in these results is the Relative Error in the translation along the $\boldsymbol{z}$ axis during walking with slipping contacts with flexibilities divided by ten. In that case, flexibilities, notably the linear stiffness, became so low that the viscoelastic model had to explain the measured reaction force by a considerable distance between the \emph{contact frame} and its \emph{rest} frame. For example, a reaction force of 500 N, which is commonly measured at RHP Friends' feet, would correspond to a distance of 16.7 cm  between both frames. This highly unrealistic discrepancy leads  the robot to quickly sink into the ground at every step. From these results, we conclude that the Kinetics Observer provides satisfactory estimation accuracy despite inaccuracies in the flexibility identification, as long as the tuned flexibility values remain within a physically meaningful range.
    We note that if no force measurement is available, flexibilities should be identified with more care.

    \begin{table} 
        \centering
        \small
        \setlength{\tabcolsep}{3pt}
        \caption{Average disturbance wrench estimation error for different tunings of the viscoelastic contact flexibilities, across the four trials of the multi-contact experiment.} \label{tab:dist_wrench_res_flexs}
            \begin{tabular}{ c  c  c  c  c  c  c }
                \toprule
             
            % \multicolumn{7}{c}{\textbf{Kinematic estimation errors}} \\
            % \midrule
 
                        &   ${}^{\Gamma}\boldsymbol{F}_{e,x}$   &   ${}^{\Gamma}\boldsymbol{F}_{e,y}$   &   ${}^{\Gamma}\boldsymbol{F}_{e,z}$   &   ${}^{\Gamma}\boldsymbol{T}_{e,x}$   &   ${}^{\Gamma}\boldsymbol{T}_{e,y}$   &   ${}^{\Gamma}\boldsymbol{T}_{e,z}$   \\
                \midrule
                Init flex &           6.4 N            &           9.7 N           &           9.2 N       &           10.3 N.m            &           4.4 N.m            &           3.6 N.m                       \\
                \midrule
                $\div 10$ &           6.8 N             &           9.8 N           &           9.1 N       &           10.3 N.m             &           4.4 N.m             &           3.6 N.m                       \\
                \midrule
                $\times 10$ &           6.8 N             &           10.9 N           &           9.2 N       &           10.3 N.m             &           4.4 N.m             &           3.6 N.m                       \\
                \bottomrule

            \end{tabular}
        \end{table}

    To conclude, all the parameters that we tuned for our experiments are summarized in Table~\ref{tab:KineticsObserver-parameters}.
    
    \begin{table}
        \newcolumntype{Y}{>{\centering\arraybackslash}X}
        \footnotesize
        \centering  
        \setlength{\tabcolsep}{2pt}
        \caption{Mean and standard deviation (in parentheses) of the kinematic estimation errors for different tunings of viscoelastic contact flexibilities, computed across all trials of the two experiments: walking with slipping contacts and multi-contact motion. The 0.3 m and 1 m Relative Error are represented, respectively.} 
        \label{tab:results_diff_flexs}
            \begin{tabularx}{\columnwidth}{@{}c c Y Y Y Y Y Y@{}}
                \toprule
                \multirow{4}{*}{} 
                & \multirow{4}{*}{} 
                & \multicolumn{2}{c}{RE Trans.} 
                & \multicolumn{2}{c}{RE Ori.} 
                & \multicolumn{2}{c}{Lin. vel.} \\

                &  
                & \multicolumn{2}{c}{[m]} 
                & \multicolumn{2}{c}{[deg]} 
                & \multicolumn{2}{c}{$[\text{m.s}^{-1}]$} \\   
            
                \addlinespace[0.2em]

                \cline{2-8}

                \addlinespace[0.2em]

                & 
                &       $\left\{\boldsymbol{x}, \boldsymbol{y}\right\}$   &         $\boldsymbol{z}$    & $\text{Tilt}$ &    $\text{Yaw}$  &   $\left\{\boldsymbol{x}, \boldsymbol{y}\right\}$  &  $\boldsymbol{z}$  \\
                \midrule

                \multirow{3}{*}[ -0.8ex ]{\makecell[c]{Slipping \\ contacts}} &     Init flex    &   \getErrorResult{Slippingodometry}{Kineticsobserver}{Relerror}{Transxy}{Meanabs}  &  \getErrorResult{Slippingodometry}{Kineticsobserver}{Relerror}{Transz}{Meanabs}  &  \getErrorResult{Slippingodometry}{Kineticsobserver}{Relerror}{Tilt}{Meanabs}    &      \getErrorResult{Slippingodometry}{Kineticsobserver}{Relerror}{Yaw}{Meanabs}     &  \getErrorResult{Slippingodometry}{Kineticsobserver}{Velerror}{EstimateXy}{Meanabs}    &   \getErrorResult{Slippingodometry}{Kineticsobserver}{Velerror}{EstimateZ}{Meanabs}  \\ 
                
                \addlinespace[0.2em]

                \cline{2-8}

                \addlinespace[0.3em]

                  & $\div$ 10      &    \getErrorResult{Slippingodometryb}{Kineticsobserver}{Relerror}{Transxy}{Meanabs}   &  \getErrorResult{Slippingodometryb}{Kineticsobserver}{Relerror}{Transz}{Meanabs}   &  \getErrorResult{Slippingodometryb}{Kineticsobserver}{Relerror}{Tilt}{Meanabs}    &       \getErrorResult{Slippingodometryb}{Kineticsobserver}{Relerror}{Yaw}{Meanabs}     &  \getErrorResult{Slippingodometryb}{Kineticsobserver}{Velerror}{EstimateXy}{Meanabs}    &   \getErrorResult{Slippingodometryb}{Kineticsobserver}{Velerror}{EstimateZ}{Meanabs}  \\ 
                
                \addlinespace[0.2em]

                \cline{2-8}

                \addlinespace[0.3em]
                            
                & $\times$    10  &    \getErrorResult{Slippingodometryc}{Kineticsobserver}{Relerror}{Transxy}{Meanabs}   &  \getErrorResult{Slippingodometryc}{Kineticsobserver}{Relerror}{Transz}{Meanabs}   &  \getErrorResult{Slippingodometryc}{Kineticsobserver}{Relerror}{Tilt}{Meanabs}    &       \getErrorResult{Slippingodometryc}{Kineticsobserver}{Relerror}{Yaw}{Meanabs}     &  \getErrorResult{Slippingodometryc}{Kineticsobserver}{Velerror}{EstimateXy}{Meanabs}    &   \getErrorResult{Slippingodometryc}{Kineticsobserver}{Velerror}{EstimateZ}{Meanabs}  \\ 

                 \midrule

                \multirow{3}{*}[ -0.8ex ]{\makecell[c]{Multi-contact \\ motion}} &  Init flex   &   \getErrorResult{Multicontact}{Kineticsobserver}{Relerror}{Transxy}{Meanabs}  &  \getErrorResult{Multicontact}{Kineticsobserver}{Relerror}{Transz}{Meanabs}   &  \getErrorResult{Multicontact}{Kineticsobserver}{Relerror}{Tilt}{Meanabs}   &       \getErrorResult{Multicontact}{Kineticsobserver}{Relerror}{Yaw}{Meanabs}     &  \getErrorResult{Multicontact}{Kineticsobserver}{Velerror}{EstimateXy}{Meanabs}    &   \getErrorResult{Multicontact}{Kineticsobserver}{Velerror}{EstimateZ}{Meanabs}  \\  
                
                \addlinespace[0.2em]

                \cline{2-8}

                \addlinespace[0.3em]

                 & $\div$ 10       &    \getErrorResult{Multicontactb}{Kineticsobserver}{Relerror}{Transxy}{Meanabs}   &  \getErrorResult{Multicontactb}{Kineticsobserver}{Relerror}{Transz}{Meanabs}   &  \getErrorResult{Multicontactb}{Kineticsobserver}{Relerror}{Tilt}{Meanabs}    &       \getErrorResult{Multicontactb}{Kineticsobserver}{Relerror}{Yaw}{Meanabs}     &  \getErrorResult{Multicontactb}{Kineticsobserver}{Velerror}{EstimateXy}{Meanabs}    &   \getErrorResult{Multicontactb}{Kineticsobserver}{Velerror}{EstimateZ}{Meanabs}  \\ 
                
                \addlinespace[0.2em]

                \cline{2-8}

                \addlinespace[0.3em]
                            
                &  $\times$ 10       &    \getErrorResult{Multicontactc}{Kineticsobserver}{Relerror}{Transxy}{Meanabs}   &  \getErrorResult{Multicontactc}{Kineticsobserver}{Relerror}{Transz}{Meanabs}   &  \getErrorResult{Multicontactc}{Kineticsobserver}{Relerror}{Tilt}{Meanabs}    &       \getErrorResult{Multicontactc}{Kineticsobserver}{Relerror}{Yaw}{Meanabs}     &  \getErrorResult{Multicontactc}{Kineticsobserver}{Velerror}{EstimateXy}{Meanabs}    &   \getErrorResult{Multicontactc}{Kineticsobserver}{Velerror}{EstimateZ}{Meanabs}  \\ 
                 
                \bottomrule
            \end{tabularx}
    \end{table}

    \begin{table}
      \footnotesize
      \centering
      \caption{Tuned parameters of the Kinetics Observer (Units Omitted)}\label{tab:KineticsObserver-parameters}
      \tabcolsep=0.09cm
      \renewcommand{\arraystretch}{1.2}
      \begin{tabularx}{\columnwidth}{l c c}
        \toprule
                                                      &           \textbf{Initial}            &         \textbf{Process noise}      \\[-0.6ex]
                                                      &           \textbf{state variances}            &         \textbf{variances}     \\
        \midrule
                      Position $\boldsymbol{p}_l$     &   $\boldsymbol{\mathbb{O}}$        &        $10^{-9}.\boldsymbol{\mathbb{I}}$           \\
        \midrule
                    Orientation $\boldsymbol{R}$    &     \footnotesize{diag}\tiny{$\left(1, 1, 0 \right).10^{-4}$} \tiny{$\: {}^{\boldsymbol{\left( 1 \right)}}$}    &    $10^{-11}.\boldsymbol{\mathbb{I}}$      \\
        \midrule
                   Lin. Velocity $\boldsymbol{v}_l$   &        $10^{-6}.\boldsymbol{\mathbb{I}}$           &        $10^{-8}.\boldsymbol{\mathbb{I}}$           \\
        \midrule
                   Ang. Velocity $\boldsymbol{\omega}_l$  &        $10^{-6}.\boldsymbol{\mathbb{I}}$           &        $10^{-10}.\boldsymbol{\mathbb{I}}$           \\
                   \midrule
                  Gyrometer bias $\boldsymbol{b}_{g,j}$     &        $10^{-8}.\boldsymbol{\mathbb{I}}$           &        $10^{-18}.\boldsymbol{\mathbb{I}}$           \\
                  \midrule
                  Disturb. force $\boldsymbol{F}_e$ &        $10^{2}.\boldsymbol{\mathbb{I}}$           &        $9.10^{-2}.\boldsymbol{\mathbb{I}}$           \\
                  \midrule
                 Disturb. moment $\boldsymbol{T}_e$ &        $10^{2}.\boldsymbol{\mathbb{I}}$        &        $9.10^{-2}.\boldsymbol{\mathbb{I}}$           \\
                 \midrule
                 Contact rest pos. $\boldsymbol{p}_{r,i}$ &   \footnotesize{diag}\tiny{$\left(5, 5, 2 \right).10^{-6}$} &  \footnotesize{diag}\tiny{$\left(10^{-5}, 10^{-5}, 10^{-9} \right)$} \\
                 \midrule
                 Contact rest ori. $\boldsymbol{R}_{r,i}$ &      $2.10^{-4}.\boldsymbol{\mathbb{I}}$     &  \footnotesize{diag}\tiny{$\left(10^{-8}, 10^{-8}, 10^{-4} \right)$}  \\
                 \midrule
                 Contact force ${^{\mathcal{C}}}\boldsymbol{F}_i$ &        $5.10^{-1}.\boldsymbol{\mathbb{I}}$           &   $900.\boldsymbol{\mathbb{I}}$         \\
                 \midrule
                 Contact moment ${^{\mathcal{C}}}\boldsymbol{T}_i$ &        $9.10^{-4}.\boldsymbol{\mathbb{I}}$           &        $225.\boldsymbol{\mathbb{I}}$           \\
                  \\
                                   &           \multicolumn{2}{c}{\textbf{Measurement noise variances}}                      \\
                                   &            \multicolumn{2}{c}{\textbf{(HRP-5P | RHP Friends)}}                      \\
                                   \midrule
                      Gyrometer     &          $25.10^{-8}.\boldsymbol{\mathbb{I}}$         &              $16.10^{-6}.\boldsymbol{\mathbb{I}}$                   \\
                      \midrule
                    Accelerometer    &         $25.10^{-4}.\boldsymbol{\mathbb{I}}$         &              $36.10^{-4}.\boldsymbol{\mathbb{I}}$                   \\
                    \midrule
                   Contact forces    &          $5.10^{-1}.\boldsymbol{\mathbb{I}}$         &              $1.\boldsymbol{\mathbb{I}}$                   \\
                   \midrule
                   Contact moments    &         $9.10^{-4}.\boldsymbol{\mathbb{I}}$         &              $9.10^{-4}.\boldsymbol{\mathbb{I}}$                   \\
                 \\
                                   &                  \multicolumn{2}{c}{\textbf{Contact flexibilities}}                \\
                                   &                  \multicolumn{2}{c}{\textbf{(HRP-5P | RHP Friends)}}                \\
                                   \midrule
                   Linear stiffness   &       \multirow{2}{*}{$3.10^{5}.\boldsymbol{\mathbb{I}}$}           &       \multirow{2}{*}{$3.10^{4}.\boldsymbol{\mathbb{I}}$}           \\
                   $\boldsymbol{K}_{p,t}\text{ [N/m]}$   &                                                   &                                                                   \\
                   \midrule
                   Linear damping   &       \multirow{2}{*}{$150.\boldsymbol{\mathbb{I}}$}            &       \multirow{2}{*}{$150.\boldsymbol{\mathbb{I}}$}           \\
                   $\boldsymbol{K}_{d,t}\text{ [N.s/m]}$   &                                                   &                                                                   \\
                   \midrule
                   Angular stiffness   &       \multirow{2}{*}{$1000.\boldsymbol{\mathbb{I}}$}           &       \multirow{2}{*}{$727.\boldsymbol{\mathbb{I}}$}           \\
                   $\boldsymbol{K}_{p,r}\text{ [N.m/rad]}$   &                                                   &                                                                   \\
                   \midrule
                  Angular damping   &        \multirow{2}{*}{$17.\boldsymbol{\mathbb{I}}$}            &        \multirow{2}{*}{$17.\boldsymbol{\mathbb{I}}$}           \\
                   $\boldsymbol{K}_{d,r}\text{ [N.m.s/rad]}$   &                                                   &                                                                   \\
                   \midrule
                 \\
                \multicolumn{3}{l}{\scriptsize{$\boldsymbol{^{\left( 1 \right)}}$ diag$\left( \right)$ is the operator that transforms a $\mathbb{R}^{n}$ vector into a $\mathbb{R}^{n\times n}$ diagonal matrix}} \tabularnewline
                \multicolumn{3}{l}{ \scriptsize{ whose diagonal terms correspond to the vector components.}} \tabularnewline
      \end{tabularx}
    \end{table}

       \section{Appendix: Summary of notation}\label{sec:appendix-notations}
    
    \begin{table}[!ht]
      \centering
      \caption{Notations for the kinematic variables of the centroid frame in the world frame}\label{tab:state-kinematics}
      \begin{tabular}{ c  c  c }
        \toprule
                                    &          Notation            &         Definition     \\
          \midrule
                Position     &      $\boldsymbol{p}_{l}$           &        $\boldsymbol{R}_{\Gamma}^{\top}\boldsymbol{p}_{\Gamma}$           \\ 
            \midrule
                \multirow{2}{*}{Orientation}      &  $\boldsymbol{R}$    &    $\boldsymbol{R}_{\Gamma}$        \\
            \cline{2-3}
                                          & $\boldsymbol{\Omega}$ &  $\boldsymbol{\Omega}_{\Gamma}$        \\
            \midrule
              Lin. Velocity    &        $\boldsymbol{v}_{l}$           &        $\boldsymbol{R}_{\Gamma}^{\top}\dot{\boldsymbol{p}}_{\Gamma}$  \\
              \midrule
              Ang. Velocity    &        $\boldsymbol{\omega}_{l}$           &        $\boldsymbol{R}_{\Gamma}^{\top}\boldsymbol{\omega}_{\Gamma}$  \\
              \midrule
              Lin. Acceleration    &        $\boldsymbol{a}_{l}$        &        $\boldsymbol{R}_{\Gamma}^{\top}\ddot{\boldsymbol{p}}_{\Gamma}$ \\
              \midrule
              Ang. Acceleration    &        $\dot{\boldsymbol{\omega}}_{l}$        &        $\boldsymbol{R}_{\Gamma}^{\top}\dot{\boldsymbol{\omega}}_{\Gamma}$  \\
          \bottomrule
      \end{tabular}
        
    \end{table}

    \small{
    
    \noindent General notations:
    \begin{itemize}[leftmargin=1em]
    \item {$\hat{\boldsymbol{\circ}}$ : estimated variable, $\bar{\boldsymbol{\circ}}$ : predicted variable, $\tilde{\boldsymbol{\circ}}$ : error variable.}
    \item {$n_{c}$ : number of current contacts.}
    \item {$n_{w}$ : number of contact wrench sensors.}
    \item {$n_{I}$ : number of IMUs.}
    \end{itemize}
    
    \noindent Main frames \ref{fig:FramesAndVars}:
    \begin{itemize}[leftmargin=1em]
    \item {$\mathcal{W}$ : world frame.}
    \item {$\Gamma$ : centroid frame.}
    \item {$\mathcal{C}_{i}$ : frame of the $i$-th contact. }
    \item {$\mathcal{C}_{r,i}$ : rest frame of the $i$-th contact. }
    \end{itemize}
    
    \noindent State variables \ref{subsec:state-vec-def}:
    \begin{itemize}[leftmargin=1em]
    \item {$\boldsymbol{p}_{l}$ : position of the centroid frame in the world frame, expressed in the centroid frame.}
    \item {$\boldsymbol{\Omega}$ : vector representation (e.g. quaternion) of the orientation of the centroid frame in the world frame.}
    \item {$\boldsymbol{v}_{l}$ : linear velocity of the centroid frame in the world frame, expressed in the centroid frame.}
    \item {$\boldsymbol{\omega}_{l}$ : angular of the centroid frame in the world frame, expressed in the centroid frame.}
    \item {$\boldsymbol{b}_{g,j}$ : bias on the $j$-th gyrometer's measurement.}
    \item {$^{\Gamma}\boldsymbol{F}_{e}$ : unmodeled disturbance force exerted on the robot, expressed in the centroid frame.}
    \item {$^{\Gamma}\boldsymbol{T}_{e}$ : unmodeled disturbance moment exerted and expressed in the centroid frame.}
    \item {$\boldsymbol{x}_{c,i}$ : contact variable regrouping $\left\{ \boldsymbol{p}_{r,i},\boldsymbol{\Omega}_{r,i},{}^{\mathcal{C}}\boldsymbol{F}_{i},{}^{\mathcal{C}}\boldsymbol{T}_{i}\right\} $.}
    \item {$\boldsymbol{p}_{r,i}$ : $i$-th contact rest position.}
    \item {$\boldsymbol{\Omega}_{r,i}$ : $i$-th contact rest orientation.}
    \item {$^{\mathcal{C}}\boldsymbol{F}_{i}$ : $i$-th contact force.}
    \item {$^{\mathcal{C}}\boldsymbol{T}_{i}$ : $i$-th contact moment.}
    \end{itemize}
    
    \noindent Centroid frame's accelerations:
    \begin{itemize}[leftmargin=1em]
    \item {$\boldsymbol{a}_{l}$ : linear acceleration of the centroid frame in the world frame, expressed in the centroid frame.}
    \item {$\dot{\boldsymbol{\omega}}_{l}$ : angular acceleration of the centroid frame in the world frame, expressed in the centroid frame.}
    \end{itemize}
    
    \noindent Measurements:
    \begin{itemize}[leftmargin=1em]
    \item {$\boldsymbol{y}_{a,j}$ : measurement of the $j$-th accelerometer.}
    \item {$\boldsymbol{y}_{g,j}$ : measurement of the $j$-th gyrometer.}
    \item {$\boldsymbol{y}_{F,i}$ : measurement of the force sensor at the $i$-th contact.}
    \item {$\boldsymbol{y}_{T,i}$: measurement of the moment sensor at the $i$-th contact.}
    \end{itemize}
    
    \noindent Inputs:
    \begin{itemize}[leftmargin=1em]
    \item {$\left\{{}^{\Gamma}\boldsymbol{I}, {}^{\Gamma}\dot{\boldsymbol{I}}\right\}$ : total inertia matrix of the robot expressed in the centroid frame and its derivative.}
    \item {$\left\{{}^{\Gamma}\boldsymbol{\sigma}, {}^{\Gamma}\dot{\boldsymbol{\sigma}}\right\}$ : total angular momentum of the robot expressed in the centroid frame and its derivative.}
    \item {$\left\{^{\Gamma}\boldsymbol{F}_{res}, ^{\Gamma}\boldsymbol{T}_{res}\right\}$ : resultant wrench measured by the sensors not associated with currently set contacts, expressed in the centroid frame.}
    \item {$\boldsymbol{\Xi}_{i}$ : input variables related to the contact $i$.}
    \item {$\left\{\boldsymbol{\check{p}}_{r,i}, \boldsymbol{\check{\Omega}}_{r,i}\right\}$ : initial guess on the rest pose of the newly created contact $i$ in the world frame.}
    \item {$\left\{ ^{\Gamma}\boldsymbol{p}_{\mathcal{C}_{i}},{}^{\Gamma}\boldsymbol{\Omega}_{\mathcal{C}_{i}},{}^{\Gamma}\dot{\boldsymbol{p}}_{\mathcal{C}_{i}},{}^{\Gamma}\boldsymbol{\omega}_{\mathcal{C}_{i}}\right\}$} : kinematics of the $i$-th contact frame in the centroid frame.
    \item {$\boldsymbol{\Psi}_{j}$ : input variables related to the $j$-th IMU.}
    \item {$\left\{ ^{\Gamma}\boldsymbol{p}_{\mathcal{S}_{j}},{}^{\Gamma}\boldsymbol{\Omega}_{\mathcal{S}_{j}},{}^{\Gamma}\dot{\boldsymbol{p}}_{\mathcal{S}_{j}},{}^{\Gamma}\boldsymbol{\omega}_{\mathcal{S}_{j}},{}^{\Gamma}\ddot{\boldsymbol{p}}_{\mathcal{S}_{j}}\right\}$ : kinematics of the $j$-th IMU frame in the centroid frame.}
    \item {$^{\Gamma}\ddot{\boldsymbol{p}}_{\mathcal{S}_{j}}$ : linear acceleration of the $j$-th IMU in the centroid frame. }
    \end{itemize}
    
    \noindent Section~\ref{subsec:state-vec-def}: Vector state definition:
    \begin{itemize}[leftmargin=1em]
    \item {$\mathcal{S}_{j}$ : frame of the $j$-th IMU.}
    \end{itemize}
    
    \noindent Section~\ref{subsec:Viscoelastic-model}: Viscoelastic model:
    \begin{itemize}[leftmargin=1em]
    \item {$\left\{\boldsymbol{p}_{\mathcal{C}_i}, \boldsymbol{R}_{\mathcal{C}_{i}, \dot{\boldsymbol{p}}_{\mathcal{C}_{i}}}, \boldsymbol{\omega}_{\mathcal{C}_{i}}\right\}$ : kinematics of the $i$-th contact's frame in the world frame obtained by forward kinematics from the centroid frame.}
    \item {$\left\{\boldsymbol{K}_{p,t}, \boldsymbol{K}_{d,t}\right\}$ : linear stiffness and damping of the contacts.}
    \item {$\left\{\boldsymbol{K}_{p,r}, \boldsymbol{K}_{d,r}\right\}$ : angular stiffness and damping of the contacts.}
    \end{itemize}
    
    }
\section{Appendix: Kinetics Observer's summary}\label{sec:appendix-ko-summary}
    
    \begin{figure*}[!ht]
        \centering
            \includegraphics[width=\textwidth]{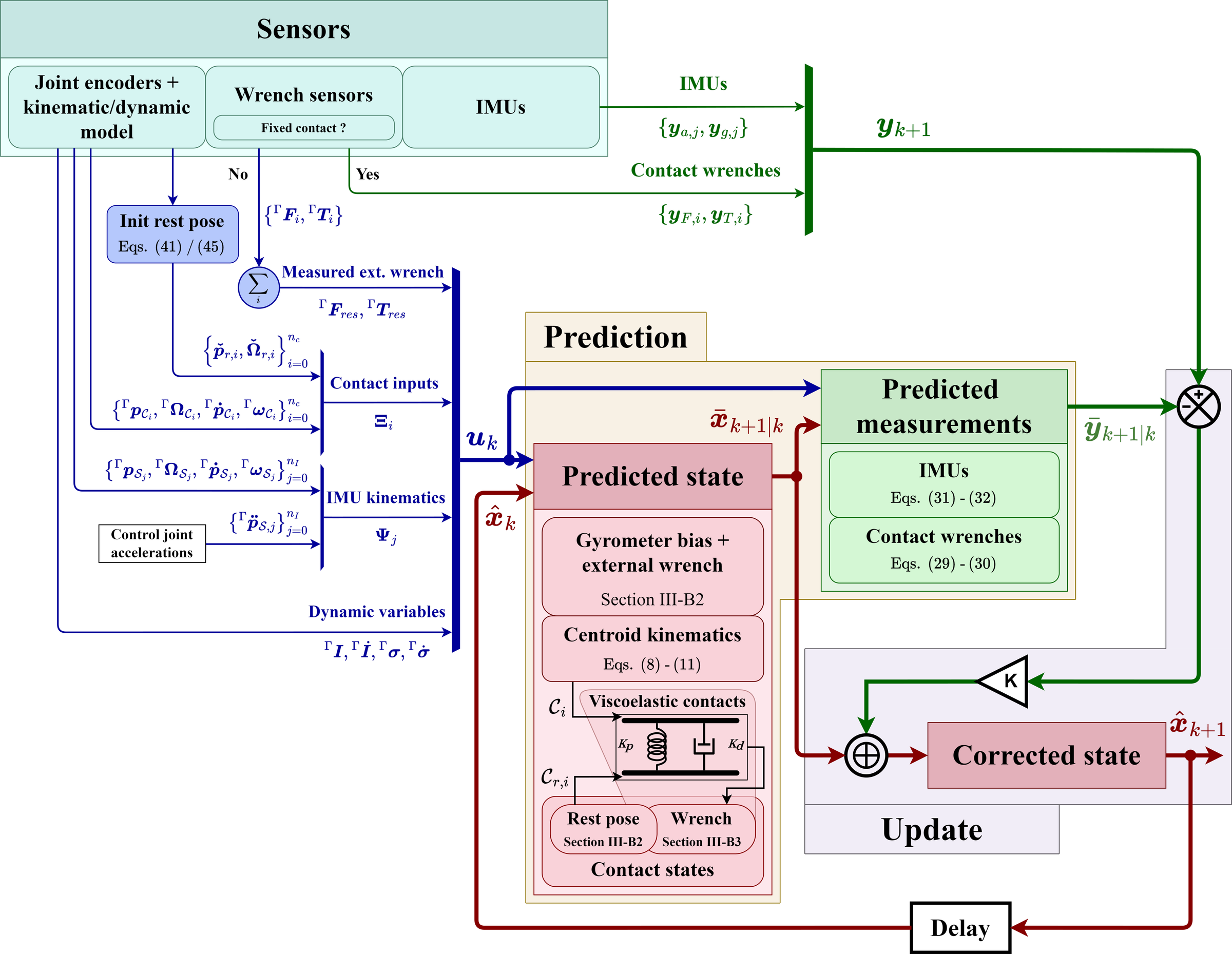}
        \caption{Summary of the structure of the Kinetics Observer. We present the main elements that make up the Kalman Filter with a focus on the dynamical model and the origin of each variable involved.}
        \label{fig:KineticsObserver-structure}  
      \end{figure*}

\theendnotes

% \bibliographystyle{SageH}
% \bibliography{main}

\end{document}